\documentclass[preprint,10pt]{elsarticle}

\usepackage[colorlinks=true, linkcolor=red, anchorcolor=red, citecolor=blue, urlcolor=red]{hyperref}
\usepackage{algorithmic}
\usepackage{algorithm}
\usepackage{xcolor}
\usepackage{graphicx}
\usepackage{colortbl}
\usepackage{pifont} 
\newcommand{\cmark}{\ding{51}}%
\newcommand{\xmark}{\ding{55}}%
\usepackage{multirow}
\usepackage{rotating}

\definecolor{HU2013-1}{RGB}{255, 0, 0}
\definecolor{HU2013-2}{RGB}{0, 255, 0}
\definecolor{HU2013-3}{RGB}{0, 0, 255}
\definecolor{HU2013-4}{RGB}{255, 255, 0}
\definecolor{HU2013-5}{RGB}{255, 0, 255}
\definecolor{HU2013-6}{RGB}{0, 255, 255}
\definecolor{HU2013-7}{RGB}{128, 0, 0}
\definecolor{HU2013-8}{RGB}{0, 128, 0}
\definecolor{HU2013-9}{RGB}{0, 0, 128}
\definecolor{HU2013-10}{RGB}{128, 128, 0}
\definecolor{HU2013-11}{RGB}{128, 0, 128}
\definecolor{HU2013-12}{RGB}{0, 128, 128}
\definecolor{HU2013-13}{RGB}{192, 192, 192}
\definecolor{HU2013-14}{RGB}{64, 64, 64}
\definecolor{HU2013-15}{RGB}{255, 128, 0}

\definecolor{PU-1}{RGB}{255, 0, 0}
\definecolor{PU-2}{RGB}{0, 255, 0}
\definecolor{PU-3}{RGB}{0, 0, 255}
\definecolor{PU-4}{RGB}{255, 255, 0}
\definecolor{PU-5}{RGB}{255, 0, 255}
\definecolor{PU-6}{RGB}{0, 255, 255}
\definecolor{PU-7}{RGB}{128, 0, 0}
\definecolor{PU-8}{RGB}{0, 128, 0}
\definecolor{PU-9}{RGB}{0, 0, 128}

\definecolor{IP-1}{RGB}{255, 0, 0}
\definecolor{IP-2}{RGB}{0, 255, 0}
\definecolor{IP-3}{RGB}{0, 0, 255}
\definecolor{IP-4}{RGB}{255, 255, 0}
\definecolor{IP-5}{RGB}{255, 0, 255}
\definecolor{IP-6}{RGB}{0, 255, 255}
\definecolor{IP-7}{RGB}{128, 0, 0}
\definecolor{IP-8}{RGB}{0, 128, 0}
\definecolor{IP-9}{RGB}{0, 0, 128}
\definecolor{IP-10}{RGB}{128, 128, 0}
\definecolor{IP-11}{RGB}{128, 0, 128}
\definecolor{IP-12}{RGB}{0, 128, 128}
\definecolor{IP-13}{RGB}{192, 192, 192}
\definecolor{IP-14}{RGB}{64, 64, 64}
\definecolor{IP-15}{RGB}{255, 128, 0}
\definecolor{IP-16}{RGB}{128, 128, 255}

\definecolor{HH-1}{RGB}{255, 0, 0}
\definecolor{HH-2}{RGB}{0, 255, 0}
\definecolor{HH-3}{RGB}{0, 0, 255}
\definecolor{HH-4}{RGB}{255, 255, 0}
\definecolor{HH-5}{RGB}{255, 0, 255}
\definecolor{HH-6}{RGB}{0, 255, 255}
\definecolor{HH-7}{RGB}{128, 0, 0}
\definecolor{HH-8}{RGB}{0, 128, 0}
\definecolor{HH-9}{RGB}{0, 0, 128}
\definecolor{HH-10}{RGB}{128, 128, 0}
\definecolor{HH-11}{RGB}{128, 0, 128}
\definecolor{HH-12}{RGB}{0, 128, 128}
\definecolor{HH-13}{RGB}{192, 192, 192}
\definecolor{HH-14}{RGB}{64, 64, 64}
\definecolor{HH-15}{RGB}{255, 128, 0}
\definecolor{HH-16}{RGB}{128, 128, 255}
\definecolor{HH-17}{RGB}{255, 192, 203}
\definecolor{HH-18}{RGB}{165, 42, 42}
\definecolor{HH-19}{RGB}{255, 215, 0}
\definecolor{HH-20}{RGB}{75, 0, 130}
\definecolor{HH-21}{RGB}{255, 20, 147}
\definecolor{HH-22}{RGB}{173, 255, 47}

\usepackage{amssymb}
\usepackage{amsmath}

\begin{document}

\begin{frontmatter}



\title{Mamba-in-Mamba: Centralized Mamba-Cross-Scan in Tokenized Mamba Model for Hyperspectral Image Classification}


\author[label1]{Weilian Zhou} 
\author[label1]{Sei-ichiro~Kamata} 
\author[label2]{Haipeng~Wang} 
\author[label3]{Man Sing~Wong} 
\author[label4]{Huiying (Cynthia)~Hou} 

\affiliation[label1]{organization={Graduate School of Information, Production and Systems, Waseda University},
            city={Kitakyushu},
            country={Japan}}
\affiliation[label2]{organization={Key Laboratory of Electromagnetic Waves (EMW) Information, Fudan University},
            city={Shanghai},
            country={China}}
\affiliation[label3]{organization={Department of Land Surveying and Geo-Informatics, Hong Kong Polytechnic University},
            city={Hong Kong},
            country={China}}
\affiliation[label4]{organization={Department of Building Environment and Energy Engineering, Hong Kong Polytechnic University},
            city={Hong Kong},
            country={China}}

\begin{abstract}
Hyperspectral image (HSI) classification is pivotal in the remote sensing (RS) field, particularly with the advancement of deep learning techniques. Sequential models, such as Recurrent Neural Networks (RNNs) and Transformers, have been tailored to this task, offering unique viewpoints. However, they face several challenges: 1) RNNs struggle with aggregating central features and are sensitive to interfering pixels; 2) Transformers require extensive computational resources and often underperform with limited HSI training samples. To address these issues, recent advances have introduced State Space Models (SSM) and Mamba, known for their lightweight and parallel scanning capabilities within linear sequence processing, providing a balance between RNNs and Transformers. However, deploying Mamba as a backbone for HSI classification has not been fully explored. Although there are improved Mamba models for visual tasks, such as Vision Mamba (ViM) and Visual Mamba (VMamba), directly applying them to HSI classification encounters problems. For example, these models do not effectively handle land-cover semantic tokens with multi-scale perception for feature aggregation when implementing patch-based HSI classifiers for central pixel prediction. Hence, the suitability of these models for this task remains an open question. In response, this study introduces the innovative Mamba-in-Mamba (MiM) architecture for HSI classification, marking the pioneering deployment of Mamba in this field. The MiM model includes: 1) A novel centralized Mamba-Cross-Scan (MCS) mechanism for transforming images into efficient-pair sequence data; 2) A Tokenized Mamba (T-Mamba) encoder that incorporates a Gaussian Decay Mask (GDM), a Semantic Token Learner (STL), and a Semantic Token Fuser (STF) for enhanced feature generation and concentration; and 3) A Weighted MCS Fusion (WMF) module coupled with a Multi-Scale Loss Design to improve model training efficiency. Experimental results from four public HSI datasets with fixed and disjoint training-testing samples demonstrate that our method outperforms existing baselines and is competitive with state-of-the-art approaches, highlighting its efficacy and potential in HSI applications.

\end{abstract}

\begin{graphicalabstract}
\end{graphicalabstract}

\begin{highlights}
\item  We explore the potential design of Mamba architecture, Mamba-in-Mamba model, for HSI classification task thereby achieving the better competitive performances and satisfying efficiency.

\item  To fully explore the ability of Mamba in visual task, we propose a novel Tokenized Mamba encoder with Gaussian Decay Mask, Semantic Token Learner, and Semantic Token Fuser to improve its suitability, aligning with multi-scale features for the HSI classification.

\item  We propose a centralized Mamba-Cross-Scan to comprehensively and continuously scan the image patch into several pair-sequences diversely. It can fully and lightly understand the HSI patch within the concentration on the center pixel for patch-wise HSI classifiers.

\item This study shows very competitive and even SOTA performance on four datasets, Indian Pines, Pavia University, Houston 2013, and WHU-Hi-HongHu, with fixed and disjoint training-testing samples, demonstrating the feasibility and efficiency.

\end{highlights}


\end{frontmatter}

\section{Introduction}
\label{introduction}
Hyperspectral imaging (HSI) is a sophisticated remote sensing (RS) technique that captures images across numerous dense and continuous spectral bands. This method spans the electromagnetic spectrum from visible to infrared wavelengths, often utilizing satellites or unmanned aerial vehicles (UAVs) \cite{c1}. Unlike traditional imaging methods, HSI provides detailed spectral information for each pixel, enabling a more precise characterization of materials \cite{hyperspectral imaging system} and objects on the Earth's surface \cite{c2}. The comprehensive data obtained from HSI makes it invaluable in various applications including agriculture \cite{agriculture}, environmental monitoring \cite{environment}, geology \cite{geology}, urban planning \cite{urban_planning}, and defense and security \cite{defense}. Given the broad utility of this technology, accurate HSI classification, which assigns a land-cover label to each pixel, is critical. This classification task is not only essential for enhancing the effectiveness of these applications but has also become a focus of research within the RS community \cite{c7}.

In recent years, the deployment of deep learning-based models for HSI classification has seen extensive growth \cite{c8}. These models typically employ a `patch-wise' learning framework, where a local patch around each pixel is processed to assign semantic labels \cite{c9, c10}. Notably, sequential models like Recurrent Neural Networks (RNNs) \cite{rnn3, rnn4, rnn5, rnn6, song multidirectional rnn} and Transformers \cite{vit, vit1, vit2, vit3, vit4, vit5, vit6, vit7, vit8, vit9, vit10, vit11, vit12, vit13, vit14, vit15}, originally developed for natural language processing (NLP), have been adapted to HSI classification. They have demonstrated impressive results by utilizing many-to-one or many-to-many schemes to capture contextual features within pixels or partitioned windows of one HSI patch, resulting in a comprehensive feature representation. Despite their success, both types of models face shortcomings that warrant further investigation to enhance their practicability and efficiency.

First, RNNs process an HSI patch by flattening it into a sequence and employing a many-to-one learning framework. This approach selects the output at the last step as the representative feature for the entire patch. However, RNNs inherently favor the latter steps of the input sequence, resulting in a feature representation that is biased towards the final processed steps. This bias can adversely affect model training, particularly when using the cross-entropy criterion to compare the predicted label with the true label of the central pixel. Although some strategies use attention-based weighted summations to balance the outputs across all steps, the attention values often remain skewed towards later pixels, diminishing the influence of the central pixel. Furthermore, RNNs also struggle to handle large HSI patches effectively due to their sequential processing nature, which limits parallel computation across pixels. Larger patches convert into longer sequences, increasing the model's exposure to interfering pixels with labels differing from the central pixel. This discrepancy can significantly degrade training performance as the model becomes more sensitive to peripheral pixel variations rather than focusing on the central context.

Second, to address the inflexibility of RNNs, Transformer-based methods have been introduced, utilizing parallel self-attention to enhance performance \cite{self-attention}. While these methods overcome some of RNNs' limitations through parallel processing and self-attention mechanisms, they hold their own set of shortcomings: 1) The self-attention mechanism scales quadratically with the size of HSI patches, leading to significant computational overhead. This becomes especially problematic when larger patches are used to capture more spatial or contextual information around the center pixel, making the approach resource-intensive and less suited for hardware-constrained environments; 2) Transformers often have a large number of parameters, increasing the risk of overfitting, especially with the limited training data typically available in HSI classification; 3) Unlike RNNs or convolutional neural networks (CNNs) \cite{cnn1, cnn2, cnn3}, which inherently capture recurrent features or local spatial features, Transformers do not have this inductive bias. While positional embeddings are used to introduce some spatial context, their effectiveness is limited without substantial training samples, potentially reducing the model's ability to accurately capture spatial contexts in HSIs. Although some research \cite{rnn trans 1, rnn trans 2, rnn trans 3, rnn trans 4} has explored hybrid approaches that integrate RNNs or CNNs with Transformers to leverage complementary strengths. However, these methods may still increase the computational burden and complexity, potentially leading to suboptimal performance.

Third, the use of sequential models for image interpretation presents challenges, particularly in how images are arranged into sequences due to the non-causal nature of visual data: 1) Many approaches rely on a simple `\textit{Raster}' scan to flatten the image, which can result in discontinuities, or `jump connections,' when the scan reaches the boundaries of the image; 2) Previous works \cite{multiscanicpr, multi-lstm} have demonstrated that a continuous scan approach yields better results in HSI classification. It can maintain the highest spatial correlation by ensuring continuity in scanning across pixels. It also preserves multi-directional information inherent in the image’s direction-sensitive properties. Despite these advantages, existing one \cite{multi-rt} often processes scans one by one in a unidirectional manner, then aggregate the results. This method does not necessarily prioritize the central pixel’s features during model training and then increase the feature redundancy and computational burden.

In response to three scenarios discussed above, two critical needs emerge: 1) Alternative Base Model: There is a need for an alternative model that enhances parallel computing or incorporates attention mechanisms, offering greater flexibility than RNNs. This model should also demand fewer resources and exhibit less complexity, proving lighter and more efficient than Transformers, and 2) Enhanced Scanning Method: A refined scanning method is required that supports continuous and multi-directional processing. This method should provide image-aware cognition without being redundant, effectively capturing the spatial correlations and dynamics inherent in image data.

Recently, structured state space sequence (S4) models \cite{s4} and the Mamba architecture \cite{mamba, survey1, survey2, survey3} are brought into the spotlight. These models are adept at handling long-sequence data and can scale linearly with sequence length. The Mamba architecture, in particular, builds on S4 by adding a selective-scan mechanism that adaptively attentions on relevant inputs, showing superior computational efficiency compared to Transformers. Although the Mamba-based methods has shown promising results in some visual tasks \cite{ssm360, vim, vmamba, rsmamba, changemamba, rs3mamba} (e.g. Vision Mamba (ViM), Visual Mamba (VMamba), etc), its potential for HSI classification has yet to be fully explored. Directly applying these advanced Mamba-based models on HSI classification task may encounter unsuitability and problems. Therefore, there is a critical need to develop a suitable vision-based Mamba model specifically designed for HSI classification.

\begin{figure}[t]
    \centering
    \includegraphics[scale=0.35]{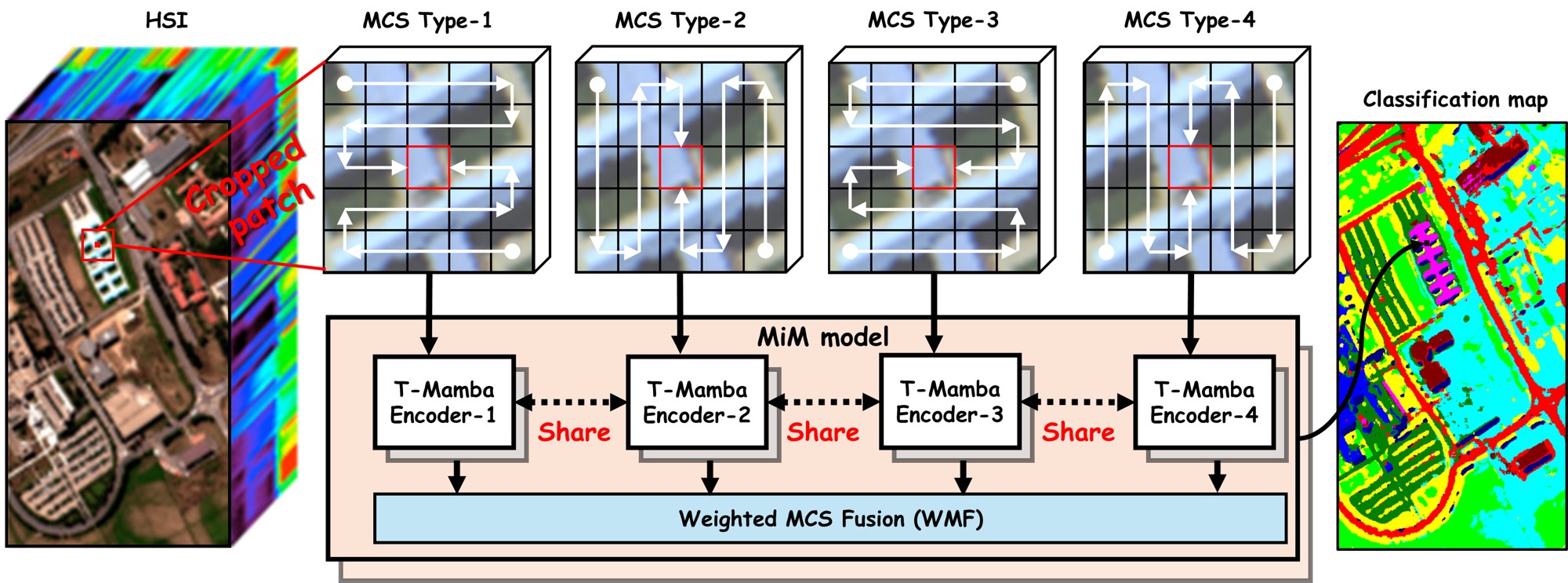}
    \caption{Conceptual diagram of the Mamba-in-Mamba (MiM) model. Initially, four distinct types of centralized Mamba-Cross-Scan (MCS) are employed to convert a cropped HSI patch into sequential data. It is then fed into four shared Tokenized Mamba (T-Mamba) encoders designed to process sequences effectively. To adjust the impact of each MCS type, a Weighted MCS Fusion (WMF) is conducted, ensuring a balanced integration of the features extracted from each MCS type. Finally, the outputs from cascaded down-sampling-driven MiM layers are fed into a decoder for classification.}
    \label{fig:concept}
\end{figure}

To this end, in this study, we introduce a innovative approach—the Mamba-in-Mamba (MiM) architecture—specifically designed for HSI classification. This method marks the pioneering application of the ViM in this task. The MiM model comprises several innovative components: 1) Centralized Mamba-Cross-Scan (MCS): This new scanning mechanism transforms images into diverse half-directional sequences, offering a more efficient, lightweight, and rational approach, 2) Tokenized Mamba (T-Mamba) Encoder: Integrates a Gaussian Decay Mask (GDM), a Semantic Token Learner (STL), and a Semantic Token Fuser (STF) to significantly enhance representative feature generation and concentration, and 3) Weighted MCS Fusion (WMF) module: Works in conjunction with a Multi-Scale Loss Design to boost decoding efficiency and overall model performance. The conceptual framework of this study is illustrated in Fig. \ref{fig:concept}.

The contributions of this study are summarized as follows:
\begin{itemize}
    \item [1)] We explore the potential applications of the improved ViM architecture, multi-scale MiM model, for HSI classification task thereby achieving the better performances and efficiency.
    \item [2)] To fully explore the ability of ViM, we propose a novel T-Mamba encoder with GDM, STL, and STF to make it more suitable for the HSI classification.
    \item [3)] We propose centralized MCS to comprehensively and continuously scan the HSI patch into several pair-sequences diversely, which can fully and lightly understand the HSI patch within the highlights on the center pixel for patch-wise HSI classifiers.
    \item [4)] The proposal shows very competitive and even SOTA performance on four datasets with fixed and disjoint training-testing samples, demonstrating the feasibility and efficiency.
\end{itemize}

The rest of the paper is organized as follows: Section \ref{sec:preliminary} presents the preliminary knowledge of Mamba, Section \ref{sec:method} details the methodology, Section \ref{sec:experiment} discusses the experimental setup and results, and Section \ref{sec:conlusion} concludes the study. 

\section{Preliminary}\label{sec:preliminary}
This section provides a foundational concepts associated with State Space Models (SSM) and the Mamba model, presenting the knowledge necessary to proceed this study.

\subsection{\textbf{Structured State Space Sequence Model (S4)}}
The SSM-based models, represent a recent class of sequence models within the field of deep learning. They are broadly related to RNNs, CNNs, and classical state space models. 

Inspired by continuous system theory \cite{continuous system theory}, SSMs are commonly considered linear time-invariant systems that map a signal $\mathbf{x}(t) \in \mathbb{R}^{1 \times \mathcal{D}}$ to $\mathbf{y}(t) \in \mathbb{R}^{1 \times \mathcal{D}}$ through a latent state $\mathbf{h}(t) \in \mathbb{R}^{1 \times \mathcal{N}}$. Here, $t$ represents the time step, $\mathcal{D}$ is the initial dimension, and $\mathcal{N}$ is the dimension of the state.

Mathematically, SSMs are typically formulated as linear ordinary differential equations (ODEs) \cite{ODE}, as follows:
\begin{subequations}
\begin{align}
    \mathbf{h}'(t) &= \mathbf{A}\, \mathbf{h}(t) + \mathbf{B}\, \mathbf{x}(t), \label{eq:ssm1} \\
    \mathbf{y}(t) &= \mathbf{C}\, \mathbf{h}(t) + \mathbf{D}\, \mathbf{x}(t), \label{eq:ssm2}
\end{align}
\end{subequations}
where $\mathbf{A} \in \mathbb{R}^{\mathcal{N} \times \mathcal{N}}$ is the evolution parameter, $\mathbf{B} \in \mathbb{R}^{\mathcal{N} \times \mathcal{D}}$ correctly aligns dimensions for input-to-state transformation, and $\mathbf{C} \in \mathbb{R}^{\mathcal{D} \times \mathcal{N}}$ is the projection parameter. $\mathbf{D} \in \mathbb{R}^{1 \times \mathcal{D}}$, which typically supports skip connections, ensures that $\mathbf{y}(t) \in \mathbb{R}^{1 \times \mathcal{D}}$.

S4 is a discrete version of a continuous system, developed to address the significant challenges posed when integrating continuous systems into deep learning algorithms. A timescale parameter $\mathbf{\Delta}\in\mathbb{R}^{1 \times\mathcal{D}}$ is used to transform the continuous parameters $\mathbf{A}$ and $\mathbf{B}$ into discrete parameters $\mathbf{\overline{A}}\in\mathbb{R}^{\mathcal{N} \times \mathcal{N}}$ and $\mathbf{\overline{B}}\in\mathbb{R}^{\mathcal{N} \times \mathcal{D}}$. The transformation commonly employs a zero-order hold (ZOH) method\footnote{\url{https://en.wikipedia.org/wiki/Zero-order_hold}}, defined as:
\begin{subequations}
\begin{align}
\mathbf{\overline{A}} & = \exp(\mathbf{\Delta} \mathbf{A}),  \\
\mathbf{\overline{B}} & = (\mathbf{\Delta} \mathbf{A})^{-1} (\exp(\mathbf{\Delta} \mathbf{A}) - \mathbf{I}) \odot \mathbf{\Delta} \mathbf{B}, 
\end{align}
\end{subequations}

\begin{figure}[t]
    \centering
    \includegraphics[scale=0.33]{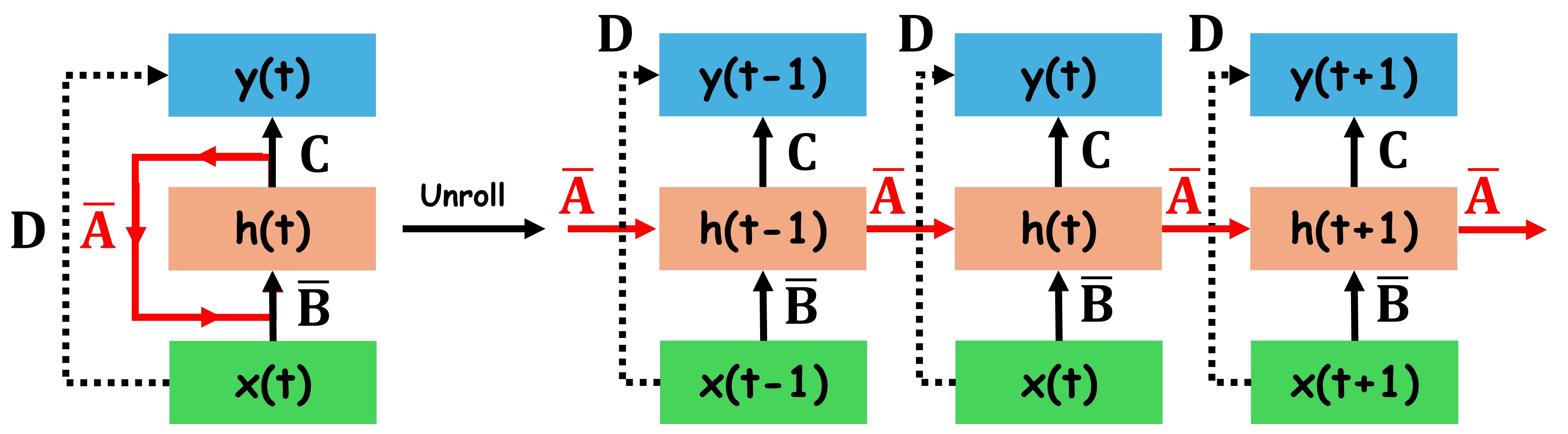}
    \caption{The graphic illustration of S4 model. $\mathbf{x}(t)$ is a input feature (i.e. scalar or vector) at time step $t$. $\mathbf{h}(t)$ represents the latent state feature. $\mathbf{\overline{A}}$, $\mathbf{\overline{B}}$, $\mathbf{C}$, and $\mathbf{D}$ represents the transformation matrices.}
    \label{fig:S4 model}
\end{figure}

Therefore, the discretized versions of Eqs. \ref{eq:ssm1} and \ref{eq:ssm2}, can be reformulated not as a function-to-function mapping, but rather as a sequence-to-sequence model, as shown in Fig. \ref{fig:S4 model}:
\begin{subequations}
\begin{align}
    \mathbf{h}(t) & = \mathbf{\overline{A}}\, \mathbf{h}(t-1) + \mathbf{\overline{B}}\, \mathbf{x}(t),\label{eq:s41}  \\
    \mathbf{y}(t) & = \mathbf{C}\, \mathbf{h}(t) + \mathbf{D}\, \mathbf{x}(t).\label{eq:s42}
\end{align}
\end{subequations}

Finally, the models compute the output through a global convolution operation as follows:
\begin{subequations}
\begin{align}
    \mathbf{\overline{K}} & = \Big(\mathbf{C}\, \mathbf{\overline{B}}, \mathbf{C}\, \mathbf{\overline{A}}\, \mathbf{\overline{B}},\, \ldots,\, \mathbf{C}\, \mathbf{\overline{A}}^{L-1}\, \mathbf{\overline{B}}\Big), \\
    \mathbf{y} & = \mathbf{x} \star \mathbf{\overline{K}},
\end{align}
\end{subequations}
where $L$ is the length of the input sequence $\mathbf{x}$, and $\mathbf{\overline{K}} \in \mathbb{R}^{\mathcal{L}}$ is a structured SSM kernel. $\star$ denotes the convolutional operation.

The notable work \cite{LSSL} states that the matrices $\mathbf{A}$, $\mathbf{B}$, and $\mathbf{C}$ in SSM share a crucial attribute known as Linear Time Invariance (LTI). It means that they are constant and remain unchanged for each token generated by the SSM. In essence, regardless of the input sequence presented to the SSM, the values of $\mathbf{A}$, $\mathbf{B}$, and $\mathbf{C}$ remain fixed, resulting in a static representation that is not content-aware. Additionally, the matrices $\mathbf{A}$, $\mathbf{B}$, and $\mathbf{C}$ are independent of the input, as their dimensions $\mathcal{N}$ and $\mathcal{D}$ do not vary, further emphasizing their time-invariant nature.

\subsection{\textbf{Selective Scan-Based S4 Model (S6)}}

To address the limitations identified in the S4 model, Mamba enhances the performance of the S4 model by introducing a selective scan mechanism as the core S4 operator, termed as S6 model. It allows the continuous parameters to vary with the input, thereby enhancing selective information processing across sequences.

It modifies the S4 model by introducing adaptive parameterization mechanisms for the matrices $\mathbf{B}$, $\mathbf{C}$, and the timescale parameter $\mathbf{\Delta}$ as follows:
\begin{subequations}
\begin{align}
    \mathbf{B} & = {Projection}_{B}(\mathbf{x}), \\
    \mathbf{C} & = {Projection}_{C}(\mathbf{x}), \\
    \mathbf{\Delta} & = \tau_{\Delta} ({Parameter} + {Projection}_{\Delta}(\mathbf{x})),
\end{align}
\end{subequations}
where $\mathbf{x} \in \mathbb{R}^{\mathcal{B} \times \mathcal{L} \times \mathcal{D}}$. Here, $\mathcal{B}$ represents the batch size, $\mathcal{L}$ denotes the sequence length, and $\mathcal{D}$ is the sequence dimension. The functions ${Projection}_{B}()$ and ${Projection}_{C}()$ perform linear projections from the $\mathcal{D}$-dimensional space to the $\mathcal{N}$-dimensional space, resulting in $\mathbf{B} \in \mathbb{R}^{\mathcal{B} \times \mathcal{L} \times \mathcal{N}}$ and $\mathbf{C} \in \mathbb{R}^{\mathcal{B} \times \mathcal{L} \times \mathcal{N}}$. Additionally, ${Projection}_{\Delta}()$ transforms $\mathbf{x}$ to a size of $\mathbb{R}^{\mathcal{B} \times \mathcal{L} \times 1}$, which is then broadcast to the $\mathcal{D}$-dimension. $\tau_{\Delta}$, typically a softplus activation \cite{softplus}, adjusts the scale, resulting in a new $\mathbf{\Delta}$ of size $\mathbb{R}^{\mathcal{B} \times \mathcal{L} \times \mathcal{D}}$.

In a summary, Mamba enhances the S4 model by making the matrices $\mathbf{B}$, $\mathbf{C}$, and the step size $\mathbf{\Delta}$ input-dependent, incorporating the sequence length and batch size to ensure that each input token is processed with distinct matrices. Moreover, these dynamic matrices selectively filter which components to maintain or omit in the hidden state, drawing an analogy to the `recurrence' mechanism in RNNs. This selective processing is implemented through a `parallel scan' operation\footnote{\url{https://developer.nvidia.com/gpugems/gpugems3/part-vi-gpu-computing/chapter-39-parallel-prefix-sum-scan-cuda}} in CUDA, which segments the sequences, processes them iteratively, and combines them efficiently. This integration of dynamic matrices and parallel processing exemplifies how Mamba captures the fast and adaptable nature of sequence handling, thereby enhancing the model's adaptability.

\subsection{\textbf{Mamba-Based Works}}
As a cutting-edge foundational model, Mamba has recently gained popularity. There have been various implementations exploring its applicability to diverse tasks.

Zhu \textit{et al.} \cite{vim} pioneered the ViM architecture for general vision tasks such as object detection and semantic segmentation, employing bidirectional scans for precise image patch positioning prior to processing with the Mamba encoder. Building on this framework, Chen \textit{et al.} \cite{rsmamba} enhanced the Mamba with a dynamic multi-path activation mechanism, significantly improving performance across various RS image classification datasets. Concurrently, Liu \textit{et al.} \cite{vmamba} developed VMamba, which utilizes 2D-Selective-Scan (SS2D) to transform non-causal visual images into ordered patch sequences for applications in image classification and object detection. Further extending this methodology, Chen \textit{et al.} \cite{changemamba} applied VMamba to RS change detection tasks, encompassing binary change detection, semantic change detection, and building damage assessment, with VMamba serving as the sequential encoder to model spatio-temporal relationships. Additionally, Zhao \textit{et al.} \cite{rsmamba2} introduced RS-Mamba for dense prediction in extensive RS images, incorporating the Omni-directional Selective Scan Module (OSSM) for feature aggregation. In the HSI classification domain, Jia \textit{et al.} \cite{spectralmamba} adapted the standard SSM-based model, SpectralMamba, for spectral domain processing, while Yang \textit{et al.} \cite{hsimamba} combined Mamba’s spectral processing with CNN-based spatial feature extraction for HSI classification. Moreover, Sheng \textit{et al.} \cite{dualmamba} proposed a lightweight spectral-spatial Mamba-convolution model for HSI classification, utilizing spatial unidirectional and spectral bidirectional scanning for enhanced feature alignment.

Despite these advancements, to our knowledge, the application of ViM with multi-directional scanning for HSI classification has not yet been fully developed or explored. Meanwhile, directly using ViM can not well handle the refined semantic token features and miss the multi-scale features existed in the HSI patch. Hence, developing an optimized ViM architecture for HSI remains an open question. Thus, we aim to explore this gap and propose a tailored HSI-driven ViM architecture.

\begin{figure}[t]
\centering
\includegraphics[scale=0.365]{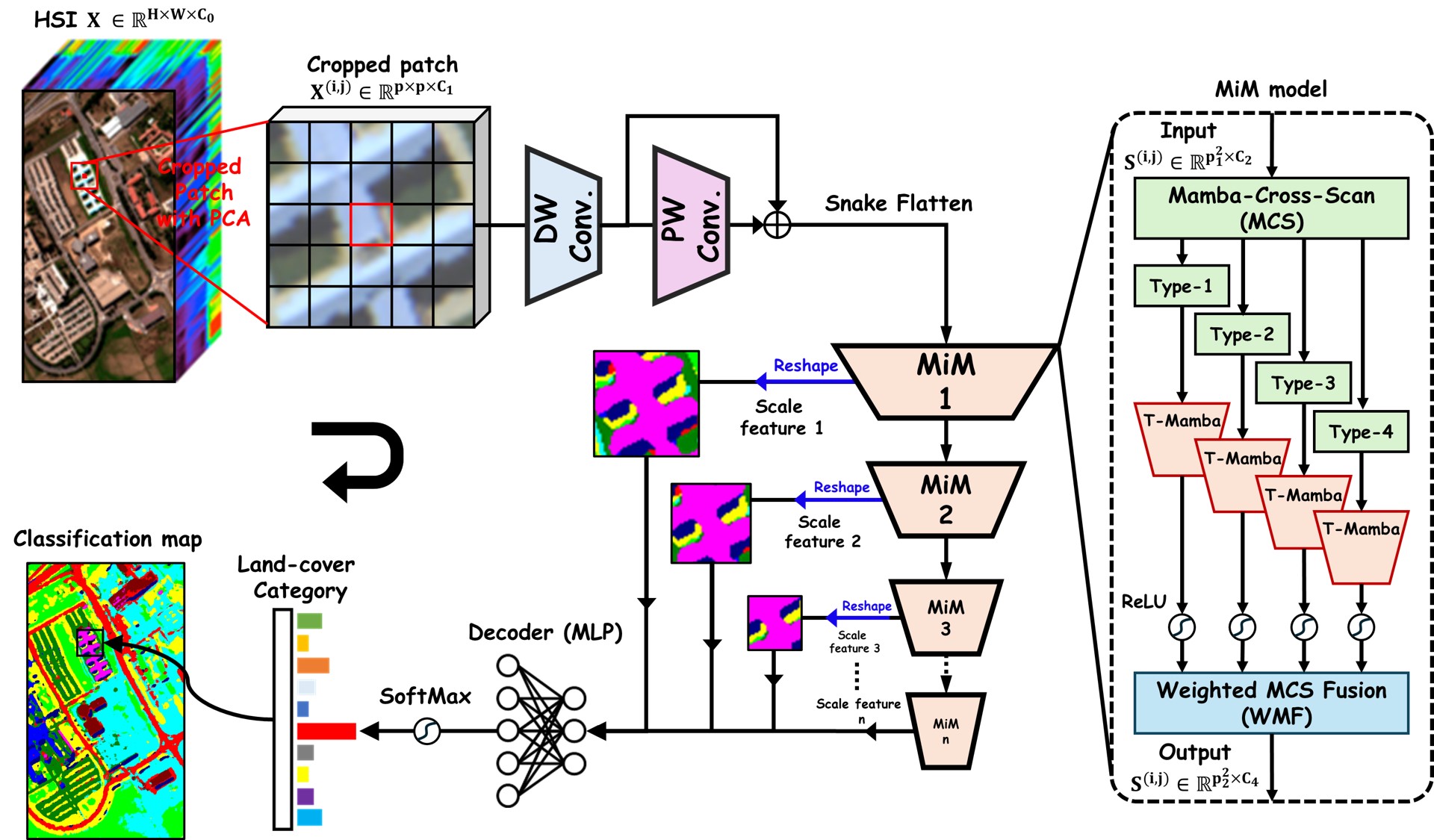}
\caption{The overall structure. It begins with a cropped HSI patch after PCA and then goes through a depth-wise convolution and a point-wise convolution for local feature generation and dynamic positional embedding. It is the reshaped by snake flattening before MiM model. In the MiM model, flattened image is first proposed by MCS, as shown in Fig. \ref{fig:four types of cross mamba scan}, to obtain four types of pair-sequences. Each pair of sequences processed by a specially designed T-Mamba encoder with sharing parameters, as detailed in Fig. \ref{fig:T_mamba}, to extract contextual features. The outputs from the four T-Mamba encoders are combined through a WMF module with dynamic weighted scan fusion. Concurrently, fused feature at current scale is recorded for final multi-scale decoding. This process is repeated, with the feature sizes being progressively down-sampled until they reach a spatial size of $1$. Finally, the accumulated multi-scale features from each MiM iteration are compiled to feed into a decoder for model training and decoding.}\label{fig:whole strcuture}
\end{figure}

\section{Methodology\label{sec:method}}
This section firstly presents (\textit{A}) methodological overview briefly. Then, it presents the detailed contents: (\textit{B}) Centralized MCS, (\textit{C}) T-Mamba Encoder, (\textit{D}) WMF Module, (\textit{E}) Multi-Scale Loss Design. The overall structure and the pseudo algorithm is shown in Fig. \ref{fig:whole strcuture} and Algorithm \ref{alg:proposed_method}, respectively. \textit{Due to the paper formatting, the summary of algorithm is placed at end of the article.}

\subsection{\textbf{Methodological Overview}}
Our method is composed of several following steps:
\begin{enumerate}
    \item[1)] We introduce a centralized MCS mechanism that transforms an HSI patch into four types of sequences, each featuring two directional scans. 
    \item[2)] The T-Mamba encoder is engineered to efficiently manage two directional scans. It utilizes the GDM, a downsample-driven STL, and an attention-driven STF to facilitate the effective learning of representative tokens.
    \item[3)] The MiM model processes the four types of sequences generated by the MCS across four T-Mamba encoders, setting the foundation for subsequent feature fusion.
    \item[4)] A WMF is developed to dynamically assign different weights to the outputs from the MiM model. Meanwhile, a scale feature, derived from the fused outputs, is used to represent this phase for later decoding.
    \item[5)] Due to the down-sampling-like operation by the MiM model, we iterate the aforementioned procedures until the spatial size of the feature reduces to 1.
    \item[6)] The accumulated multi-scale features, are prepared for the decoder. Each feature is linked to a specific loss function to optimize performance during training.
\end{enumerate}

Detailed explanations for each component are provided in the subsequent subsections later.

\subsection{\textbf{Centralized Mamba-Cross-Scan (MCS)}}
The original HSI symbolized as $\mathbf{\mathbf{X}}\in\mathbb{R}^{\mathcal{H} \times \mathcal{W} \times \mathcal{C}_0}$ can be defined as follows:
\begin{equation}
\label{eq:X}
\centering
    \mathbf{\mathbf{X}} = \Big\{\mathbf{\mathbf{x}}^{(i,j)} \, \Big| \,  i=0,1,...,\mathcal{H}-1, \, j=0,1,..., \mathcal{W}-1 \Big\},
\end{equation}
where $\mathbf{\mathbf{x}}^{(i,j)}\in\mathbb{R}^{\mathcal{C}_0}$ signifies the spectral signature at the spatial position $(i, j)$. Consequently,
\begin{equation}
\centering
	    \mathbf{\mathbf{X}} = \left( \begin{matrix}
	               \mathbf{\mathbf{x}}^{(0,0)} & \mathbf{\mathbf{x}}^{(0,1)} & \dots & \mathbf{\mathbf{x}}^{(0, \mathcal{W}-1)}\\
	               \mathbf{\mathbf{x}}^{(1,0)} & \mathbf{\mathbf{x}}^{(1,1)} & \dots & \mathbf{\mathbf{x}}^{(1,\mathcal{W}-1)}\\
	               \vdots & \vdots & \ddots & \vdots\\
                   \mathbf{\mathbf{x}}^{(\mathcal{H}-1,0)} & \mathbf{\mathbf{x}}^{(\mathcal{H}-1,1)}  & \dots &\mathbf{\mathbf{x}}^{(\mathcal{H}-1,\mathcal{W}-1)}.
	\end{matrix} \right).
\end{equation}

Designed as a patch-wise framework, the input for the neural model is a cropped patch from the original HSI. 
For one cropped HSI patch \( \mathbf{X}^{(i,\,j)} \) centered at \( \mathbf{x}^{(i,\,j)} \) with patch size \( p \), an odd number, the patch can be represented as:
\begin{equation}
    \mathbf{X}^{(i,\,j)} = \Big\{\, \mathbf{x}^{(i+\alpha,\,j+\beta)} \, \Big| \, \alpha, \beta = -\frac{p-1}{2}, \ldots, \frac{p-1}{2} \Big\},
\end{equation}
where \( \alpha, \beta \in \mathbb{Z} \) and \( i+\alpha \) and \( j+\beta \) record the position.

Taking \( p = 5 \) as an example, the cropped HSI patch \( \mathbf{X}^{(i,\,j)} \) can be represented as:
\begin{equation}
\centering
\label{eq:Xij}
\mathbf{X}^{(i,\,j)} = \left( \begin{matrix}
    \mathbf{x}^{(i-2,\,j-2)} & \mathbf{x}^{(i-2,\,j-1)} & \dots & \mathbf{x}^{(i-2,\,j+2)} \\
    \mathbf{x}^{(i-1,\,j-2)} & \mathbf{x}^{(i-1,\,j-1)} & \dots & \mathbf{x}^{(i-1,\,j+2)} \\
    \vdots & \vdots & \ddots & \vdots \\
    \mathbf{x}^{(i+2,\,j-2)} & \mathbf{x}^{(i+2,\,j-1)} & \dots & \mathbf{x}^{(i+2,\,j+2)}
\end{matrix} \right),
\end{equation}
where \( 2 \leq i \leq \mathcal{H}-3 \), \( 2 \leq j \leq \mathcal{W}-3 \), and \( \mathbf{X}^{(i,\,j)} \in \mathbb{R}^{p \times p \times \mathcal{C}_0} \). The center pixel \( \mathbf{X}^{(i,\,j)}[2,2] \) is \( \mathbf{x}^{(i,\,j)} \).

The existing centralized scanning method, SS2D, as utilized in \cite{vmamba}, employs traditional \textit{Raster} flattening across the spatial domain of images. This approach can lead to misalignment of image contents, particularly through jump connections that occur when the scan meets image boundaries. Furthermore, SS2D does not fully embrace the two-dimensional structure of the image, limiting its ability to analyze from multiple perspectives.

Consequently, we have opted for the \textit{Snake} scan method (e.g., U-Turn scan), which has been endorsed in various studies \cite{multiscanicpr, multi-lstm} and serves as the foundation for our proposed MCS methodology. The \textit{Snake} scan method preserves spatial continuity between pixels or partitioned windows when transforming them into sequences, reducing the incidence of jump connections at image boundaries. This method's effectiveness is further supported by the \textit{multi-scanning strategy} concept \cite{multi-rt}, which demonstrates that using sequences scanned from multiple directions can enhance sequential model training.

\begin{figure}[t]
    \centering
    \includegraphics[scale = 0.42]{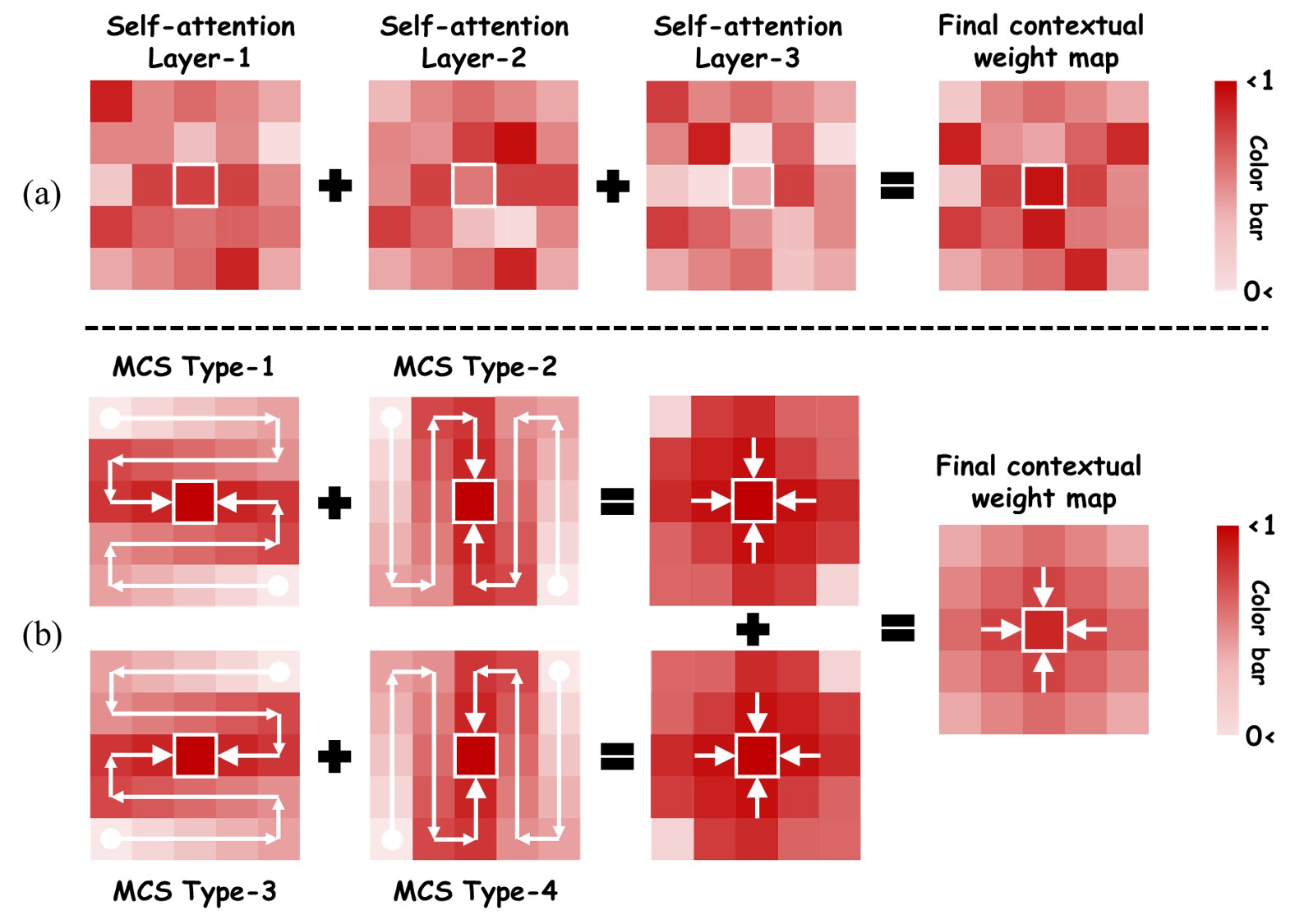}
    \caption{The conceptual difference between (a) self-attention mechanism in Transformers, and (b) proposed MCS mechanism with T-Mamba in overall structure for recognizing global contextual information in the patch.}
    \label{fig:MCS concept}
\end{figure}

Leveraging this background, we have developed four types of centralized MCS that optimize the transformation of images into sequences from various starting and ending points, ensuring both potential maximization and system efficiency. The fundamental distinction between the Transformer’s self-attention and our MCS mechanism is illustrated in Fig. \ref{fig:MCS concept}.

As depicted in Fig. \ref{fig:four types of cross mamba scan}, each MCS type initiates from two opposing positions and converges at the center pixel. This method is reminiscent of the bi-directional processing typical of RNNs but incorporates a centralized design, providing a novel approach to image sequence processing.

Before feeding the $\mathbf{X}^{(i,\,j)}$ into MCS, we first deploy the principal component analysis (PCA), depth-wise convolution (DWconv)\footnote{\url{https://github.com/rosinality/depthwise-conv-pytorch}} and point-wise convolution (PWconv)\footnote{\url{https://github.com/anshulpaigwar/Pointwise-Convolutional-Neural-Network}} on $\mathbf{X}^{(i,\,j)}$ for dimension reduction, local feature aggregation and dynamic positional embedding, as:
\begin{equation}
    \mathbf{X}^{(i,\,j)} \gets PWconv\Big(DWconv\Big(PCA\Big(\mathbf{X}^{(i,\,j)}\Big)\Big)\Big)
\end{equation}
where the PCA preserves the number of $\mathcal{C}_1$ components. The PWconv and DWconv outputs number of $\mathcal{C}_2$ feature dimensions, where $\mathcal{C}_2 > \mathcal{C}_1$. Hence, the resultant $\mathbf{X}^{(i,\,j)} \in\mathbb{R}^{p \times p \times \mathcal{C}_2}$.

\subsubsection{\textbf{MCS Type-1}}
\begin{figure}[t]
    \centering
    \includegraphics[scale=0.51]{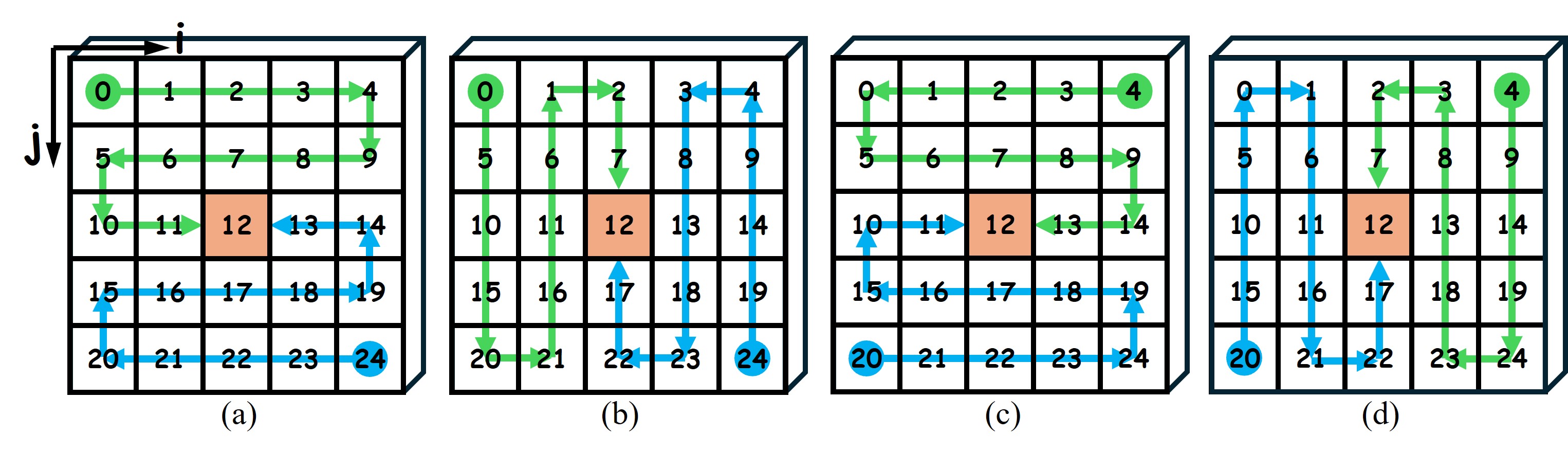}
    \caption{Four types of MCS with a $5 \times 5$ patch. (a) Type-1, (b) Type-2, (c) Type-3, and (d) Type-4. They cover the all the possible continuous \textit{snake} scanning for a patch.}
    \label{fig:four types of cross mamba scan}
\end{figure}

In details, as shown in Fig. \ref{fig:four types of cross mamba scan}(a), the type-1 MCS starts from the two positions, $\mathbf{x}^{(i-2,\, j-2)}$ and $\mathbf{x}^{(i+2,\, j+2)}$. Instead of directly preparing two directional sub-sequences, we firstly flatten the whole patch into a complete sequence by \textit{snake} scanning, starting from $\mathbf{x}^{(i-2,\, j-2)}$, as:
\begin{equation}
\begin{array}{ccccccccc}
    &\overbrace{\textcolor{green}{\mathbf{\mathbf{x}}^{(i-2,j-2)}}}^{Start} &\rightarrow &\mathbf{\mathbf{x}}^{(i-2,j-1)} &\rightarrow &\cdots &\rightarrow \mathbf{\mathbf{x}}^{(i-2,j+2)}\\

    & & & & & &\downarrow\\

    &\mathbf{\mathbf{x}}^{(i-1,j-2)} &\leftarrow &\mathbf{\mathbf{x}}^{(i-1,j-1)} &\leftarrow &\cdots 
    &\leftarrow \mathbf{\mathbf{x}}^{(i-1,j+2)}\\

    &\downarrow & & & & &\\

    &\mathbf{\mathbf{x}}^{(i,j-2)} &\rightarrow &\mathbf{\mathbf{x}}^{(i,j-1)} &\rightarrow &\cdots 
    &\rightarrow \mathbf{\mathbf{x}}^{(i,j+2)}\\

    & & & & & &\downarrow\\

    &\mathbf{\mathbf{x}}^{(i+1,j-2)} &\leftarrow &\mathbf{\mathbf{x}}^{(i+1,j-1)} &\leftarrow &\cdots 
    &\leftarrow \mathbf{\mathbf{x}}^{(i+1,j+2)}\\

    &\downarrow & & & & &\\

    &\mathbf{\mathbf{x}}^{(i+2,j-2)} &\rightarrow &\mathbf{\mathbf{x}}^{(i+2,j-1)} &\rightarrow &\cdots 
    &\rightarrow \underbrace{\textcolor{blue}{\mathbf{\mathbf{x}}^{(i+2,j+2)}}}_{End}\\

\end{array}
\end{equation}

Therefore, the generated complete sequence $\mathbf{S}_1^{(i,\,j)}\in\mathbb{R}^{p^2 \times \mathcal{C}_2}$ can be represented as:
\begin{equation}\label{eq:5}
    \begin{split}
    \mathbf{S}^{(i,j)}_1 = & [\textcolor{green}{\mathbf{x}^{(i-2,j-2)}}, \mathbf{x}^{(i-2,j-1)}, \cdots, \mathbf{x}^{(i-2,j+2)}, \cdots\\ 
    &\cdots, \mathbf{x}^{(i+2,j-2)}, \mathbf{x}^{(i+2,j-1)},\cdots, \textcolor{blue}{\mathbf{x}^{(i+2,j+2)}}]^\top,
    \end{split}
\end{equation}
where each element of the $\mathbf{S}_1^{(i,\,j)}$ is a spectral vector, in which the $\mathbf{S}_1^{(i,\,j)}[\frac{p^2-1}{2}]$ is the center step $\mathbf{x}^{(i,j)}$ of the sequence.

Subsequently, we split the $\mathbf{S}_1^{(i,\,j)}$ into two sub-sequences, $\mathbf{S}_{1,\,f}^{(i,\,j)}$ and $\mathbf{S}_{1,\,b}^{(i,\,j)}$. In this case, the $\mathbf{S}_{1,\,f}^{(i,\,j)}$ starts from $\mathbf{x}^{(i-2,\, j-2)}$, and ends at $\mathbf{x}^{(i,\,j)}$; meanwhile, the  $\mathbf{S}_{1,\,b}^{(i,\,j)}$ starts from $\mathbf{x}^{(i+2,\, j+2)}$, and ends at $\mathbf{x}^{(i,\,j)}$, as follows:
\begin{equation}\label{eq:split1}
    \mathbf{S}^{(i,j)}_{1,\,f},\, \mathbf{S}^{(i,j)}_{1,\,b} \gets {Split}\Big(\mathbf{S}_1^{(i,\,j)}\Big),
\end{equation}
where $\mathbf{S}_{1,\,f}^{(i,\,j)}$ and $\mathbf{S}_{1,\,b}^{(i,\,j)}$ are arranged as:
\begin{subequations}
    \begin{align}
        \mathbf{S}^{(i,j)}_{1,\,f} &= \Big[\underbrace{\textcolor{green}{\mathbf{x}^{(i-2,j-2)}}, \mathbf{x}^{(i-1,j-2)}, \mathbf{x}^{(i,j-2)},\cdots, \textcolor{orange}{\mathbf{x}^{(i,j)}}}_{{length} = (p^2 + 1) / 2}\Big]^\top, \label{eq:s1f}\\
        \mathbf{S}^{(i,j)}_{1,\,b} &=\Big[\underbrace{\textcolor{blue}{\mathbf{x}^{(i+2,j+2)}}, \mathbf{x}^{(i+1,j+2)}, \mathbf{x}^{(i,j+2)},\cdots, \textcolor{orange}{\mathbf{x}^{(i,j)}}}_{{length} = (p^2 + 1) / 2}\Big]^\top,\label{eq:s1b} 
    \end{align}
\end{subequations}
where two sub-sequences termed as forward-direction and backward-direction, both include the center step $\mathbf{x}^{(i,j)}$ of the complete sequence and assign it at the last step in the sub-sequence. In this case, when patch size $p$ equals 5, the length of sub-sequence is 13, as shown in Fig. \ref{fig:four types of cross mamba scan}(a).

Consequently, take $\mathbf{S}^{(i,j)}_{1,\,f}$ as an example, we can feed it into Mamba model respectively, referring Eqs. \ref{eq:s41} and \ref{eq:s42} as:
\begin{subequations}
    \begin{align}
    \mathbf{h}^{(i,j)}_{1,\,f}[t] & = \mathbf{\overline{A}}\, \mathbf{h}^{(i,j)}_{1,\,f}[t-1] + \mathbf{\overline{B}}\, \mathbf{S}^{(i,j)}_{1,\,f}[t], \\
    \mathbf{y}^{(i,j)}_{1,\,f}[t] & = \mathbf{C}\, \mathbf{h}^{(i,j)}_{1,\,f}[t] + \mathbf{D}\, \mathbf{S}^{(i,j)}_{1,\,f}[t], 
\end{align}
\end{subequations}
where $t\in\mathbb{Z}$, and $0 \leq t \leq \frac{p^2+1}{2}-1$, denoting the element index (step) in the sub-sequence. Therefore, the last step, $\mathbf{x}^{(i,j)}$, center of the patch, can iterate all previous steps' features. By doing the same operation for the another sub-sequence, $\mathbf{S}^{(i,j)}_{1,\,b}$, the whole image patch's feature can be merged by two of them.
\begin{equation}
    \mathbf{y}^{(i,j)}_1 = {Merge} \Big(\mathbf{y}^{(i,j)}_{1,\,f},\, \mathbf{y}^{(i,j)}_{1,\,b}\Big),
\end{equation}
where $\mathbf{y}^{(i,j)}_{1,\,f}$ and $\mathbf{y}^{(i,j)}_{1,\,b}$ are the output sequence for two sub-sequences. And $\mathbf{y}^{(i,j)}_{1}$ is the merged feature from type-1 MCS for representing whole image feature. The ${Merge}()$ operation will be detailed in the next section.

\subsubsection{\textbf{MCS Type-2}}

Hence, following the same description of type-1 MCS, the type-2 MCS with vertical manner, referring Fig. \ref{fig:four types of cross mamba scan}(b), can be formulated as:

\begin{subequations}
    \begin{align}
        \mathbf{S}^{(i,j)}_{2,\,f} &= \Big[\textcolor{green}{\mathbf{x}^{(i-2,j-2)}}, \mathbf{x}^{(i-2,j-1)}, \mathbf{x}^{(i-2,j)},\cdots, \textcolor{orange}{\mathbf{x}^{(i,j)}}\Big]^\top,\\
        \mathbf{S}^{(i,j)}_{2,\,b} &= \Big[\textcolor{blue}{\mathbf{x}^{(i+2,j+2)}}, \mathbf{x}^{(i+2,j+1)}, \mathbf{x}^{(i+2,j)},\cdots, \textcolor{orange}{\mathbf{x}^{(i,j)}}\Big]^\top. 
    \end{align}
\end{subequations}

In this case, the two sub-sequences $\mathbf{S}^{(i,j)}_{2,\,f}$ and $\mathbf{S}^{(i,j)}_{2,\,b}$ starts at $\mathbf{x}^{(i-2,\, j-2)}$ and $\mathbf{x}^{(i+2,\, j+2)}$, respectively and ends at $\mathbf{x}^{(i,j)}$ both with a vertical scanning manner. The length of them is same with the previous description.

\subsubsection{\textbf{MCS Type-3}}

Different from the type-1 and type-2 which starts from two positions of $\mathbf{x}^{(i-2,\, j-2)}$ and $\mathbf{x}^{(i+2,\, j+2)}$, the type-3 cross Mamba scan is starting from $\mathbf{x}^{(i-2,\, j+2)}$ and $\mathbf{x}^{(i-2,\, j+2)}$ with a horizontal manner, shown in Fig. \ref{fig:four types of cross mamba scan}(c). The formulated two sub-sequences are listed below:
\begin{subequations}
    \begin{align}
        \mathbf{S}^{(i,j)}_{3,\,f} &= \Big[\textcolor{green}{\mathbf{x}^{(i-2,j+2)}}, \mathbf{x}^{(i-1,j+2)}, \mathbf{x}^{(i,j+2)},\cdots, \textcolor{orange}{\mathbf{x}^{(i,j)}}\Big]^\top,\\
        \mathbf{S}^{(i,j)}_{3,\,b} &=\Big[\textcolor{blue}{\mathbf{x}^{(i+2,j-2)}}, \mathbf{x}^{(i+1,j-2)}, \mathbf{x}^{(i,j-2)},\cdots, \textcolor{orange}{\mathbf{x}^{(i,j)}}\Big]^\top.
    \end{align}
\end{subequations}

\subsubsection{\textbf{MCS Type-4}}
Similar with the type-3 MCS, the type-4 MCS starts from $\mathbf{x}^{(i-2,\, j-2)}$ and $\mathbf{x}^{(i+2,\, j+2)}$ positions with a vertical manner, shown in Fig. \ref{fig:four types of cross mamba scan}(d). Two sub-sequences are formulated below:
\begin{subequations}
    \begin{align}
        \mathbf{S}^{(i,j)}_{4,\,f} &= \Big[\textcolor{green}{\mathbf{x}^{(i-2,j+2)}}, \mathbf{x}^{(i-2,j+1)}, \mathbf{x}^{(i-2,j)},\cdots, \textcolor{orange}{\mathbf{x}^{(i,j)}}\Big]^\top, \label{eq:s4b}\\
        \mathbf{S}^{(i,j)}_{4,\,b} &= \Big[\textcolor{blue}{\mathbf{x}^{(i+2,j+2)}}, \mathbf{x}^{(i+2,j+1)}, \mathbf{x}^{(i+2,j)},\cdots, \textcolor{orange}{\mathbf{x}^{(i,j)}}\Big]^\top, 
    \end{align}
\end{subequations}

Consequently, we have generate four types of MCS with totally eight sub-sequences. Each type of them will be input for the following proposed T-Mamba encoder with bi-directional design for feature generation.

Different from the other works, our proposed MCS is designed as a centralized manner with the \textit{snake} scanning method, which is more suitable for patch-wise HSI classifier where the center pixel's label is assigned for the patch for training. Moreover, the lightweight design of using two sub-sequences for representing whole image patch can largely save computational burden and complexity than other works which use two complete sequences for whole image interpretation.

\begin{figure*}[t]
    \centering
    \includegraphics[scale=0.35]{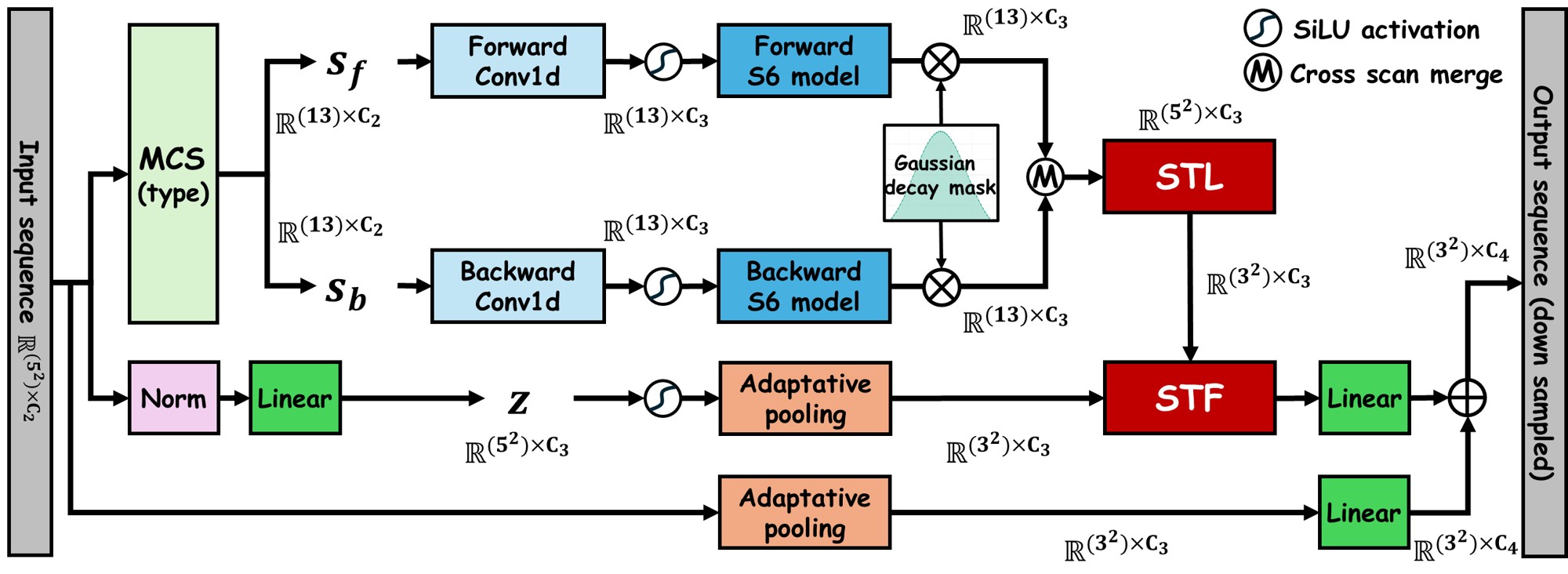}
    \caption{The T-Mamba encoder. Take inputs of size $\mathbb{R}^{(5^2) \times C_2}$ as an example. Initially, the input is split into forward and backward sequences $\mathbf{s}_f$ and $\mathbf{s}_b$ by MCS, each processed independently through a Conv1d and an S6 model. A Gaussian decay mask then statically adjusts the features of each sequence. Two processed feature sequences are then merged for feature aggregation. A downsample-driven Semantic Token Learner (STL) subsequently generates representative semantic tokens, which are correlated with $\mathbf{z}$ by an attention-driven Semantic Token Fuser (STF) to implement feature enhancement. Finally, a residual connection from the original input is used to enhance feature integration for output to the next stage.
}
    \label{fig:T_mamba}
\end{figure*}

\subsection{\textbf{Tokenized Mamba (T-Mamba)}}
After obtaining two sub-sequences from each type of MCS, we introduce the novel T-Mamba encoder, which processes them in a bi-directional manner. An overview of the T-Mamba is shown in Fig. \ref{fig:T_mamba}.

\subsubsection{\textbf{Whole Flowchart of T-Mamba}}
Taking Eq. \ref{eq:5} as an example, for a type-1 MCS, we start with an embedded sequence $\mathbf{S}_1^{(i,\,j)}\in\mathbb{R}^{5^2 \times \mathcal{C}_2}$. This sequence is first normalized and then projected into a sequence as follows:
\begin{equation}
    \mathbf{z} = {Linear}\Big({Norm}\Big(\mathbf{S}_1^{(i,\,j)}\Big)\Big),
\end{equation}
where $\mathbf{z}$ is of size $\mathbb{R}^{5^2 \times \mathcal{C}_3}$. 

Next, $\mathbf{S}_1^{(i,\,j)}$ is split into two directional sub-sequences, $\mathbf{s}_f$ and $\mathbf{s}_b$ by MCS, where $f$ denotes the forward sub-sequence, and $b$ denotes the backward sub-sequence with each sub-sequence having dimensions $\mathbb{R}^{13 \times \mathcal{C}_3}$, referring Eq. \ref{eq:split1}:
\begin{equation}\label{eq:split f b}
    \mathbf{s}_f,\, \mathbf{s}_b \gets {MCS_1}\Big(\mathbf{S}_1^{(i,\,j)}\Big),
\end{equation}
where $MCS_1()$ denotes the split operation under type-1 MCS.

Both $\mathbf{s}_f$ and $\mathbf{s}_b$ are then processed through a $SiLU()$ activation function\footnote{\url{https://pytorch.org/docs/stable/generated/torch.nn.SiLU.html}} and a 1D convolution to prepare the features:
\begin{subequations}
    \begin{align}
        \overline{\mathbf{s}_f} &= {SiLU}\Big({Conv1d}_f\Big(\mathbf{s}_f\Big)\Big), \label{eq:conv1d f}\\
        \overline{\mathbf{s}_b} &= {SiLU}\Big({Conv1d}_b\Big(\mathbf{s}_b\Big)\Big).
    \end{align}
\end{subequations}

Subsequently, these processed sub-sequences are input to the Mamba (S6) model as described in Section. \ref{sec:preliminary}:
\begin{subequations}
    \begin{align}
        \widetilde{\mathbf{s}_f} &= {Mamba}_f\Big(\overline{\mathbf{s}_f}\Big),\\
        \widetilde{\mathbf{s}_b} &= {Mamba}_b\Big(\overline{\mathbf{s}_b}\Big).\label{eq:mamba b}
    \end{align}
\end{subequations}

A Gaussian decay mask (GDM) is then applied to softly select features from the outputs of the Mamba models:
\begin{subequations}
    \begin{align}
        \widehat{\mathbf{s}_f} &= \widetilde{\mathbf{s}_f} \otimes {GDM}_f\Big(\widetilde{\mathbf{s}_f}\Big),\label{eq:GDM f}\\
        \widehat{\mathbf{s}_b} &= \widetilde{\mathbf{s}_b} \otimes {GDM}_b\Big(\widetilde{\mathbf{s}_b}\Big),\label{eq:GDM b}
    \end{align}
\end{subequations}
where ${GDM}()$ applies the Gaussian decay mask and will be detailed later, and $\otimes$ denotes the Hadamard multiplication.

The resultant features, $\widehat{\mathbf{s}_f}$ and $\widehat{\mathbf{s}_b}$, are merged to form a complete feature sequence:
\begin{equation}\label{eq:merge1}
    \widehat{\mathbf{s}} = {Merge} \Big(\widehat{\mathbf{s}_f},\, \widehat{\mathbf{s}_b}\Big),
\end{equation}
where $\widehat{\mathbf{s}}\in\mathbb{R}^{5^2 \times C_3}$ and the ${Merge}()$ operation details will be provided later.

The merged sequence $\widehat{\mathbf{s}}$ is then processed by a Semantic Token Learner (STL) module to automatically learn representative semantic tokens and perform down-sampling:
\begin{equation}\label{eq:stl}
    \mathbf{u} = {STL}\left(\widehat{\mathbf{s}}\right),
\end{equation}
where $\mathbf{u}\in\mathbb{R}^{3^2 \times C_3}$. It is crucial to maintain an odd spatial patch size to keep a center pixel.

These down-sampled semantic tokens $\mathbf{u}$ are then multiplied with the previously prepared $\mathbf{z}$ for analogous attention application:
\begin{equation}\label{eq:stf}
    \widehat{\mathbf{u}} = STF\Big(\mathbf{u},\, {AdaptivePool}\Big({SiLU}\Big(\mathbf{z}\Big)\Big)\Big),
\end{equation}
where $\widehat{\mathbf{u}}\in\mathbb{R}^{3^2 \times C_3}$. ${AdaptivePool}()$ is an adaptive pooling operation that reduces an odd patch size to a smaller odd patch size. $STF()$ represents the operation in Semantic Token Fuser (STF), detailed later.

Finally, the residual connection from the original input is added as follows:
\begin{equation}\label{eq:output current}
    \mathbf{o}^{(i,\,j)}_1 = {Linear}\Big(\widehat{\mathbf{u}}\Big) + {Linear}\Big({AdaptivePool}\Big(\mathbf{S}_1^{(i,\,j)}\Big)\Big).
\end{equation}

These operations at type-1 MCS results in semantic tokens representing the entire image patch, as illustrated in Fig. \ref{fig:T_mamba}.

\subsubsection{\textbf{Gaussian Decay Mask (GDM)}}
To well distribute the influence of features from each step in the sequence, it is essential to apply nuanced weighting to each step’s features. Rather than relying on binary values of 0 and 1 to decide which features to retain or eliminate, we introduce the GDM as a soft mask. This GDM assigns weights based on spectral and index distances within the sequence, focusing around the central pixel. 

For instance, consider a forward feature sequence $\widetilde{\mathbf{s}_f}$ outputted by the Mamba model. It can be represented by:
\begin{equation}
    \widetilde{\mathbf{s}_f} = \Big[\underbrace{\widetilde{\mathbf{s}_f}[0],\, \widetilde{\mathbf{s}_f}[1],\, \widetilde{\mathbf{s}_f}[2],\, \cdots,\, \widetilde{\mathbf{s}_f}[T]}_{length = (p^2 + 1) / 2}\Big]^\top,
\end{equation}
where $t\in\mathbb{Z}$, $0 \leq t \leq T$, and $T = \frac{p^2+1}{2}-1$, indexing the elements (steps) in the sub-sequence. The last step of the sub-sequence represents the center pixel in the patch. Therefore, the spatial GDM is calculated by evaluating the index difference between each step and the last step. 

Specifically, the \( \sigma_{idx} \), representing the average index distance to the last index, is computed as:
\begin{equation}
    \sigma_{idx} = \frac{1}{T} \sum_{t=0}^{T} \Big|t - T\Big|
\end{equation}

The weights for each index \( t \) in the sub-sequence are then calculated using a Gaussian distribution centered at \( T \):
\begin{equation}
    w_t = \exp\left(-\frac{1}{2} \left(\frac{t - T}{\sigma_{idx}}\right)^2\right)
\end{equation}

These weights are normalized to ensure they sum to one, forming the final spatial GDM for the sub-sequence:
\begin{equation}
    \mathbf{W}_{idx}[t] = \frac{w_t}{\sum_{t=0}^{T} w_t},
\end{equation}
where index-based weights \( \mathbf{W}_{idx} \) are applied to each element in the sub-sequence $\widetilde{\mathbf{s}_f}$. 

Additionally, the feature-based GDM is calculated based on the Euclidean distance between each step (feature vector) and the last step, where the \( \sigma_{spe} \), representing the average feature distance to the last step, is computed as:
\begin{equation}
    \sigma_{fea} = \frac{1}{T} \sum_{t=0}^{T} \Big|\Big| \widetilde{\mathbf{s}_f}[t] - \widetilde{\mathbf{s}_f}[T] \Big|\Big|^2
\end{equation}

The weights are then calculated by:
\begin{equation}
    v_t = \exp\left(-\frac{1}{2} \left(\frac{\Big|\Big| \widetilde{\mathbf{s}_f}[t] - \widetilde{\mathbf{s}_f}[T] \Big|\Big|^2}{\sigma_{fea}}\right)^2\right),
\end{equation}
where $||.||^2$ represents the L2-distance (Euclidean distance).

Finally, spectral-based weights \( \mathbf{W}_{fea} \) can be obtained by:
\begin{equation}
    \mathbf{W}_{fea}[t] = \frac{v_t}{\sum_{t=0}^{T} v_t}
\end{equation}

Consequently, two kinds of GDMs are designed to softly select features from the sequence as:
\begin{equation}
    \mathbf{W}_{idx}, \mathbf{W}_{fea} \gets {GDM}(\widetilde{\mathbf{s}_f})
\end{equation}
Therefore, the Eq. \ref{eq:GDM f} can be rewritten as:
\begin{equation}
    \widehat{\mathbf{s}_f} = \widetilde{\mathbf{s}_f} \otimes {Norm}\Big(\mathbf{W}_{idx} \otimes \mathbf{W}_{fea}\Big)
\end{equation}

The same operation is applied on the $\widetilde{\mathbf{s}_b}$ as well. 

This approach may allow for a more refined control over feature selection, enhancing the model's ability to capture and emphasize relevant spatial and spectral information.

\subsubsection{\textbf{Cross Scan Merge}}
To effectively merge features from two half-directional sequences and restore them to the original sequence length, we have designed the ${Merge}()$ function. This function is crafted to seamlessly integrate two sub-sequences, honoring their temporal dynamics and optimally preparing them for subsequent processing layers. The computational steps involved in this merging process are detailed below, as referenced in Eq. \ref{eq:merge1}. 

Given two sub-sequences, $\widehat{\mathbf{s}_f}$ and $\widehat{\mathbf{s}_b}$, each comprising $T$ vectors plus a final vector in $\mathbb{R}^{\mathcal{C}_3}$, we first partition these sequences as follows:
\begin{subequations}
\begin{align}
\widehat{\mathbf{q}_f} & = \Big[\widetilde{\mathbf{s}_f}[0],\, \widetilde{\mathbf{s}_f}[1],\, \widetilde{\mathbf{s}_f}[2],\, \cdots,\, \widetilde{\mathbf{s}_f}[T-1]\Big]^\top, \\
\widehat{\mathbf{q}_b} & = \Big[\widetilde{\mathbf{s}_b}[T-1],\, \widetilde{\mathbf{s}_b}[T-2],\, \widetilde{\mathbf{s}_b}[T-3],\, \cdots,\, \widetilde{\mathbf{s}_b}[0]\Big]^\top, \\
\mathbf{c} & = \frac{\widetilde{\mathbf{s}_f}[T] + \widetilde{\mathbf{s}_b}[T]}{2},
\end{align}
\end{subequations}
where $\widehat{\mathbf{q}_f}\in\mathbb{R}^{T \times \mathcal{C}_3}$ and $\widehat{\mathbf{q}_b}\in\mathbb{R}^{T \times \mathcal{C}_3}$ represent the initial segments of the two sub-sequences, and $\mathbf{c}\in\mathbb{R}^{1 \times \mathcal{C}_3}$ denotes the averaged feature from the final steps of both sub-sequences.

The ${Merge}()$ function then concatenates these elements:
\begin{equation}
\widehat{\mathbf{s}} = Linear\Big({Concat}\Big(\widehat{\mathbf{q}_f},\, \mathbf{c},\, \widehat{\mathbf{q}_b}\Big)\Big),
\end{equation}
where $\widehat{\mathbf{s}}$ resumes the original length of sequence, $\mathbb{R}^{p^2 \times \mathcal{C}_3}$.

This approach effectively combines the two sub-sequences into a unified sequence while preserving a logical progression of features, which is particularly beneficial in models that process sequential data requiring a contextual understanding of feature evolution.

\begin{figure}[t]
    \centering
    \includegraphics[scale=0.35]{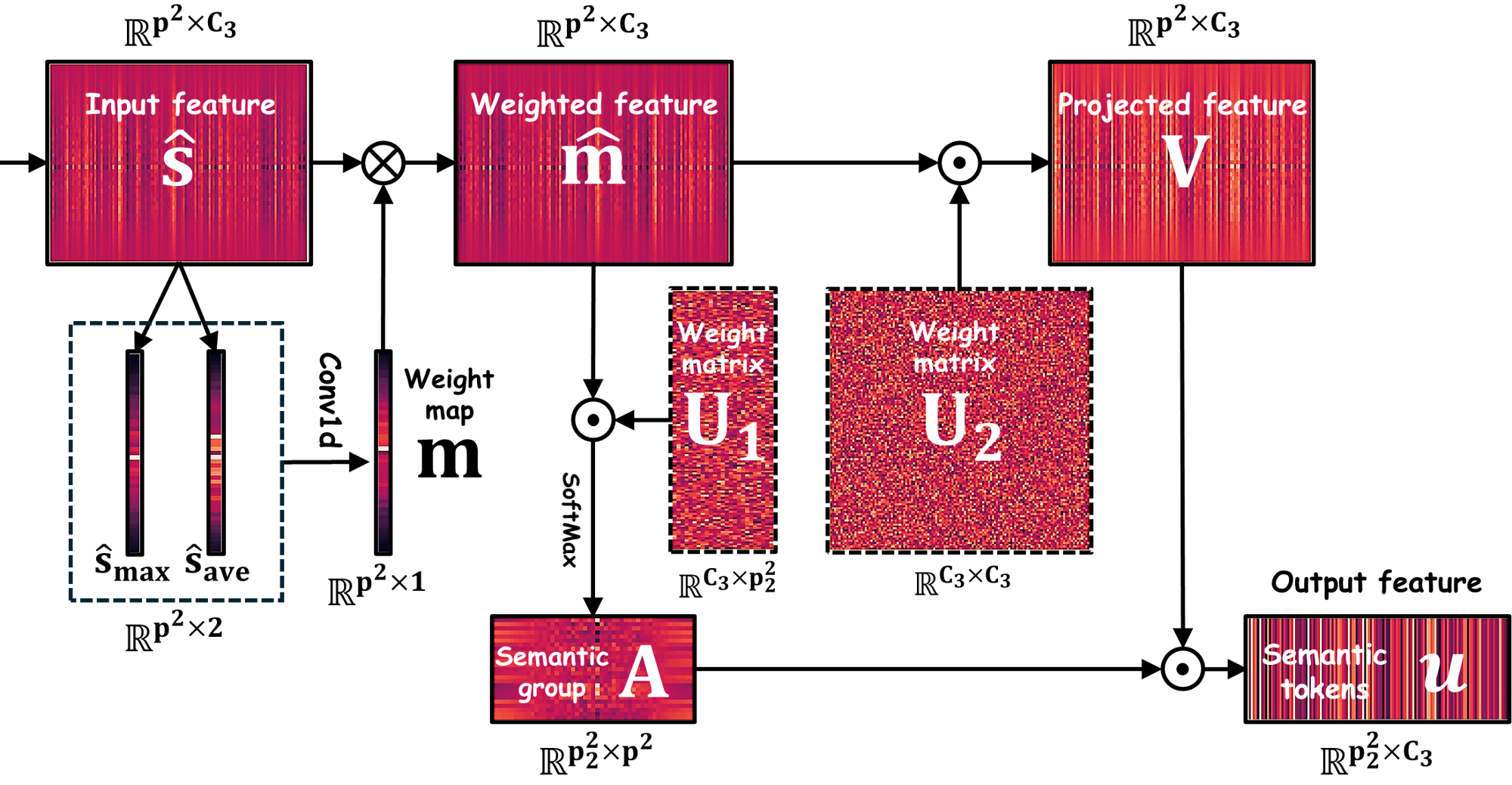}
    \caption{The visualization process of STL. The merged sequence feature is denoted by $\widehat{\mathbf{s}}$. After applying sequential attention to $\widehat{\mathbf{s}}$, the resulting features are represented as $\widehat{\mathbf{m}}$. The learnable transformation matrices $\mathbf{U}_1$ and $\mathbf{U}_2$ are utilized to extract length-level and dimension-level semantic features, denoted by $\mathbf{A}$ and $\mathbf{V}$, respectively. Finally, the down-sampled semantic tokens $\mathbf{u}$ are derived through the inner product of $\mathbf{A}$ and $\mathbf{V}$. }
    \label{fig:STL}
\end{figure}

\subsubsection{\textbf{Semantic Token Learner (STL)}}
Efficiently condensing coarse features into fine and dynamic semantic features within a sequence remains a question. Drawing inspiration from recent advancements \cite{vit8, tokenlearner}, we introduce the STL in our study. STL is tailored to abstract and condense sequence input into a set of representative semantic tokens, utilizing a sequential attention mechanism to enhance the focus and relevance of features. Fig. \ref{fig:STL} depicts an example of the transformation process facilitated by the STL.

For this process, the input feature sequence is denoted as $\widehat{\mathbf{s}} \in \mathbb{R}^{p^2 \times \mathcal{C}_3}$, where $p$ is the patch size, and $\mathcal{C}_3$ is the dimensionality.

Initially, we adopt a sequential attention module, adapted for $\widehat{\mathbf{s}}$, utilizes a 1D convolution. This module emphasizes salient features within the $\widehat{\mathbf{s}}$, allowing for a focused analysis and feature extraction without 2D spatial dimensions. 

It operates as follows:
\begin{subequations}
    \begin{align}\label{eq:seq attn1}
     \widehat{\mathbf{s}}_{max} &= max(\widehat{\mathbf{s}}),\\
     \widehat{\mathbf{s}}_{ave} &= mean(\widehat{\mathbf{s}}),
\end{align}
\end{subequations}
where $\widehat{\mathbf{s}}_{max} \in \mathbb{R}^{p^2 \times 1}$ and $\widehat{\mathbf{s}}_{ave} \in \mathbb{R}^{p^2 \times 1}$ represent the max and average pooling features along the sequence, respectively.

Then, these features are concatenated and subjected to a 1D convolution followed by a sigmoid activation $Sigmoid()$ to produce a feature weighting map:
\begin{equation}
    \mathbf{m} = Sigmoid\Big(Conv1d\Big(Concat\Big(\widehat{\mathbf{s}}_{max},\, \widehat{\mathbf{s}}_{ave}\Big)\Big)\Big),
\end{equation}
where $\mathbf{m} \in \mathbb{R}^{p^2 \times 1}$.

The weighting map is then applied to the original input:
\begin{equation}\label{eq:seq attn2}
\widehat{\mathbf{m}} = \mathbf{m} \otimes \widehat{\mathbf{s}},
\end{equation}
yielding $\widehat{\mathbf{m}} \in \mathbb{R}^{p^2 \times \mathcal{C}_3}$.

Subsequently, semantic tokens are initialized as $\mathbf{u} \in \mathbb{R}^{p_2^2 \times \mathcal{C}_3}$, with $p_2$ being an odd number smaller than $p$ to indicate a reduced number of semantic tokens. The transformation of the weighted feature sequence $\widehat{\mathbf{m}}$ into semantic tokens is performed as follows:
\begin{equation}
    \mathbf{u} = \underbrace{Softmax\left(\left(\widehat{\mathbf{m}} \odot \mathbf{U}_1\right)^\top\right)}_{\mathbf{A}} \odot \underbrace{\Big(\widehat{\mathbf{m}} \odot \mathbf{U}_2\Big)}_{\mathbf{V}},
\end{equation}
where $\mathbf{U}_1 \in \mathbb{R}^{\mathcal{C}_3 \times p_2^2}$ and $\mathbf{U}_2 \in \mathbb{R}^{\mathcal{C}_3 \times \mathcal{C}_3}$ are weight matrix initialized using a Gaussian distribution. $\odot$ denotes the inner-product multiplication. This configuration allows $\widehat{\mathbf{m}}$ to map into two semantic groups effectively, with $\mathbf{A} \in \mathbb{R}^{p_2^2 \times p_2^2}$ and $\mathbf{V} \in \mathbb{R}^{p^2 \times \mathcal{C}_3}$ representing the resultant matrices in length-level and dimension-level which are then emphasized by a $Softmax()$ function for the most relevant semantic components. Finally, $\mathbf{u}$, the semantic tokens, is derived by inner-product from $\mathbf{A}$ and $\mathbf{V}$.

This STL aids in downsampling and captures crucial high-level features for downstream tasks.

\begin{figure}[t]
    \centering
    \includegraphics[scale=0.35]{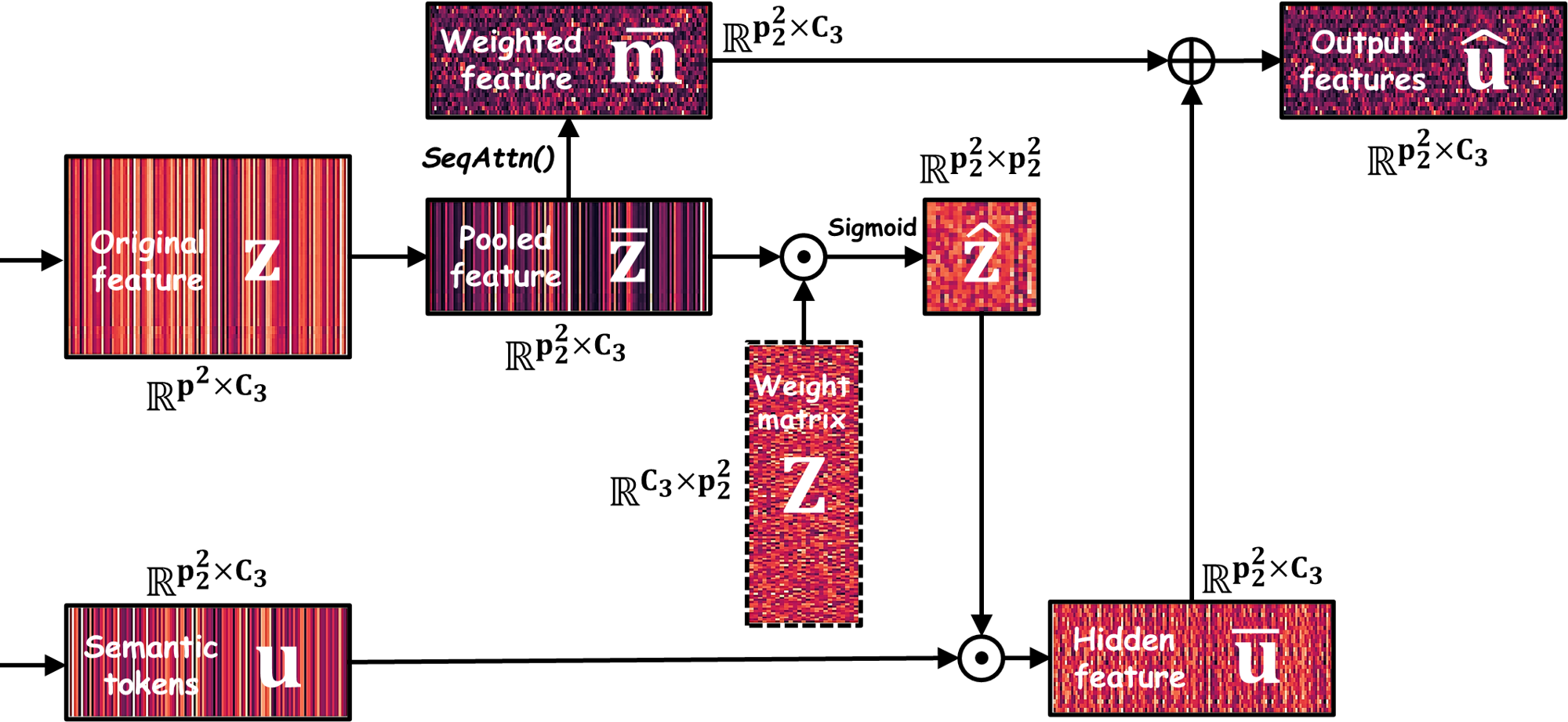}
    \caption{The visualization process of STF. It illustrates the semantic feature fusion of $\mathbf{z}$ and $\mathbf{u}$. To ensure compatibility in size with $\mathbf{u}$, $\mathbf{z}$ is first adaptively pooled into $\overline{\mathbf{z}}$. Subsequently, the pooled feature $\overline{\mathbf{z}}$ is processed through a learnable transformation matrix $\mathbf{Z}$ to produce $\widehat{\mathbf{z}}$, which captures the influence score. This allows the correlation between $\overline{\mathbf{z}}$ and $\mathbf{u}$ to be assessed through the inner product of $\widehat{\mathbf{z}}$ and the semantic token $\mathbf{u}$. Finally, to integrate the token-influenced data and the attention-modified data, $\overline{\mathbf{u}}$ and $\overline{\mathbf{m}}$ are summed to yield the final output $\widehat{\mathbf{u}}$.}
    \label{fig:STF}
\end{figure}

\subsubsection{\textbf{Semantic Token Fuser (STF)}}
To gate with the original input, previously prepared the $\mathbf{z}$ serves as the residual information for incorporating with encoded semantic features $\mathbf{u}$, as shown in Fig. \ref{fig:T_mamba}. Instead of simply multiplying them together, we aim to fuse them by a STF operation to merge the information contained in a set of semantic tokens back into the initial sequence data, effectively fusing learned abstract representations with original sequence features. This is particularly useful in scenarios where you want to reinforce or adjust the sequence data using higher-level learned representations, which could be more informative or refined than raw features. Fig. \ref{fig:STF} illustrates a sample of STF transformation process.

Given two features $\mathbf{u}\in \mathbb{R}^{p_2^2 \times \mathcal{C}_3}$ and $\mathbf{z}\in \mathbb{R}^{p^2 \times \mathcal{C}_3}$, we first pool the $\mathbf{z}$ into the same sequential length with $\mathbf{u}$:
\begin{equation}
    \overline{\mathbf{z}} = {AdaptivePool}\Big({SiLU}\Big(\mathbf{z}\Big)\Big).
\end{equation}

After that, we compute the influence scores for each token based on the features in $\overline{\mathbf{z}}\in \mathbb{R}^{p_2^2 \times \mathcal{C}_3}$:
\begin{equation}
        \widehat{\mathbf{z}} = Sigmoid\Big(\overline{\mathbf{z}}\odot\mathbf{Z}\Big)
\end{equation}
where $\widehat{\mathbf{z}}\in \mathbb{R}^{p_2^2 \times p_2^2}$ and $\mathbf{Z}\in \mathbb{R}^{\mathcal{C}_3 \times p_2^2}$ is a relevant parameter matrix. $Sigmoid()$ function ensures the scores range between 0 and 1.
    
Then, the tokens $\mathbf{u}$ are weighted by the influence scores $\widehat{\mathbf{z}}$ and summed over the tokens to blend the token information into the sequence:
\begin{equation}
        \overline{\mathbf{u}} = \widehat{\mathbf{z}} \odot \mathbf{u}, 
\end{equation}
where $\overline{\mathbf{u}}\in \mathbb{R}^{p_2^2 \times \mathcal{C}_3}$, effectively spreading the token features across the sequence dimensions.

Next, the sequential attention $SeqAttn()$, mentioned in Eqs. \ref{eq:seq attn1}-\ref{eq:seq attn2} is re-applied to $\overline{\mathbf{z}}$ to emphasize significant features:
\begin{equation}
        \overline{\mathbf{m}} = SeqAttn(\overline{\mathbf{z}}),
\end{equation}
where $\overline{\mathbf{m}} \in \mathbb{R}^{p_2^2 \times \mathcal{C}_3}$ represents the attention-weighted sequence features.

Finally, The final sequence output integrates both the token-influenced data and the attention-modified data:
\begin{equation}
        \widehat{\mathbf{u}} = \overline{\mathbf{u}} + \overline{\mathbf{m}},
\end{equation}
where $\widehat{\mathbf{u}}\in \mathbb{R}^{p_2^2 \times \mathcal{C}_3}$, providing a rich, detail-enhanced sequence for downstream tasks.

These operations ensure that the fused feature $\widehat{\mathbf{u}}$ is well-suited for downstream tasks, capturing both the detailed and global contextual information necessary for effective model performance. The addition of the attention mechanism particularly enhances the model's ability to focus on relevant features dynamically based on the given input context.

\begin{algorithm}[t]
\caption{Summary of procedure for the proposed method}
\label{alg:proposed_method}
\begin{algorithmic}[1]
\STATE \textbf{Input:} Input HSI data $\mathbf{X} \in \mathbb{R}^{\mathcal{H} \times \mathcal{W} \times \mathcal{C}_0}$, ground truth $\mathbf{L} \in \mathbb{R}^{\mathcal{H} \times \mathcal{W}}$; patch size $p$; fixed training samples.
\STATE \textbf{Output:} Predicted labels of the testing samples.
\STATE \textbf{Initialization:} Optimizer: AdamW; Learning rate: 0.0005; Criterion: CrossEntropy; Epochs: 300; Batch size: 64.
\STATE \textbf{Procedure:}
\STATE Apply principal component analysis (PCA) on $\mathbf{X} \in \mathbb{R}^{\mathcal{H} \times \mathcal{W} \times \mathcal{C}}$ for dimension reduction, getting $\mathcal{C}_1$.  
\STATE Crop patch $\textbf{X}^{(i,\,j)}$ centered at pixel $\textbf{x}^{(i,\,j)}$ from samples.
\STATE Perform depth-wise convolution on $\textbf{X}^{(i,\,j)}$ from feature dimension $\mathcal{C}_1$ to $\mathcal{C}_2$, generating local features. 
\STATE Perform point-wise convolution on $\textbf{X}^{(i,\,j)}$ then, serving as dynamic positional embedding. 

\STATE \textbf{MiM model:}
\STATE Apply four types of MCS on $\textbf{X}^{(i,\,j)}$, getting sequences, $\mathbf{S}_{1}^{(i,\,j)}$,  $\mathbf{S}_{2}^{(i,\,j)}$, $\mathbf{S}_{3}^{(i,\,j)}$, and $\mathbf{S}_{4}^{(i,\,j)}$ (Eqs. \ref{eq:s1f}-\ref{eq:s4b}).
    \FOR{$m = 1$ \TO $4$ in $\mathbf{S}_m^{(i,\,j)}$}
    \STATE Split $\mathbf{S}_m^{(i,\,j)}$ into sub-sequences, $\mathbf{s}_f$ and $\mathbf{s}_b$ (Eq. \ref{eq:split f b}).
        \STATE \textbf{T-Mamba encoder:}
            \FOR{Both forward pass $\mathbf{s}_f$ and backward pass $\mathbf{s}_b$}
                \STATE Feed sub-sequences $\mathbf{s}_f$, $\mathbf{s}_b$ into respective Conv1d and S6 model, obtaining $\widetilde{\mathbf{s}_f}$ and $\widetilde{\mathbf{s}_b}$ (Eqs. \ref{eq:conv1d f}-\ref{eq:mamba b}).
                \STATE Apply GDMs to $\widetilde{\mathbf{s}_f}$ and $\widetilde{\mathbf{s}_b}$, resulting in $\widehat{\mathbf{s}_f}$ and $\widehat{\mathbf{s}_b}$ (Eqs. \ref{eq:GDM f} and \ref{eq:GDM b}).
                \STATE Merge two features $\widehat{\mathbf{s}_f}$ and $\widehat{\mathbf{s}_b}$, getting $\widehat{\mathbf{s}}$ (Eq. \ref{eq:merge1}).
                \STATE Deploy STL on $\widehat{\mathbf{s}}$ to generate down-sampled semantic tokens $\mathbf{u}$ (Eq. \ref{eq:stl}).
                \STATE Apply STF on $\mathbf{u}$ and original feature $\mathbf{z}$ to generate enhanced tokens $\mathbf{\widehat{u}}$ (Eq. \ref{eq:stf}).
                \STATE Apply a residual connection between $\mathbf{\widehat{u}}$ and pooled $\mathbf{S}_m^{(i,\,j)}$, producing current output $\mathbf{o}_m^{(i,\,j)}$ (Eq. \ref{eq:output current}).
                \STATE Activate $\mathbf{o}_m^{(i,\,j)}$ using the Tanh function.
            \ENDFOR
    \ENDFOR
\STATE Merge four outputs, $\mathbf{o}_1^{(i,\,j)}$, $\mathbf{o}_2^{(i,\,j)}$, $\mathbf{o}_3^{(i,\,j)}$, and $\mathbf{o}_4^{(i,\,j)}$, by WMF for weighted fusion, recorded as current scale ($p$) feature $\mathbf{O}_p^{(i,\,j)}$ (Eq. \ref{eq:weight fusion 1}).
\STATE Iterate MiM models on $\mathbf{o}_1^{(i,\,j)}$, $\mathbf{o}_2^{(i,\,j)}$, $\mathbf{o}_3^{(i,\,j)}$, and $\mathbf{o}_4^{(i,\,j)}$, until the patch size $p$ is reduced to $1$ (Eq. \ref{eq:mim iteration}).
\STATE Train the model using a decoder (MLP) and a multi-scale loss function (Eq. \ref{eq:multi-scale loss}).
\end{algorithmic}
\end{algorithm}

\begin{sidewaystable}[t]
\caption{Description of land-cover types and number of train-test samples for four datasets\label{tab:traning and testing samples}}
\tiny
\centering
\begin{tabular}{|c|c||ccc||ccc||ccc||ccc|}
\hline
 & &\multicolumn{3}{c||}{\textbf{Indian Pines (IP)}} & \multicolumn{3}{c||}{\textbf{Houston 2013 (HU2013))}} & \multicolumn{3}{c||}{\textbf{Pavia University (PU)}} & \multicolumn{3}{c|}{\textbf{WHU-Hi-HongHu (HongHu)}}\\ \hline
No. & Color &\multicolumn{1}{c|}{Class Name}  & \multicolumn{1}{c|}{Train} & Test & \multicolumn{1}{c|}{Class Name}  & \multicolumn{1}{c|}{Train} & Test & \multicolumn{1}{c|}{Class Name} & \multicolumn{1}{c|}{Train} & Test & \multicolumn{1}{c|}{Class Name} & \multicolumn{1}{c|}{Train} & Test\\ \hline

1 & \cellcolor{HH-1} &\multicolumn{1}{c|}{Corn-n.}  & \multicolumn{1}{c|}{50} & 1,384 & \multicolumn{1}{c|}{Grass-h.}  & \multicolumn{1}{c|}{198} & 1,053 & \multicolumn{1}{c|}{Asphalt}  & \multicolumn{1}{c|}{548} & 6,304 & \multicolumn{1}{c|}{Red-r.}  & \multicolumn{1}{c|}{25} & 14,016\\ \hline

2 & \cellcolor{HH-2}&\multicolumn{1}{c|}{Corn-m.}  & \multicolumn{1}{c|}{50} & 784 & \multicolumn{1}{c|}{Grass-s.}  & \multicolumn{1}{c|}{190} & 1,064 & \multicolumn{1}{c|}{Meadow}  & \multicolumn{1}{c|}{540} & 18,146 & \multicolumn{1}{c|}{Road}  & \multicolumn{1}{c|}{25} & 3,487 \\ \hline

3 & \cellcolor{HH-3}&\multicolumn{1}{c|}{Corn}  & \multicolumn{1}{c|}{50} & 184 & \multicolumn{1}{c|}{Grass-s.}  & \multicolumn{1}{c|}{192} & 505 & \multicolumn{1}{c|}{Gravel}  & \multicolumn{1}{c|}{392} & 1,815 & \multicolumn{1}{c|}{Bare-s.}  & \multicolumn{1}{c|}{25} & 21,796\\ \hline

4 & \cellcolor{HH-4}&\multicolumn{1}{c|}{Grass-p.} & \multicolumn{1}{c|}{50} & 447 & \multicolumn{1}{c|}{Tree}  & \multicolumn{1}{c|}{188} & 1,056 & \multicolumn{1}{c|}{Tree}  & \multicolumn{1}{c|}{524} & 2,912 & \multicolumn{1}{c|}{Cotton}  & \multicolumn{1}{c|}{25} & 163,260\\ \hline

5 & \cellcolor{HH-5}&\multicolumn{1}{c|}{Grass-t.}  & \multicolumn{1}{c|}{50} & 697 & \multicolumn{1}{c|}{Soil}  & \multicolumn{1}{c|}{186} & 1,056 & \multicolumn{1}{c|}{Metal-s.}  & \multicolumn{1}{c|}{265} & 1,113 & \multicolumn{1}{c|}{Cotton-f.}  & \multicolumn{1}{c|}{25} & 6,193\\ \hline

6 & \cellcolor{HH-6}&\multicolumn{1}{c|}{Hay-w.}  & \multicolumn{1}{c|}{50} & 439 & \multicolumn{1}{c|}{Water}  & \multicolumn{1}{c|}{182} & 143 & \multicolumn{1}{c|}{Bare-s.}  & \multicolumn{1}{c|}{532} & 4,572 & \multicolumn{1}{c|}{Rape}  & \multicolumn{1}{c|}{25} & 44,532\\ \hline

7 & \cellcolor{HH-7}&\multicolumn{1}{c|}{Soybean-n.}  & \multicolumn{1}{c|}{50} & 918 & \multicolumn{1}{c|}{Residential}  & \multicolumn{1}{c|}{196} & 1,072 & \multicolumn{1}{c|}{Bitumen} & \multicolumn{1}{c|}{375} & 981 & \multicolumn{1}{c|}{Chinese-c.}  & \multicolumn{1}{c|}{25} & 24,078\\ \hline

8 & \cellcolor{HH-8}&\multicolumn{1}{c|}{Soybean-m.}  & \multicolumn{1}{c|}{50} & 2,418 & \multicolumn{1}{c|}{Commercial}  & \multicolumn{1}{c|}{191} & 1,053 & \multicolumn{1}{c|}{Brick} & \multicolumn{1}{c|}{514} & 3,364 & \multicolumn{1}{c|}{Pakchoi}  & \multicolumn{1}{c|}{25} & 4,029\\ \hline

9 & \cellcolor{HH-9}&\multicolumn{1}{c|}{Soybean-c.} & \multicolumn{1}{c|}{50} & 564 & \multicolumn{1}{c|}{Road}  & \multicolumn{1}{c|}{193} & 1,059 & \multicolumn{1}{c|}{Shadow} & \multicolumn{1}{c|}{231} & 795 & \multicolumn{1}{c|}{Cabbage}  & \multicolumn{1}{c|}{25} & 10,794\\ \hline

10 & \cellcolor{HH-10}&\multicolumn{1}{c|}{Wheat}  & \multicolumn{1}{c|}{50} & 162 & \multicolumn{1}{c|}{Highway}  & \multicolumn{1}{c|}{191} & 1,036 & \multicolumn{1}{c|}{}  & \multicolumn{1}{c|}{} &  & \multicolumn{1}{c|}{Tuber-m.}  & \multicolumn{1}{c|}{25} & 12,369\\ \hline

11 & \cellcolor{HH-11}&\multicolumn{1}{c|}{Wood}  & \multicolumn{1}{c|}{50} & 1,244 & \multicolumn{1}{c|}{Railway}& \multicolumn{1}{c|}{181} & 1,054 & \multicolumn{1}{c|}{}  & \multicolumn{1}{c|}{} &  & \multicolumn{1}{c|}{Brassica-p.}  & \multicolumn{1}{c|}{25} & 10,990\\ \hline

12 & \cellcolor{HH-12}&\multicolumn{1}{c|}{Buildings-g.-t.}  & \multicolumn{1}{c|}{50} & 330 & \multicolumn{1}{c|}{Park-l-1}  & \multicolumn{1}{c|}{192} & 1,041 & \multicolumn{1}{c|}{}  & \multicolumn{1}{c|}{} &  & \multicolumn{1}{c|}{Brassica-c.}  & \multicolumn{1}{c|}{25} & 8,929\\ \hline

13 & \cellcolor{HH-13}&\multicolumn{1}{c|}{Stone-s.-t.}  & \multicolumn{1}{c|}{50} & 45 & \multicolumn{1}{c|}{Park-l-2}  & \multicolumn{1}{c|}{184} & 285 & \multicolumn{1}{c|}{}  & \multicolumn{1}{c|}{} &  & \multicolumn{1}{c|}{Small-b.-c.}  & \multicolumn{1}{c|}{25} & 22,482\\ \hline

14 & \cellcolor{HH-14}&\multicolumn{1}{c|}{Alfalfa}  & \multicolumn{1}{c|}{15} & 39 & \multicolumn{1}{c|}{Tennis-g.}  & \multicolumn{1}{c|}{181} & 247 & \multicolumn{1}{c|}{}  & \multicolumn{1}{c|}{} &  & \multicolumn{1}{c|}{Lactuca-s.}  & \multicolumn{1}{c|}{25} & 7,331\\ \hline

15 & \cellcolor{HH-15}&\multicolumn{1}{c|}{Grass-p.-m.} & \multicolumn{1}{c|}{15} & 11 & \multicolumn{1}{c|}{Running-t.}  & \multicolumn{1}{c|}{187} & 473 & \multicolumn{1}{c|}{}  & \multicolumn{1}{c|}{} &  & \multicolumn{1}{c|}{Celtuce}  & \multicolumn{1}{c|}{25} & 977\\ \hline

16 & \cellcolor{HH-16}&\multicolumn{1}{c|}{Oat}  & \multicolumn{1}{c|}{15} & 5 & \multicolumn{1}{c|}{} & \multicolumn{1}{c|}{} &  & \multicolumn{1}{c|}{}  & \multicolumn{1}{c|}{} &  & \multicolumn{1}{c|}{Film-c.-l.}  & \multicolumn{1}{c|}{25} & 7,237\\ \hline

17 & \cellcolor{HH-17}&\multicolumn{1}{c|}{}  & \multicolumn{1}{c|}{} &  & \multicolumn{1}{c|}{}  & \multicolumn{1}{c|}{} &  & \multicolumn{1}{c|}{}  & \multicolumn{1}{c|}{} &  & \multicolumn{1}{c|}{Romaine-l.}  & \multicolumn{1}{c|}{25} & 2,985\\ \hline

18 & \cellcolor{HH-18}&\multicolumn{1}{c|}{}  & \multicolumn{1}{c|}{} &  & \multicolumn{1}{c|}{}  & \multicolumn{1}{c|}{} &  & \multicolumn{1}{c|}{}  & \multicolumn{1}{c|}{} &  & \multicolumn{1}{c|}{Carrot}  & \multicolumn{1}{c|}{25} & 3,192\\ \hline

19 & \cellcolor{HH-19}&\multicolumn{1}{c|}{}  & \multicolumn{1}{c|}{} &  & \multicolumn{1}{c|}{}  & \multicolumn{1}{c|}{} &  & \multicolumn{1}{c|}{}  & \multicolumn{1}{c|}{} &  & \multicolumn{1}{c|}{White-r.}  & \multicolumn{1}{c|}{25} & 8,687\\ \hline

20 & \cellcolor{HH-20}&\multicolumn{1}{c|}{}  & \multicolumn{1}{c|}{} &  & \multicolumn{1}{c|}{}  & \multicolumn{1}{c|}{} &  & \multicolumn{1}{c|}{}  & \multicolumn{1}{c|}{} &  & \multicolumn{1}{c|}{Garlic-s.}  & \multicolumn{1}{c|}{25} & 3,461\\ \hline

21 & \cellcolor{HH-21}&\multicolumn{1}{c|}{}  & \multicolumn{1}{c|}{} &  & \multicolumn{1}{c|}{}  & \multicolumn{1}{c|}{} &  & \multicolumn{1}{c|}{}  & \multicolumn{1}{c|}{} &  & \multicolumn{1}{c|}{Broad-b.}  & \multicolumn{1}{c|}{25} & 1,303\\ \hline

22 & \cellcolor{HH-22}&\multicolumn{1}{c|}{}  & \multicolumn{1}{c|}{} &  & \multicolumn{1}{c|}{}  & \multicolumn{1}{c|}{} &  & \multicolumn{1}{c|}{}  & \multicolumn{1}{c|}{} &  & \multicolumn{1}{c|}{Tree}  & \multicolumn{1}{c|}{25} & 4,015\\ \hline

Total & &\multicolumn{1}{c|}{}  & \multicolumn{1}{c|}{695} & 9,671 & \multicolumn{1}{c|}{}  & \multicolumn{1}{c|}{2,832} & 12,197 & \multicolumn{1}{c|}{} & \multicolumn{1}{c|}{3,921} & 40,002 & \multicolumn{1}{c|}{} & \multicolumn{1}{c|}{550} & 386,143\\ \hline

Ratio & &\multicolumn{1}{c|}{}  & \multicolumn{1}{c|}{6.7\%} & 93.3\% & \multicolumn{1}{c|}{}  & \multicolumn{1}{c|}{18.8\%} & 81.2\% & \multicolumn{1}{c|}{} & \multicolumn{1}{c|}{8.9\%} & 91.1\% & \multicolumn{1}{c|}{} & \multicolumn{1}{c|}{0.14\%} & 99.86\%\\ \hline
\end{tabular}

\end{sidewaystable}

\begin{sidewaystable}[!t]
\centering
\caption{The hyperparameter settings and size of notations for four datasets.\label{tab:hyperparameter settings}}
\begin{tabular}{|c|c||c||c||c||c|}
\hline
Definition & Notation & \textbf{IP} & \textbf{HU2013} & \textbf{PU} & \textbf{HongHu} \\ \hline

Image Size: & $\mathcal{H} \times \mathcal{W} \times \mathcal{C}_0$ & $145 \times 145 \times 200$ & $349 \times 1905 \times 144$ & $610 \times 340 \times 103$ & $940 \times 475 \times 270$\\ \hline

Dimension after PCA: & $\mathcal{C}_1$ & 60 & 30 & 30 & 100 \\ \hline

Initial Patch Size: & $p \times p$ & $7 \times 7$  & $9 \times 9$ & $11 \times 11$ &$9 \times 9$\\ \hline

Feature Dimension Setting: & $\mathcal{C}_2,\, \mathcal{C}_3,\, \mathcal{C}_4$ & 64 & 64 & 32 & 128\\ \hline

Number of T-Mamba Layers: & & 4 & 4 & 2 & 3 \\ \hline

Number of MiM Layers: & & 4 & 5 & 6 & 5 \\ \hline

Learning Rate:  & & 0.0005 & 0.0005 & 0.001 & 0.001 \\ \hline

Feature Dropout: & & 0.1 & 0.1 & 0.2 & 0.1 \\ \hline

\end{tabular}

\end{sidewaystable}

\subsection{\textbf{Weighted MCS Fusion (WMF)}\label{sec:Weighted Token Fuser}}
After obtaining four tokenized features, $\mathbf{o}^{(i,\,j)}_1$, $\mathbf{o}^{(i,\,j)}_2$, $\mathbf{o}^{(i,\,j)}_3$, and $\mathbf{o}^{(i,\,j)}_4$, from different T-Mamba encoders, we recognize that each feature set varies depending on the scan order of the sequence. This variability is critical in the context of our proposed MCS, which involves multiple scan orders, each providing a unique output representation. To effectively integrate these features, a thoughtful fusion strategy is essential, as indiscriminate merging could dilute their distinct contributions.

In response, we introduce the WMF, designed to assign attention weights to each feature set according to its relevance and importance in the overall context. The fusion process is formulated as follows:
\begin{equation}\label{eq:weight fusion 1}
    \mathbf{O}^{(i,\,j)}_p = k_1\mathbf{o}^{(i,\,j)}_1 + k_2\mathbf{o}^{(i,\,j)}_2 + k_3\mathbf{o}^{(i,\,j)}_3 + k_4\mathbf{o}^{(i,\,j)}_4,
\end{equation}
where \( k_o\in \mathbb{R}^{1} \) (for \( o \in \{1, 2, 3, 4\} \)) represents the weight assigned to each encoded feature, and $k_1 + k_2 + k_3 + k_4 = 1$. $\mathbf{O}^{(i,\,j)}_p$ denotes the fused feature at scale $p$. These weights are dynamically integrated into the network architecture, allowing for optimization through learning, ensuring that each feature set contributes optimally to the final model output.

Following this feature fusion, the weighted feature $\mathbf{O}^{(i,\,j)}_p$ is processed through a decoder, specifically a Multi-Layer Perceptron (MLP), essential for refining and interpreting the fused features for model training:
\begin{equation}\label{eq:decoder}
    \mathbf{Y}^{(i,\,j)}_p = MLP\Big(Tanh\Big(\mathbf{O}^{(i,\,j)}_p\Big)\Big),
\end{equation}
where $Tanh()$ denotes the Hyperbolic Tangent activation function\footnote{\url{https://pytorch.org/docs/stable/generated/torch.nn.Tanh.html}}.

The training of the model is governed by the cross-entropy loss function, commonly used for classification tasks. This loss function measures the discrepancy between the predicted and actual labels of the training samples and is defined as:
\begin{equation}\label{eq:loss function}
    \mathcal{L}_p = -\frac{1}{N_{tr}}\sum^{N_{tr}}_{tr=1}\Big[\mathbf{L}_{tr}\log(\mathbf{Y}_p) + (1-\mathbf{L}_{tr})\log(1-\mathbf{Y}_p)\Big],
\end{equation}
where \( N_{tr} \) is the total number of training samples, \( \mathbf{L}_{tr} \) denotes the true label, and \( \mathbf{Y}_p \) the predicted label for the \( tr \)-th sample. This function effectively minimizes the error between predicted and true labels, driving the training process towards better accuracy.

\begin{sidewaystable}[t]
\centering
\caption{Different classification methods in terms of OA, AA, and Kappa as well as the accuracies for each
class on the \underline{\textbf{Indian Pines}} dataset. The \textcolor{red}{\textbf{best one}} is highlighted as red. The \textcolor{blue}{\textbf{second best}} is highlighted as blue. \label{tab:IP quant results}}

\begin{tabular}{|c|c||c|c|c|c|c|c|c|c|c|}
\hline
                 No.& Color. & vanilla ViT & SpeFormer & MAEST & HiT & SSFTT & 3DViT & HUSST & ViM & MiM (Ours)\\ \hline
                
1&\cellcolor{IP-1} & 64.480 & 70.538 & 76.721 & \textcolor{blue}{\textbf{87.948}} & 84.825 & 81.328 & 87.137 & 74.787 & \textcolor{red}{\textbf{90.291}} \\ \hline
2&\cellcolor{IP-2} & 61.374 & 65.570 & 68.970 & 80.650 & \textcolor{blue}{\textbf{84.251}} & 73.540 & 82.510 & 68.233 & \textcolor{red}{\textbf{90.269}} \\ \hline
3&\cellcolor{IP-3} & 69.919 & 86.352 & 89.600 & 82.353 & \textcolor{blue}{\textbf{93.583}} & 93.557 & 90.515 & 91.153 & \textcolor{red}{\textbf{97.082}} \\ \hline
4&\cellcolor{IP-4} & 84.032 & 90.566 & 88.602 & 82.629 & 89.825 & 89.070 & 85.414 & \textcolor{blue}{\textbf{91.815}} & \textcolor{red}{\textbf{93.537}} \\ \hline
5&\cellcolor{IP-5} & 93.333 & 96.379 & 97.787 & 98.501 & 96.591 & 96.847 & 97.951 & \textcolor{red}{\textbf{98.860}} & \textcolor{blue}{\textbf{98.722}} \\ \hline
6&\cellcolor{IP-6} & 94.348 & \textcolor{blue}{\textbf{99.773}} & 98.731 & 95.897 & 99.658 & 96.512 & 99.312 & 98.848 & \textcolor{red}{\textbf{100}} \\ \hline
7&\cellcolor{IP-7} & 69.006 & 77.595 & 79.433 & 88.548 & \textcolor{blue}{\textbf{89.398}} & 82.875 & 82.641 & 76.452 & \textcolor{red}{\textbf{91.243}} \\ \hline
8&\cellcolor{IP-8} & 71.266 & 78.037 & 85.752 & 87.729 & \textcolor{blue}{\textbf{90.015}} & 87.600 & 88.812 & 81.906 & \textcolor{red}{\textbf{93.109}} \\ \hline
9&\cellcolor{IP-9} & 68.959 & 72.697 & 76.994 & 86.124 & 82.871 & 80.681 & \textcolor{blue}{\textbf{86.925}} & 79.850 & \textcolor{red}{\textbf{88.794}} \\ \hline
10&\cellcolor{IP-10} & 99.690 & 99.690 & \textcolor{blue}{\textbf{99.962}} & 97.885 & 99.387 & \textcolor{red}{\textbf{100}} & 99.083 & \textcolor{blue}{\textbf{99.692}} & \textcolor{red}{\textbf{100}} \\ \hline
11&\cellcolor{IP-11} & 95.215 & 97.756 & 96.940 & \textcolor{blue}{\textbf{98.355}} & 96.634 & 97.438 & 96.226 & 96.870 & \textcolor{red}{\textbf{98.396}} \\ \hline
12&\cellcolor{IP-12} & 65.172 & 74.237 & 71.945 & 70.753 & 76.562 & 71.958 & \textcolor{blue}{\textbf{76.923}} & 73.851 & \textcolor{red}{\textbf{81.198}} \\ \hline
13&\cellcolor{IP-13} & 66.176 & \textcolor{blue}{\textbf{92.784}} & \textcolor{red}{\textbf{93.750}} & 84.906 & 88.235 & 90.909 & 90.909 & 90.000 & 84.906 \\ \hline
14&\cellcolor{IP-14} & 60.000 & \textcolor{blue}{\textbf{97.500}} & 78.000 & 90.698 & \textcolor{red}{\textbf{98.734}} & 69.474 & 95.000 & 88.636 & \textcolor{blue}{\textbf{97.500}} \\ \hline
15&\cellcolor{IP-15} & 60.606 & 71.667 & 52.381 & \textcolor{red}{\textbf{84.615}} & \textcolor{blue}{\textbf{78.571}} & 64.286 & 73.333 & 64.706 & 75.862 \\ \hline
16&\cellcolor{IP-16} & 52.632 & 35.714 & 35.714 & \textcolor{red}{\textbf{83.333}} & 58.824 & 35.714 & 27.778 & 33.333 & \textcolor{blue}{\textbf{76.923}} \\ \hline
OA& & 74.0875 & 79.9607 & 83.0731 & 86.0511 & \textcolor{blue}{\textbf{88.7498}} & 85.1411 & 87.3333 & 81.9667 & \textcolor{red}{\textbf{92.0794}} \\ \hline
AA& & 77.6380 & 82.2704 & 84.7051 & 89.5578 & \textcolor{blue}{\textbf{90.9978}} & 87.9868 & 89.0293 & 84.8289 & \textcolor{red}{\textbf{94.1145}} \\ \hline
Kappa& & 0.7079 & 0.7733 & 0.8074 & 0.8420 & \textcolor{blue}{\textbf{0.8715}} & 0.8305 & 0.8556 & 0.7952 & \textcolor{red}{\textbf{0.9096}} \\ \hline
\end{tabular}
\end{sidewaystable}

\begin{sidewaystable}[t]
\caption{Different classification methods in terms of OA, AA, and Kappa as well as the accuracies for each
class on the \underline{\textbf{Pavia University}} dataset. The \textcolor{red}{\textbf{best one}} is highlighted as red. The \textcolor{blue}{\textbf{second best}} is highlighted as blue. \label{tab:PU quant results}}

\centering
\begin{tabular}{|c|c||c|c|c|c|c|c|c|c|c|}
\hline
                 No.& Color. & vanilla ViT & SpeFormer & MAEST & HiT & SSFTT & 3DViT & HUSST & ViM & MiM (Ours)\\ \hline
                
1&\cellcolor{PU-1} & 87.245 & 89.487 & 90.084 & 90.559 & \textcolor{blue}{\textbf{96.721}} & 95.985 & 96.443 & 78.878 & \textcolor{red}{\textbf{97.255}} \\ \hline
2&\cellcolor{PU-2} & 74.871 & 90.862 & \textcolor{blue}{\textbf{92.186}} & 91.322 & 89.994 & 88.007 & 87.622 & 85.758 & \textcolor{red}{\textbf{92.947}} \\ \hline
3&\cellcolor{PU-3} & 60.796 & 78.886 & 79.862 & 66.773 & 83.342 & \textcolor{red}{\textbf{88.233}} & 75.083 & 46.655 & \textcolor{blue}{\textbf{87.009}} \\ \hline
4&\cellcolor{PU-4} & 76.601 & 88.610 & 90.488 & 94.595 & 94.804 & \textcolor{red}{\textbf{97.830}} & 95.025 & 90.010 & \textcolor{blue}{\textbf{96.251}} \\ \hline
5&\cellcolor{PU-5} & 99.194 & \textcolor{red}{\textbf{100}} & \textcolor{red}{\textbf{100}}    & 98.670 & \textcolor{red}{\textbf{100}} & \textcolor{red}{\textbf{100}} & \textcolor{red}{\textbf{100}} & \textcolor{blue}{\textbf{99.326}} & \textcolor{red}{\textbf{100}} \\ \hline
6&\cellcolor{PU-6} & 64.266 & 75.571 & \textcolor{blue}{\textbf{82.845}} & \textcolor{red}{\textbf{87.232}} & 73.704 & 68.196 & 74.182 & 71.951 & 80.032 \\ \hline
7&\cellcolor{PU-7} & 95.694 & 80.966 & 89.988 & 84.953 & 98.616 & \textcolor{red}{\textbf{99.217}} & 94.418 & 77.917 & \textcolor{blue}{\textbf{99.128}} \\ \hline
8&\cellcolor{PU-8} & 82.894 & 83.960 & 84.277 & \textcolor{blue}{\textbf{90.554}} & 88.452 & 90.886 & 90.376 & 76.343 & \textcolor{red}{\textbf{91.252}} \\ \hline
9&\cellcolor{PU-9} & 94.375 & \textcolor{blue}{\textbf{99.949}} & 99.310 & 95.719 & 92.159 & 93.173 & 99.619 & 95.831 & \textcolor{red}{\textbf{100}} \\ \hline

OA&                & 76.0313 & 86.2784 & 88.0433 & 87.9183 & \textcolor{blue}{\textbf{88.8707}} & 86.5258 & 86.2483 & 78.1811 & \textcolor{red}{\textbf{91.5756}} \\ \hline
AA&                & 84.4339 & 89.8503 & 90.5291 & 90.7538 & \textcolor{blue}{\textbf{91.4693}} & 91.3783 & 90.6566 & 78.5284 & \textcolor{red}{\textbf{92.7601}} \\ \hline
Kappa&             & 0.7086 & 0.8288 & 0.8341 & 0.8449 & \textcolor{blue}{\textbf{0.8608}} & 0.8581 & 0.8572 & 0.7145 & \textcolor{red}{\textbf{0.8983}} \\ \hline
\end{tabular}

\end{sidewaystable}

\subsection{\textbf{Multi-Scale Loss Design}}
To account for the multi-scale features potentially present in an image patch for different objects, we propose integrating multiple scale features into the decoding process.

In our approach, each MiM model performs a down-sampling operation on the input image patch, which is initially processed by the STL in the T-Mamba encoder.

Upon obtaining four distinct features, $\mathbf{o}^{(i,\,j)}_1$, $\mathbf{o}^{(i,\,j)}_2$, $\mathbf{o}^{(i,\,j)}_3$, and $\mathbf{o}^{(i,\,j)}_4$, from the first MiM model at the initial patch scale $p$, each feature is activated and subsequently fed into the next MiM model for further feature tokenization and down-sampling, as shown in the Fig. \ref{fig:whole strcuture}.

This process allows us to capture multi-scale features in a cascaded and iterative manner, which can be analogous as:
\begin{equation}\label{eq:mim iteration}
\mathbf{O}^{(i,\,j)}_{p_n} = \underbrace{MiM(MiM(MiM(\cdots MiM(\mathbf{O}^{(i,,j)}_{p}))))}_{n\, times},
\end{equation}
where $p_n < p$, recording the patch size at $n$-th $MiM()$ iteration. It must be noted that all patch sizes should be an odd number. The iteration is ending at $p_n = 1$. For example, $p=11$, $p_2 = 9$, $p_3 = 7$, $p_4 = 5$, $p_5 = 3$, and $p_6 = 1$ by 6 times of operation under MiM model.

To effectively utilize these multi-scale features, we employ an output-level based decoding strategy, where each scale feature is independently integrated into the loss function. Consequently, as per Eq. \ref{eq:loss function}, we formulate the total loss function as:
\begin{equation}\label{eq:multi-scale loss}
\mathcal{L}_{total} = \frac{1}{n}(\mathcal{L}_{p} + \mathcal{L}_{p_2} + \mathcal{L}_{p_3} + \cdots + \mathcal{L}_{p_n}).
\end{equation}

The iterative process concludes when the final patch size $p_n$ is reduced to 1. Thus, our total loss function effectively incorporates the multi-scale features from a single image patch to enhance training effectiveness.

\begin{figure*}[t]
    \centering
    \includegraphics[scale=0.60]{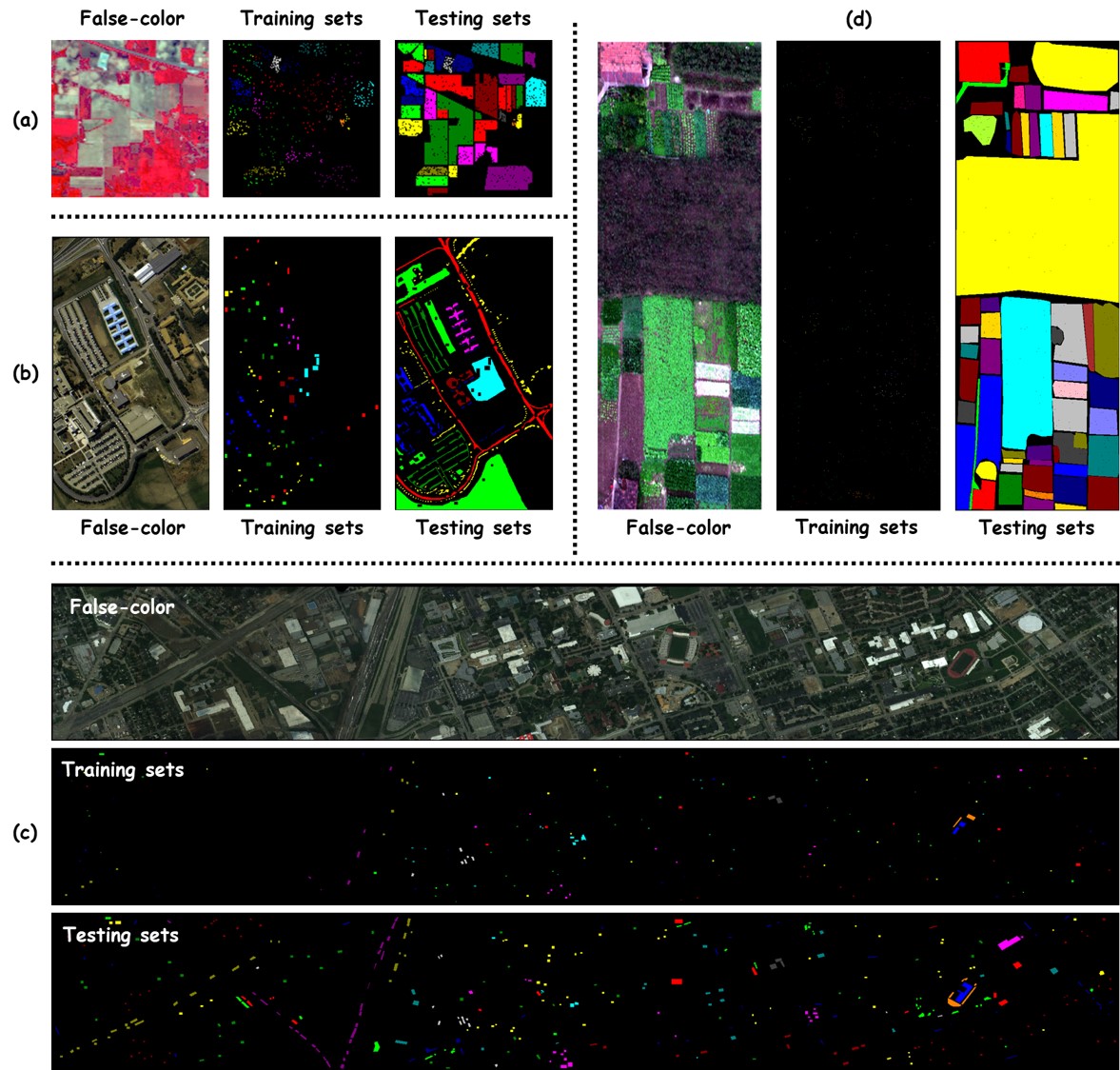}
    \caption{Four experimental datasets, (a) Indian Pines (IP), (b) Pavia University (PU), (c) Houston 2013 (HU2013), and (d) WHU-Hi-HongHu (HongHu) are used for experiments. The false-color images, fixed and disjoint training sets and testing sets are shown above.}
    \label{fig:four dataset}
\end{figure*}

\section{Experimental results and analysis \label{sec:experiment}}
In this section, we first describe four well-known HSI datasets. We then outline the implementation details and the methods used for comparison. Finally, we conduct extensive experiments to evaluate the performance of the proposed method both quantitatively and qualitatively. \textit{Due to the paper formatting, all evaluation tables are placed at end of the article.}

\subsection{\textbf{Datasets Description}}
Four HSI datasets are illustrated in Fig. \ref{fig:four dataset}, along with their corresponding false-color images, training and testing samples. The numbers of training and testing samples for each class within these datasets are listed in Table \ref{tab:traning and testing samples}. 

The Indian Pines (IP) dataset is a mixed vegetation site with $145 \times 145$ pixels and 220 spectral bands. To remove water absorption bands, the number of spectral bands was reduced to 200, resulting in 16 different land-cover classes.

The Pavia University (PU) dataset consists of $610 \times 340$ pixels and initially has 115 bands. In the experiment, 12 of the noisy bands were removed, leaving 103 bands to be used. The dataset has 9 different ground cover types.

The Houston 2013 (HU2013) dataset\footnote{\url{http://www.grss-ieee.org/community/technical-committees/data-fusion}} was used in the 2013 IEEE GRSS Data Fusion Contest. It features 144 spectral bands and spans an area of $349 \times 1905$ pixels, covering 15 different land-cover classes.

The WHU-Hi-HongHu (HongHu) dataset\footnote{\url{http://rsidea.whu.edu.cn/resource_WHUHi_sharing.htm}} was acquired by the UAV platform on November 20, 2017, in Honghu City, Hubei Province, China. The experimental area is a complex agricultural scene with many classes of crops, and different cultivars of the same crop are also planted in the region. It consists of $940 \times 475$ pixels and initially has 270 bands. This dataset has 22 different ground cover classes.

For fair comparison, all the training and testing samples are manually selected in a fixed and disjoint manner. Details can be referred to \cite{vit1} and corresponding website.

\subsection{\textbf{Experimental Evaluation Indicators}}
To quantitatively assess the effectiveness of our proposed method and compare it with other methods, we rely on four key performance indicators: overall accuracy (OA), average accuracy (AA), the kappa coefficient (kappa), and accuracy within each class. They provide a comprehensive view of the classification performance, with higher values indicating superior classification results. 

All experiments are conducted using Pytorch 2.3 with a Geforce RTX 3090 GPU and Ubuntu 22.04 system.

\begin{sidewaystable}[t]
\caption{Different classification methods in terms of OA, AA, and Kappa as well as the accuracies for each
class on the \underline{\textbf{Houston 2013}} dataset. The \textcolor{red}{\textbf{best one}} is highlighted as red. The \textcolor{blue}{\textbf{second best}} is highlighted as blue. \label{tab:HU2013 quant results}}

\centering
\begin{tabular}{|c|c||c|c|c|c|c|c|c|c|c|}
\hline
                 No.& Color. & vanilla ViT & SpeFormer & MAEST & HiT & SSFTT & 3DViT & HUSST & ViM & MiM (Ours)\\ \hline
                
1&\cellcolor{HU2013-1} & 89.129 & 90.815 & 89.644 & 89.616 & 91.085 & \textcolor{red}{\textbf{95.123}} & 92.073 & 87.271 & \textcolor{blue}{\textbf{94.887}} \\ \hline
2&\cellcolor{HU2013-2} & 94.852 & 93.950 & 91.537 & 92.643 & 93.056 & \textcolor{blue}{\textbf{97.550}} & 94.473 & 93.035 & \textcolor{red}{\textbf{98.273}} \\ \hline
3&\cellcolor{HU2013-3} & 96.188 & 97.087 & \textcolor{red}{\textbf{100}} & \textcolor{blue}{\textbf{99.901}} & 98.232 & \textcolor{red}{\textbf{100}} & \textcolor{red}{\textbf{100}} & 97.527 & \textcolor{red}{\textbf{100}} \\ \hline
4&\cellcolor{HU2013-4} & 93.285 & 96.843 & 95.083 & \textcolor{blue}{\textbf{97.016}} & 94.711 & \textcolor{red}{\textbf{97.681}} & 96.279 & 87.680 & 96.866 \\ \hline
5&\cellcolor{HU2013-5} & 96.445 & 94.575 & 96.748 & \textcolor{blue}{\textbf{97.959}} & 96.925 & 97.823 & 96.378 & 93.160 & \textcolor{red}{\textbf{99.858}} \\ \hline
6&\cellcolor{HU2013-6} & 74.011 & 93.464 & 93.770 & 91.374 & 84.713 & \textcolor{blue}{\textbf{99.306}} & 97.260 & 94.595 & \textcolor{red}{\textbf{99.652}} \\ \hline
7&\cellcolor{HU2013-7} & 82.431 & 91.276 & 90.431 & 91.541 & 87.516 & 90.009 & \textcolor{red}{\textbf{91.864}} & 84.735 & \textcolor{blue}{\textbf{91.423}} \\ \hline
8&\cellcolor{HU2013-8} & 74.693 & 68.837 & 75.459 & 81.531 & \textcolor{red}{\textbf{88.889}} & 72.029 & 76.727 & 84.537 & \textcolor{blue}{\textbf{87.167}} \\ \hline
9&\cellcolor{HU2013-9} & 73.527 & \textcolor{blue}{\textbf{86.340}} & 85.249 & 84.658 & \textcolor{red}{\textbf{88.835}} & 85.297 & 86.209 & 85.089 & 84.309 \\ \hline
10&\cellcolor{HU2013-10} & 72.931 & 71.959 & 86.763 & 88.628 & \textcolor{blue}{\textbf{91.663}} & 82.154 & 81.541 & 84.514 & \textcolor{red}{\textbf{92.380}} \\ \hline
11&\cellcolor{HU2013-11} & 69.631 & 75.279 & 86.429 & 84.187 & 87.778 & 87.748 & \textcolor{red}{\textbf{92.630}} & 83.675 & \textcolor{blue}{\textbf{90.028}} \\ \hline
12&\cellcolor{HU2013-12} & 70.809 & 78.114 & 75.945 & 82.155 & \textcolor{blue}{\textbf{88.132}} & 78.154 & 81.899 & 79.158 & \textcolor{red}{\textbf{89.581}} \\ \hline
13&\cellcolor{HU2013-13} & 63.805 & 90.429 & 91.667 & 91.501 & 90.203 & \textcolor{blue}{\textbf{92.632}} & 90.657 & 87.896 & \textcolor{red}{\textbf{93.080}} \\ \hline
14&\cellcolor{HU2013-14} & 96.509 & 97.436 & \textcolor{red}{\textbf{100}} & 93.690 & 97.041 & 95.367 & \textcolor{blue}{\textbf{99.597}} & 99.190 & 97.436 \\ \hline
15&\cellcolor{HU2013-15} & 98.943 & 95.075 & 96.728 & 96.629 & 91.245 & 97.399 & 95.122 & \textcolor{blue}{\textbf{97.612}} & \textcolor{red}{\textbf{98.333}} \\ \hline
OA&                      & 82.9548 & 86.4557 & 88.8415 & 90.0303 & \textcolor{blue}{\textbf{91.4241}} & 89.7024 & 90.1205 & 87.5871 & \textcolor{red}{\textbf{92.8917}} \\ \hline
AA&                      & 84.0284 & 88.3335 & 89.8294 & 91.7086 & \textcolor{blue}{\textbf{93.0019}} & 90.1268 & 93.0002 & 89.1248 & \textcolor{red}{\textbf{94.2108}}  \\ \hline
Kappa&                   & 0.8151 & 0.8530 & 0.8787 & 0.8918 & \textcolor{blue}{\textbf{0.9069}} & 0.8882 & 0.8928 & 0.8653 & \textcolor{red}{\textbf{0.9228}} \\ \hline
\end{tabular}

\end{sidewaystable}

\subsection{\textbf{Brief Description and Settings of Compared Methods}}
We conducted comparative analyses against a range of baseline and state-of-the-art models. These included vanilla ViT (ViT) \cite{vit}, SpeFormer \cite{vit1}, MAEST \cite{vit3}, HiT \cite{vit4}, SSFTT \cite{vit8}, 3DViT \cite{vit9}, HUSST \cite{vit10}, and ViM \cite{vim}. All models were employed as patch-wise classifiers, with parameter settings adhering to their respective references, especially the initial patch size for different datasets.
The comparison is grounded in the same number of the training samples without data augmentation.
\begin{itemize}
    \item [1)] ViT is configured with a feature dimension of 64 and incorporates 4 multi-heads self-attention. The network depth is set to 4 layers. Across all datasets, the initial size of the cropped patch is $9 \times 9$. We use a sub-patch size of $3 \times 3$, partitioned in a non-overlapping manner.
    \item [2)] SpeFormer integrates group-wise spectral embedding into the Transformer encoder and includes cross-layer adaptive fusion modules to boost the encoder's learning capabilities. It is configured with an initial patch size of $7 \times 7$ across all datasets in our experiments.
    \item [3)] MAEST integrates a reconstruction path with a classification path for HSI classification, employing masking auto-encoding (MAE). The initial patch size is set at $7 \times 7$.
    \item [4)] HiT embeds convolution operations into the Transformer structure to capture the subtle spectral discrepancies and convey the local spatial context information. For all datasets, the initial patch size is set at $15 \times 15$ for HiT. 
    \item [5)] SSFTT combines 3D-CNNs and 2D-CNNs, featuring a Gaussian-weighted feature tokenizer to prepare extracted features for the Transformer encoder. In our experiments, patch sizes vary by dataset: $13 \times 13$ for the IP and PU dataset and $9 \times 9$ for both HU2013 and HongHu dataset.
    \item [6)]3DViT integrates spectral and spatial features into sub-cubes using a 3D configuration within the ViT framework. It implements 3D coordinate positional embedding and a local-global feature fusion mechanism, merging local details with global context. For all datasets, the initial patch size for the 3DViT model is set at $9 \times 9$. 
    \item [7)] HUSST implements a cascaded approach, combining a spectral Transformer with a spatial local-global Transformer hierarchically. It introduces cross-scale local-global feature fusion for multi-scale feature aggregation. The initial patch size set at $11 \times 11$ for all datasets.
    \item [8)] ViM, the first Mamba model tailored for vision tasks, is applied to HSI classification to assess its viability. The hyperparameters are aligned with those of our proposed MiM model, detailed later.
    \item [9)] Our MiM model standardizes all feature dimensions at 64, with T-Mamba encoders having a depth of 2. The initial patch sizes are set at $7 \times 7$ for the IP dataset, $11 \times 11$ for the PU dataset, and $9 \times 9$ for the HU2013 and HongHu dataset. Others can be found in Algorithm \ref{alg:proposed_method}.
\end{itemize}

\begin{sidewaystable}[t]

\caption{Different classification methods in terms of OA, AA, and Kappa as well as the accuracies for each
class on the \underline{\textbf{WHU-Hi-HongHu}} dataset. The \textcolor{red}{\textbf{best one}} is highlighted as red. The \textcolor{blue}{\textbf{second best}} is highlighted as blue. \label{tab:Honghu quant results}}
\centering
\begin{tabular}{|c|c||c|c|c|c|c|c|c|c|c|}
\hline
                 No.& Color. & vanilla ViT & SpeFormer & MAEST & HiT & SSFTT & 3DViT & HUSST & ViM & MiM (Ours)\\ \hline
                
1&\cellcolor{HH-1} &83.308  &92.414  &93.150  &92.397  &\textcolor{red}{\textbf{94.062}}  &\textcolor{blue}{\textbf{93.913}}  &93.392  &90.774  &91.922  \\ \hline
2&\cellcolor{HH-2} &51.588  &77.377  &69.992  &83.118  &80.321  &\textcolor{red}{\textbf{84.327}}  &82.783  &78.092  &\textcolor{blue}{\textbf{83.239}} \\ \hline
3&\cellcolor{HH-3} &66.129  &87.049  &87.753  &86.075  &86.447  &\textcolor{blue}{\textbf{88.706}}  &86.066  &85.647  &\textcolor{red}{\textbf{89.772}} \\ \hline
4&\cellcolor{HH-4} &81.673  &92.768  &95.519  &\textcolor{blue}{\textbf{95.807}}  &92.688  &93.738  &94.840  &90.939  &\textcolor{red}{\textbf{95.993}}  \\ \hline
5&\cellcolor{HH-5} &30.789  &42.216  &52.979  &50.280  &39.060  &45.475  &\textcolor{blue}{\textbf{59.836}}  &40.896  &\textcolor{red}{\textbf{59.971}}  \\ \hline
6&\cellcolor{HH-6} &80.313  &93.287  &94.380  &94.295  &93.865  &93.009  &\textcolor{blue}{\textbf{94.936}}  &92.830  &\textcolor{red}{\textbf{96.003}}  \\ \hline
7&\cellcolor{HH-7} &55.159  &69.221  &81.213  &66.949  &\textcolor{red}{\textbf{84.649}}  &74.366  &79.445  &81.871  &\textcolor{blue}{\textbf{83.873}}  \\ \hline
8&\cellcolor{HH-8} &29.825  &48.326  &45.361  &47.745  &46.911  &\textcolor{red}{\textbf{64.771}}  &45.650  &54.258  &\textcolor{blue}{\textbf{64.170}}  \\ \hline
9&\cellcolor{HH-9} &87.712  &95.937  &95.455  &93.624  &\textcolor{blue}{\textbf{96.133}}  &94.108  &95.912  &95.570  &\textcolor{red}{\textbf{96.399}}  \\ \hline
10&\cellcolor{HH-10} &50.244  &78.034  &\textcolor{blue}{\textbf{85.437}}  &78.864  &82.318  &85.436  &83.367  &77.348  &\textcolor{red}{\textbf{87.094}}  \\ \hline
11&\cellcolor{HH-11} &32.714  &68.521  &80.407  &64.692  &76.228  &\textcolor{blue}{\textbf{82.055}}  &72.825  &75.274  &\textcolor{red}{\textbf{88.261}}  \\ \hline
12&\cellcolor{HH-12} &38.737  &60.166  &72.607  &64.964  &\textcolor{blue}{\textbf{76.981}}  &65.600  &69.342  &68.752  &\textcolor{red}{\textbf{77.643}}  \\ \hline
13&\cellcolor{HH-13} &41.373  &66.599  &\textcolor{blue}{\textbf{75.962}}  &60.533  &72.911  &70.993  &75.059  &66.550  &\textcolor{red}{\textbf{76.611}}  \\ \hline
14&\cellcolor{HH-14} &53.612  &75.796  &88.200  &85.221  &85.478  &\textcolor{red}{\textbf{90.432}}  &86.554  &74.109  &\textcolor{blue}{\textbf{89.186}}  \\ \hline
15&\cellcolor{HH-15} &48.922  &80.839  &84.730  &83.587  &\textcolor{blue}{\textbf{84.911}}  &70.855  &72.665  &37.689  &\textcolor{red}{\textbf{87.631}}  \\ \hline
16&\cellcolor{HH-16} &85.969  &\textcolor{red}{\textbf{96.110}}  &94.660  &94.505  &95.649  &95.244  &\textcolor{blue}{\textbf{96.007}}  &93.233  &95.948  \\ \hline
17&\cellcolor{HH-17} &71.222  &89.171  &74.695  &\textcolor{red}{\textbf{94.118}}  &85.529  &87.956  &87.207  &85.676  &\textcolor{blue}{\textbf{92.016}}  \\ \hline
18&\cellcolor{HH-18} &49.649  &84.597  &83.587  &85.643  &83.333  &84.548  &86.010  &\textcolor{red}{\textbf{89.097}}  &\textcolor{blue}{\textbf{86.875}}  \\ \hline
19&\cellcolor{HH-19} &59.138  &83.529  &86.927  &82.922  &\textcolor{blue}{\textbf{90.572}}  &86.385  &88.418  &86.205  &\textcolor{red}{\textbf{91.872}}  \\ \hline
20&\cellcolor{HH-20} &34.296  &71.906  &80.832  &32.043  &79.922  &80.556  &\textcolor{blue}{\textbf{82.181}}  &74.823  &\textcolor{red}{\textbf{88.583}}  \\ \hline
21&\cellcolor{HH-21} &20.711  &50.048  &52.946  &52.932  &46.456  &\textcolor{blue}{\textbf{62.122}}  &44.932  &56.133  &\textcolor{red}{\textbf{71.989}}  \\ \hline
22&\cellcolor{HH-22} &32.600  &66.921  &80.489  &87.111  &68.578  &\textcolor{red}{\textbf{83.324}}  &67.961  &53.190  &\textcolor{blue}{\textbf{83.057}}  \\ \hline

OA&                      &65.2688  &82.9934  &\textcolor{blue}{\textbf{87.3104}}  &84.2750  &85.2845  &85.8091  &86.9473  &82.7789  &\textcolor{red}{\textbf{89.5749}}  \\ \hline
AA&                      &67.4351  &84.6977  &85.7987  &83.7510  &87.4042  &\textcolor{blue}{\textbf{88.7126}}  &87.9006  &85.7671  &\textcolor{red}{\textbf{90.9751}}   \\ \hline
Kappa&                   &0.5873  &0.7911  &\textcolor{blue}{\textbf{0.8427}}  &0.8048  &0.8196  &0.8254  &0.8387  &0.7897  &\textcolor{red}{\textbf{0.8705}}  \\ \hline
\end{tabular}

\end{sidewaystable}

\begin{figure*}[!t]
    \centering
    \includegraphics[scale=0.4]{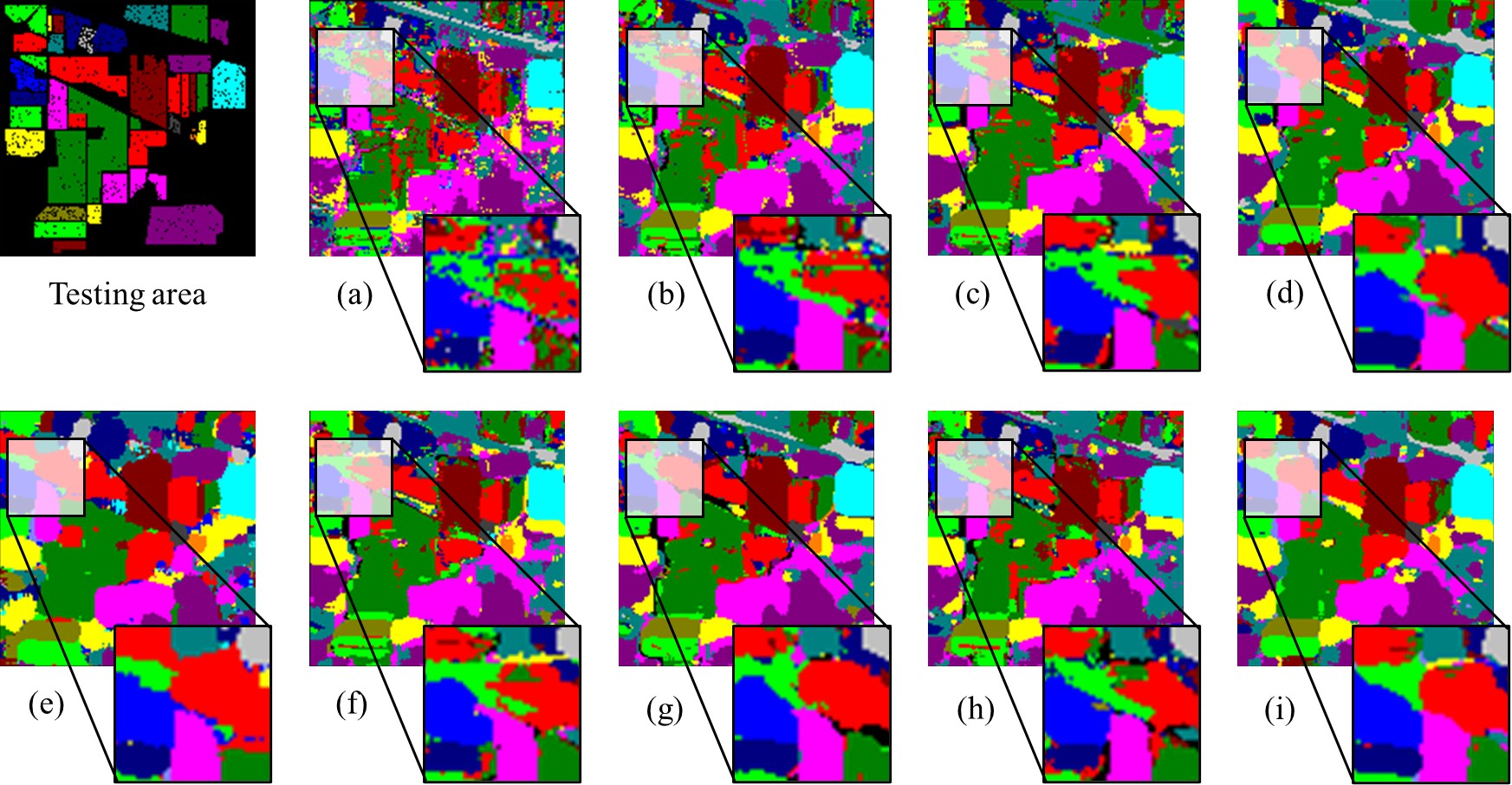}
    \caption{The classification maps with different methods for the IP dataset. (a) vanilla ViT (OA=74.0875\%). (b) SpeFormer (OA=79.9607\%). (c) MAEST (OA= 83.0731\%) (d) HiT (OA=86.0511\%). (e) SSFTT (OA=88.7498\%). (f) 3DViT (OA=85.1411\%). (g) HUSST (OA=87.3333\%). (h) ViM (OA= 81.9667\%). (i) MiM (OA=92.0794\%).}
    \label{fig:IP classification results}
\end{figure*}
\begin{figure*}[t]
    \centering
    \includegraphics[scale=0.6]{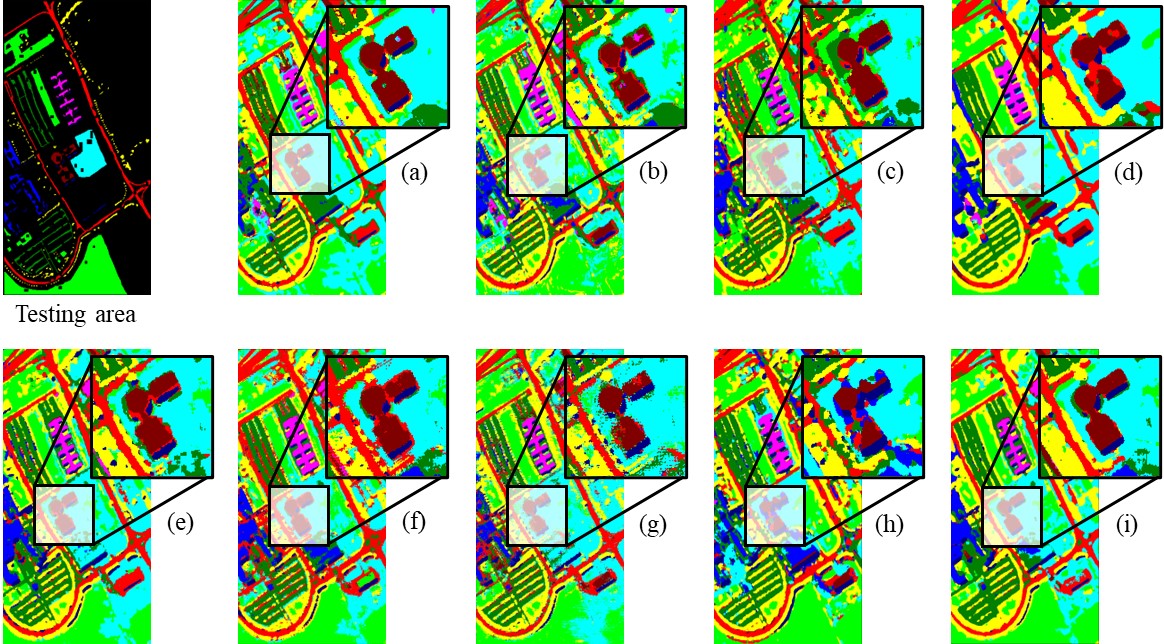}
    \caption{The classification maps with different methods for the PU dataset. (a) vanilla ViT (OA=76.0313\%). (b) SpeFormer (OA=86.2784\%). (c) MAEST (OA=88.0433\%) (d) HiT (OA=87.9183\%). (e) SSFTT (OA=88.8707\%). (f) 3DViT (OA=86.5258\%). (g) HUSST (OA=86.2483\%). (h) ViM (OA=78.1811\%). (i) MiM (OA=91.5756\%).}
    \label{fig:PU classification results}
\end{figure*}
\begin{figure*}[t]
    \centering
    \includegraphics[scale=0.6]{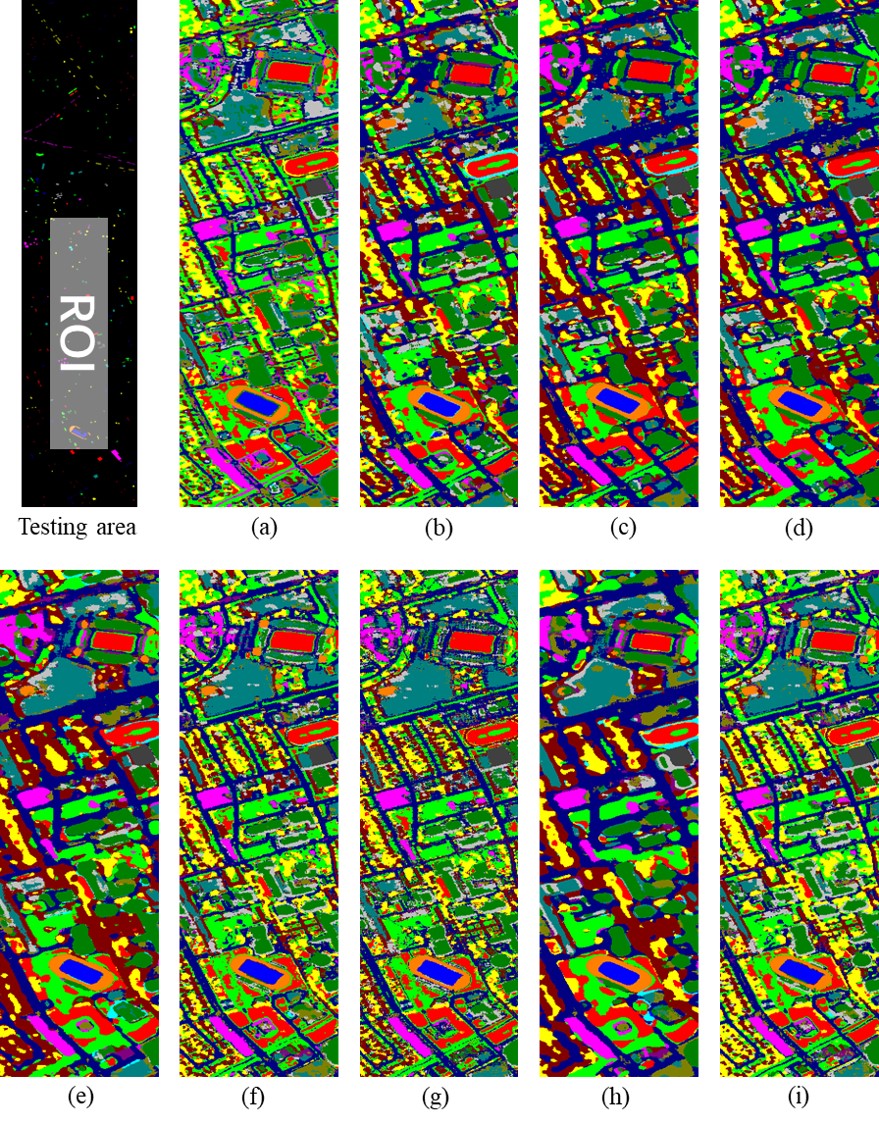}
    \caption{The region of interest (ROI) in classification maps with different methods for the HU2013 dataset. (a) vanilla ViT (OA=82.9548\%). (b) SpeFormer (OA=86.4557\%). (c) MAEST (OA=88.8415\%) (d) HiT (OA=90.0303\%). (e) SSFTT (OA=91.4241\%). (f) 3DViT (OA=89.7024\%). (g) HUSST (OA=90.1205\%). (h) ViM (OA=87.5871 \%). (i) MiM (OA=92.8917\%).}
    \label{fig:HU2013 classification results}
\end{figure*}
\begin{figure*}[t]
    \centering
    \includegraphics[scale=0.4]{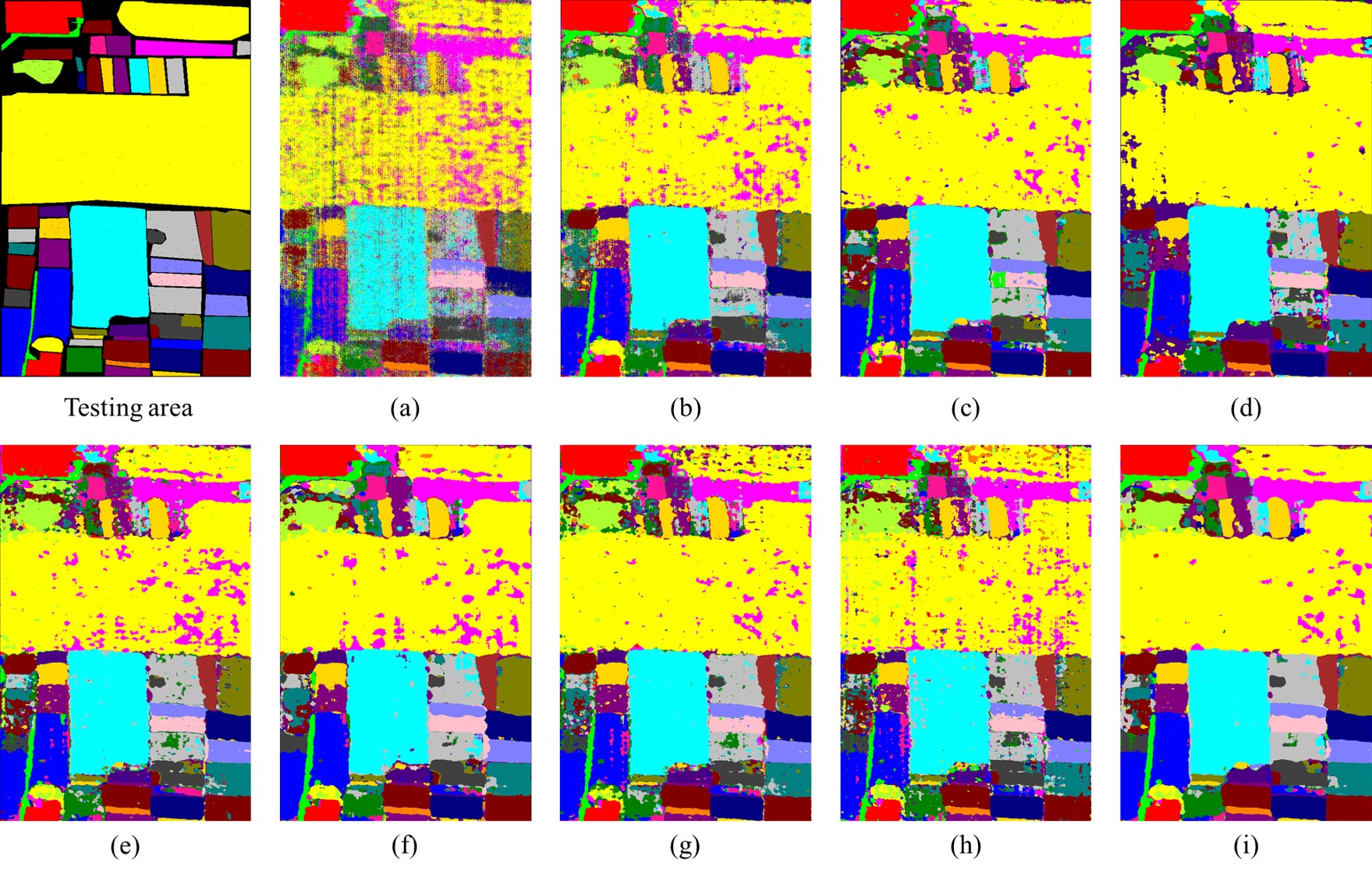}
    \caption{The classification maps with different methods for the HongHu dataset. (a) vanilla ViT (OA=65.2688\%). (b) SpeFormer (OA=82.9934\%). (c) MAEST (OA=87.3104\%) (d) HiT (OA=84.2750\%). (e) SSFTT (OA=85.2845\%). (f) 3DViT (OA=85.8091\%). (g) HUSST (OA=86.9473\%). (h) ViM (OA=82.7789\%). (i) MiM (OA=89.5749\%).}
    \label{fig:Honghu classification results}
\end{figure*}

\subsection{\textbf{Experimental Results}}

\subsubsection{\textbf{Quantitative Results and Classification Maps}}
Tables \ref{tab:IP quant results}, \ref{tab:PU quant results}, \ref{tab:HU2013 quant results}, and \ref{tab:Honghu quant results} present average classification metrics under 5 times of running, such as OA, AA, Kappa coefficient, and class-wise accuracy for the IP, PU, HU2013, and HongHu datasets, respectively. Corresponding classification maps are displayed in Figs. \ref{fig:IP classification results}, \ref{fig:PU classification results}, \ref{fig:HU2013 classification results}, and \ref{fig:Honghu classification results}. 

From these results, it's evident that baseline methods like vanilla ViT and ViM do not yield satisfactory outcomes, especially the vanilla ViT, which struggles due to its purely parallel and simultaneous processing approach (e.g., self-attention). This often results in noise-like classification artifacts, particularly noticeable in smooth regions and at the boundaries of land-covers, as the classification map for vanilla ViT exhibits much sharper visual artifacts. Conversely, ViM, which integrates bidirectional sequence processing with Mamba encoders, typically shows better performance. This improvement could be attributed to an ordering bias, akin to that found in RNNs, which is advantageous when training samples are limited.

SpeFormer and MAEST, enhancements over vanilla ViT, achieve significantly better results. SpeFormer introduces local feature aggregation within spatial and spectral domains and incorporates cross-layer connections to enhance the robustness of its encoder. Compared to vanilla ViT, SpeFormer produces smoother classifications in land-cover regions. MAEST integrates a masked auto-encoder into SpeFormer, adding a noise mask to the original data and reconstructing it, which enhances generative performance on unseen data, especially when training and testing samples are uncorrelated. MAEST presents better improvements on SpeFormer.

On another front, 3DViT extends ViT into a 3D sequential model across spatial and spectral domains, effectively linking spatial and spectral features to enhance classification accuracy. Despite being a pure Transformer-based model without hybrid elements like RNNs or CNNs, it shows promising results.

Moreover, HiT, SSFTT, and HUSST combine CNNs with Transformers to leverage local and global feature awareness, generally yielding superior results. These models provide clearer distinctions in flat land-cover regions and at boundaries. However, their larger model sizes contribute to higher computational demands and are less hardware-friendly.

\begin{sidewaystable}[t]
\caption{Total running time (minutes) and the number of parameters required by different methods on the \underline{\textbf{Indian Pines}} dataset.\label{tab:computation costs}}
\centering
\begin{tabular}{|c||c|c|c|c|c|c|c|c|c|}
\hline
       & vanilla ViT & SpeFormer & MAEST & HiT & SSFTT & 3DViT & HUSST & ViM & MiM (Ours) \\ \hline
Train (minutes.)  & 2.1     &  3.5   &4.7      & 5.6       & 5.5        & 4.5        & 3.5            & 2.5          & 3.6      \\ \hline
Test (seconds.)   & 2.0     &  4.0   &5.0      & 11.0       & 5.0         & 8.0       & 3.0             & 7.0          & 5.0     \\ \hline
Parameters     & 416K    & 644K      &953K      & 45.8M       & 2.9M   & 2.4M        &9.1M            &357K          &1.1M      \\ \hline
\end{tabular}
\end{sidewaystable}

Our method, MiM, introduces a MCS mechanism with a T-Mamba encoder, facilitating lightweight, multi-directional feature generation. Compared to ViM, our specialized T-Mamba encoder and extended bidirectional scan improve classification performance significantly. Our approach, notably lighter than HiT, SSFTT, and HUSST, produces classification maps with smoother regions and more distinct land-cover boundaries. The class-wise accuracies are competitive and generally superior to other methods.

Key differences in classification performance are highlighted and marked in Figs. \ref{fig:IP classification results}, and \ref{fig:PU classification results} for detailed comparisons.

This exploration paves the way for our future research into adapting pure Mamba models and MCS into a 3D configuration, similar to 3DViT which extends ViT into 3D version.

\begin{figure*}[t]
    \centering
    \includegraphics[scale = 0.47]{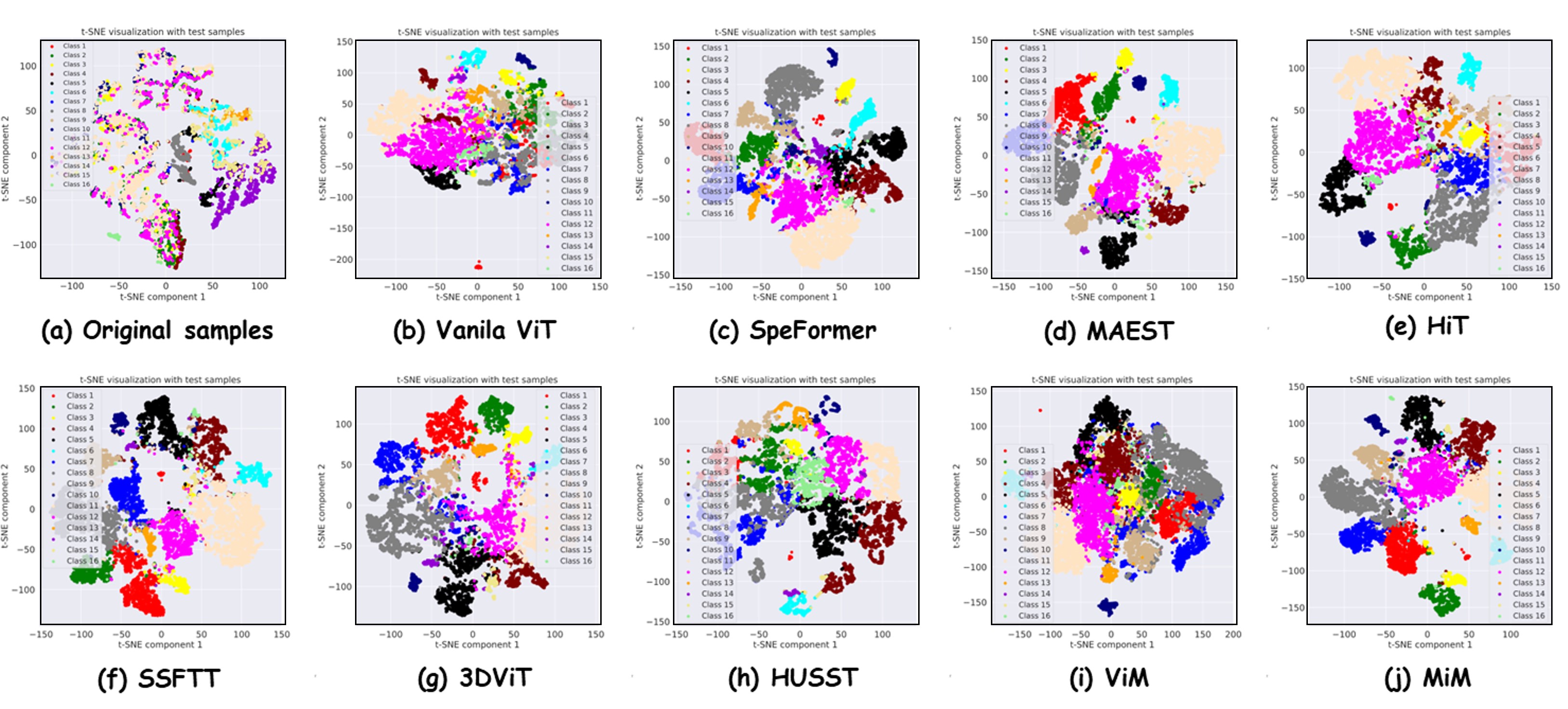}
    \caption{Visualization of T-SNE results on the IP test samples by different methods. (a) original test samples' T-SNE results. (b) vanilla ViT. (c) SpeFormer. (d) MAEST. (e) HiT. (f) SSFTT. (g) 3DViT. (h) HUSST. (i) ViM. (j) MiM.}
    \label{fig:IP TSNE}
\end{figure*}

\subsubsection{\textbf{Learned Feature Visualizations by T-SNE}}

Deep learning-based methods excel in creating new feature representations that replace the original inputs through representation learning. Typically, these methods train a discriminative model to transform input data into a distinct output representation for tasks such as HSI classification, subsequently utilizing these new features for clustering.

To evaluate the representational capabilities of various methods, we employed t-distributed stochastic neighbor embedding (T-SNE)\footnote{\url{https://en.wikipedia.org/wiki/T-distributed_stochastic_neighbor_embedding}}, a tool that visualizes the distribution of learned features from test samples.

Using the IP dataset as an example, Fig. \ref{fig:IP TSNE} illustrates the visualization outcomes of different methods. From this, two key points emerge. Firstly, while some clusters represented by different colors may appear closely positioned in T-SNE results, this proximity should not be interpreted as indicative of similarity due to T-SNE’s internal mechanics. Secondly, the presence of same-color clusters at considerable distances from each other highlights T-SNE’s focus on preserving local structure while optimizing a non-convex function, which may lead to local minima.

The visualization clearly demonstrates that our MiM model effectively discriminates between learned feature representations. Through T-SNE, the new feature representations not only cluster but also show increased discriminative properties, with several clusters unifying under the same color and appearing smoother.

\subsubsection{\textbf{Comparison of Computational Costs}}
To evaluate computational efficiency across all methods, we conducted a detailed statistical analysis, summarized in Table \ref{tab:computation costs}. This analysis adheres to previously described experimental setups and includes metrics such as total training time (minutes), testing time (seconds), and model size (number of parameters) for the IP dataset. 


Our method showcases competitive processing times, significantly outperforming advanced models like HiT, MAEST, 3DViT, and SSFTT in terms of efficiency. However, it records slightly longer times compared to simpler models such as vanilla ViT, SpeFormer, HUSST, and ViM. When it comes to the number of trainable parameters, our approach proves to be more efficient, requiring fewer parameters than other leading methods, which underscores its parameter efficiency.

In summary, our method strikes an effective balance between running time and model complexity, measured by parameter count. While it doesn't achieve the fastest training or testing times, it offers lower operational times than several sophisticated models and maintains a manageable parameter count. This combination of efficiency and simplicity makes our method particularly suitable for scenarios where both performance and model manageability are priorities.

\subsection{\textbf{Other Analysis and Ablation Study}}
In this part, a comprehensive analysis, experimental evaluation, and ablation studies have been presented.

\begin{table}[t]
\caption{Impact of different insertable components in T-Mamba encoder on the overall classification results (\%). The \textcolor{red}{\textbf{best one}} is highlighted as red. \label{tab:ablation}}
\centering
\begin{tabular}{|c|c|c||c|c|c|c|}
\hline
    $STL$ & $GDM$ & $STF$ & \textbf{IP} & \textbf{PU} & \textbf{HU2013} & \textbf{HongHu} \\ \hline
      \xmark    &       \xmark      &   \xmark  &  82.559   &    80.999   &    88.112    &84.123 \\ \hline
   \textcolor{green}{\cmark}    &      \xmark       &  \xmark   &  86.129  &    85.698    &    90.995    &86.889 \\ \hline
   \xmark   &      \textcolor{green}{\cmark}       &  \xmark   &  84.006  &    82.159    &    89.459    &86.114 \\ \hline
   \xmark  &      \xmark       &  \textcolor{green}{\cmark}   &  84.561  &    83.226    &     90.642  &85.013 \\ \hline
   \textcolor{green}{\cmark} & \textcolor{green}{\cmark} &  \xmark   &  87.021   &    86.111    &    91.324    &88.546 \\ \hline
   \textcolor{green}{\cmark} & \xmark&  \textcolor{green}{\cmark}   &  91.969  &    89.123    &    91.452    &88.982 \\ \hline
   \xmark   & \textcolor{green}{\cmark} & \textcolor{green}{\cmark} &  87.159  &   87.661     &    91.003   & 88.002\\ \hline
   \textcolor{green}{\cmark}   & \textcolor{green}{\cmark} & \textcolor{green}{\cmark} &   \textcolor{red}{\textbf{92.079}}   &   \textcolor{red}{\textbf{91.576}}    &   \textcolor{red}{\textbf{92.892}}    &\textcolor{red}{\textbf{89.575}}\\ \hline
         
\end{tabular}
\end{table}

\subsubsection{\textbf{Ablation on Proposed Different Components}} 
In the T-Mamba encoder, we integrated several key components: the STL, GDM, and STF. To assess their effectiveness, we conducted an ablation study, the results of which are summarized in Table \ref{tab:ablation}. The ablation study details how each component was tested. In scenarios where STL and STF were not used, we substituted them with general average pooling and simple summation, respectively, to gauge their relative performance.

STL outperforms average pooling in merging MCS features, suggesting its superior ability to capture contextual information within the merged features, significantly enhancing accuracy.

GDM refines feature focus after Mamba encoding by reweighting the sequence based on the proximity to the central pixel, which is the last index in a sub-sequence. This targeted approach enhances the central pixel's prominence, reducing the impact of interfering pixels and yielding modest improvements.

STF fuses enhanced features more effectively than simple summation, leveraging higher-level learned representations to enrich the raw feature set. Incorporating STF alone leads to better outcomes than summation, underscoring its utility in strengthening feature representations.

Additionally, combining STL and STF yielded the most substantial improvements, likely due to their complementary mechanisms both utilizing sequential attention.

Overall, this ablation study demonstrates that the integration of STL, GDM, and STF significantly boosts performance by synergistically refining the handling of image band features in a sequential model, particularly with the continuous scanning features critical in our proposed MiM model.

\begin{table}[t]
\centering
\caption{Impact of different initial patch size for the overall accuracy (\%) on four datasets. The \textcolor{red}{\textbf{best one}} is highlighted as red. \label{tab:patch size}}

\begin{tabular}{|c||c|c|c|c|}
\hline
Patch Size ($p$) & \textbf{IP} & \textbf{PU} & \textbf{HU2013} & \textbf{HongHu} \\ \hline
$3 \times 3$ & 89.001 \textcolor{red}{$\uparrow$}     &86.331 \textcolor{red}{$\uparrow$}    &89.123 \textcolor{red}{$\uparrow$}     & 83.127 \textcolor{red}{$\uparrow$}\\ \hline
$5 \times 5$ & 91.332 \textcolor{red}{$\uparrow$}     &89.126 \textcolor{red}{$\uparrow$}    &91.028 \textcolor{red}{$\uparrow$}     &87.269 \textcolor{red}{$\uparrow$}\\ \hline
$7 \times 7$ & \textcolor{red}{\textbf{92.079}} \textcolor{red}{$\uparrow$}   &90.214 \textcolor{red}{$\uparrow$}    &91.999 \textcolor{red}{$\uparrow$}     &89.113 \textcolor{red}{$\uparrow$}\\ \hline
$9 \times 9$ & 91.554 \textcolor{green}{$\downarrow$}         &91.002 \textcolor{red}{$\uparrow$}    &  \textcolor{red}{\textbf{92.892}} \textcolor{red}{$\uparrow$}  &\textcolor{red}{\textbf{89.575}} \textcolor{red}{$\uparrow$}\\ \hline
$11 \times 11$ & 90.667 \textcolor{green}{$\downarrow$}        & \textcolor{red}{\textbf{91.576}} \textcolor{red}{$\uparrow$}  &92.091 \textcolor{green}{$\downarrow$}      &88.990 \textcolor{green}{$\downarrow$}\\ \hline
$13 \times 13$ & 89.214 \textcolor{green}{$\downarrow$}        &90.885 \textcolor{green}{$\downarrow$}      &90.451 \textcolor{green}{$\downarrow$}      &87.317 \textcolor{green}{$\downarrow$}\\ \hline
$15 \times 15$ & 87.332 \textcolor{green}{$\downarrow$}       &88.213 \textcolor{green}{$\downarrow$}      &88.519 \textcolor{green}{$\downarrow$}      &86.451 \textcolor{green}{$\downarrow$}\\ \hline
\end{tabular}

\end{table}

\subsubsection{\textbf{Analysis on Initial Patch Size}}
We investigated the effect of initial patch size on classification accuracy. For instance, with the IP dataset, the optimal starting patch size is 7 $\times$ 7. This setup allows us to derive four scale features—scale 7, 5, 3, and 1—each contributing to the loss function. Therefore, the final result is obtained by training four loss functions together.

As shown in Table \ref{tab:patch size}, we evaluated a range of initial patch sizes (3, 5, 7, 9, 11, 13, 15). The results, specifically OA for four datasets, are detailed in the same table.

An appropriately large initial patch size can improve model performance by capturing more spatial context. However, excessively large patch sizes can be detrimental, introducing noise from interfering pixels.

For the relatively compact IP dataset, the ideal initial patch size is small, confirmed at 7 $\times$ 7. In contrast, for the HU2013 dataset, which features a dense spatial distribution but closely spaced land-covers, a slightly larger patch size of 9 $\times$ 9 is optimal. The PU dataset, characterized by its flat and smooth terrain, warrants an even larger optimal initial patch size of 11 $\times$ 11. Finally, the HongHu dataset gets the ideal initial patch size at 9 $\times$ 9.

\subsubsection{\textbf{Investigation and Comparison on Scan Design}}
In the proposed centralized MCS, selecting the appropriate scan manner is essential for optimal feature allocation and positioning within sequential models. It is critical to choose a cross-scan design that effectively spans spatial domains.

\begin{figure}[t]
    \centering
    \includegraphics[scale=0.45]{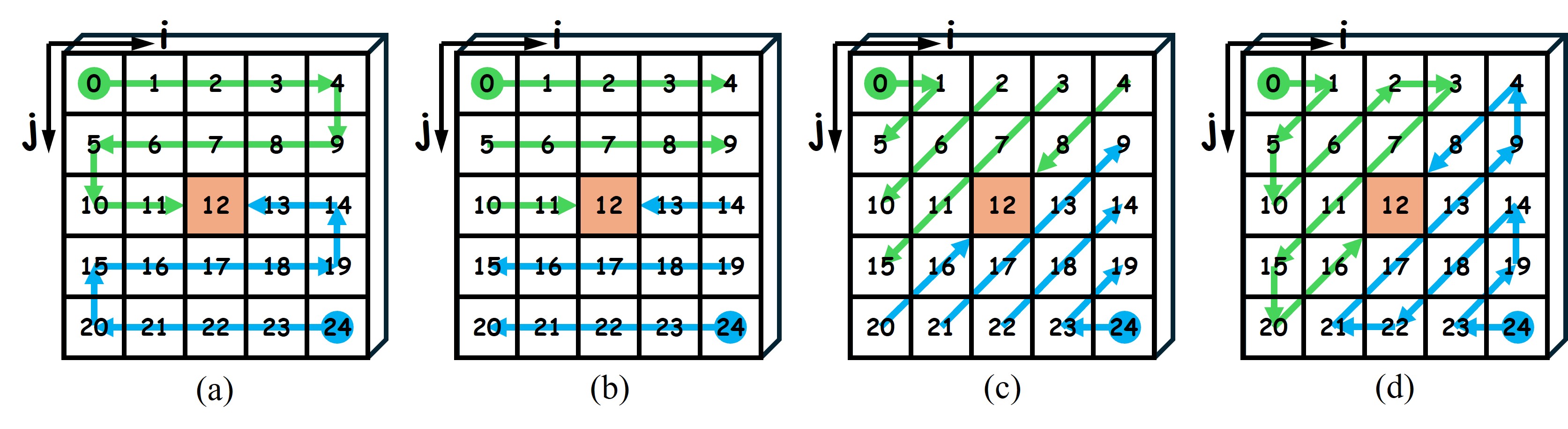}
    \caption{Investigated scan designs. (a) proposed Mamba-scan design. (b) general Raster-scan design. (c) Diagonal-scan design. (d) Zig-Zag-scan design. Each design can be extended into four types from different start and end positions, similar to those shown in Fig. \ref{fig:four types of cross mamba scan}.}
    \label{fig:scan type}
\end{figure}

Given that random and casual scanning may lead to significant redundancy and poor feature arrangement, we established a certain rule: spatially, the scanning should keep continuous as much as possible. Thus, through investigation, we identified several scanning designs, as shown in Fig. \ref{fig:scan type}. For each scan design, it can be extended into four types, similar to those shown in Fig. \ref{fig:four types of cross mamba scan}.

The Mamba-scan achieves the greatest continuity when converting an image into a sequence. The Raster-scan, a common method, linearizes the image into a sequence but introduces jump connections at the boundaries of the image. The Diagonal-scan approaches this by scanning images diagonally, selecting positions along the diagonals. Conversely, the Zig-zag-scan modifies the diagonal approach by maintaining partial continuity at image boundaries.

\begin{table}[t]
\footnotesize
\caption{The comparison of using different scan designs for OA (\%). The \textcolor{red}{\textbf{best one}} is highlighted as red. The \textcolor{blue}{\textbf{second best}} is highlighted as blue.  \label{tab:scan type}}
\centering
\begin{tabular}{|c||c||c|c|c|c|}
\hline
                               & Num. of used & \textbf{IP} & \textbf{PU} & \textbf{HU2013} & \textbf{HongHu}\\ \hline
                               & 1                 & 89.123    & 89.559    & 90.123        &87.315\\ \cline{2-6} 
\multirow{2}{*}{Mamba-scan}    & 2                 & 90.156    & 90.452    & 91.029       &88.569\\ \cline{2-6} 
                               & 3                 & 91.221    & 91.001    & 91.993        &89.226\\ \cline{2-6} 
                               & 4                 &\textcolor{red}{\textbf{92.079}}   &   \textcolor{red}{\textbf{91.576}}    &   \textcolor{red}{\textbf{92.892}}        &\textcolor{red}{\textbf{89.575}}\\ \hline
                               & 1                    & 88.326    & 87.009    & 89.163       &85.429\\ \cline{2-6} 
\multirow{2}{*}{Raster-scan}     & 2                    & 89.555    & 88.322    & 90.442       &86.129\\ \cline{2-6} 
                               & 3                    & 90.122    & 89.125    & 91.105       &88.226\\ \cline{2-6} 
                               & 4                    & \textcolor{blue}{\textbf{91.612}}    & 90.761    & \textcolor{blue}{\textbf{91.986}}       &\textcolor{blue}{\textbf{89.125}} \\ \hline
                               & 1                    & 88.039    & 86.002    & 89.013       &83.664\\ \cline{2-6} 
\multirow{2}{*}{Diagonal-scan}   & 2                    & 89.196    & 87.569    & 89.669       &86.548\\ \cline{2-6} 
                               & 3                    & 90.236    & 88.555    & 90.139       &87.841\\ \cline{2-6} 
                               & 4                    & 91.002    & 89.571    & 91.004        &88.221\\ \hline
                               & 1                    & 88.129    & 87.563    & 87.125       &84.003\\ \cline{2-6} 
 \multirow{2}{*}{Zig-Zag-scan}   & 2                    & 90.009    & 87.993    & 89.699       &86.443\\ \cline{2-6} 
                               & 3                    & 91.011    & 88.961    & 90.861       &87.231\\ \cline{2-6} 
                               & 4                    & 91.536    & \textcolor{blue}{\textbf{91.299}}    & 91.834        &87.992\\ \hline
\end{tabular}

\end{table}

We tested these scanning methods to evaluate their performance, with results summarized in Table \ref{tab:scan type}. Each scan design was tested using one to four types. The most effective results were achieved using the Mamba-scan with four directional types, confirming that a continuous sequence arrangement enhances content recognition. This finding is further supported by comparing the Diagonal-scan and the Zig-zag-scan, where the latter showed slight improvements due to its semi-continuous positioning. The commonly used Raster-scan, despite its prevalence in image transformation tasks, tends to disrupt spatial continuity, particularly when spatial context is significant in RS land-cover analysis, resulting in inferior performance compared to the Mamba-scan.

Additionally, employing four types of scans enhances each scanning approach, demonstrating the viability of multi-directional feature learning.

\subsubsection{\textbf{Exploring the Feature Fusion Strategy in T-Mamba}}
\begin{figure}[t]
    \centering
    \includegraphics[scale=0.35]{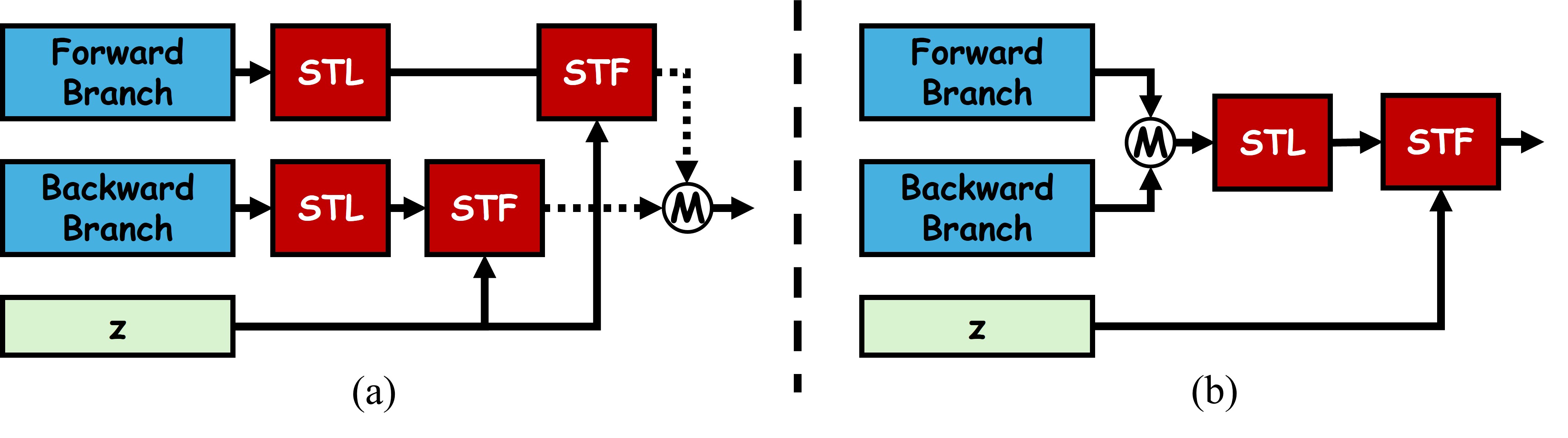}
    \caption{Designed bi-directional feature fusion manner in T-Mamba encoder. (a) Post-Merge strategy, and (b) Pre-Merge strategy. }
    \label{fig:two fusion manners}
\end{figure}

We developed our bi-directional feature fusion strategy in T-Mamba encoder with two ways, as shown in Fig. \ref{fig:two fusion manners}: 1) Post-Merge strategy, and 2) Pre-Merge strategy. To evaluate the effectiveness of these two strategies, we conducted experiments across four datasets. The results of these experiments are detailed in Table \ref{tab:feature fusion}.

\begin{table}[t]
\centering
\caption{Impact of different feature fusion strategy in T-Mamba encoder for the OA (\%) on four datasets. The \textcolor{red}{\textbf{best one}} is highlighted as red. \label{tab:feature fusion}}
\begin{tabular}{|c||c|c|c|c|}
\hline
             & \textbf{IP} & \textbf{PU} & \textbf{HU2013} & \textbf{HongHu} \\ \hline
Post-Merge & 91.326            & 90.332            & 91.229    & 88.450\\ \hline
Pre-Merge  & \textcolor{red}{\textbf{92.079}}   &   \textcolor{red}{\textbf{91.576}}    &   \textcolor{red}{\textbf{92.892}}     &\textcolor{red}{\textbf{89.575}}\\ \hline
\end{tabular}

\end{table}

We observed that the Pre-Merge strategy tends to be more effective for feature condensation and aggregation. This is likely because the STL module is applied to the merged features, which encapsulate the entirety of the image's features. Consequently, the STL can comprehensively analyze the whole image's contents, facilitating the condensation and abstraction of features, akin to a 'global' awareness of the image. Subsequently, the STF module leverages the pooled feature of the entire image, $\mathbf{z}$, to achieve a holistic understanding of image features. In contrast, the Post-Merge strategy, applies the STL and STF to each branch independently without integrating semantic information across the branches. This approach results in less effective feature integration compared to the Pre-Merge strategy.

Therefore, rather than directly adopting the ViM model, separating two individual branches,  for our HSI classification task, we have developed a more suitable and innovative approach that enhances performance through strategic feature merging and processing.

\begin{figure}[t]
    \centering
    \includegraphics[scale=0.33]{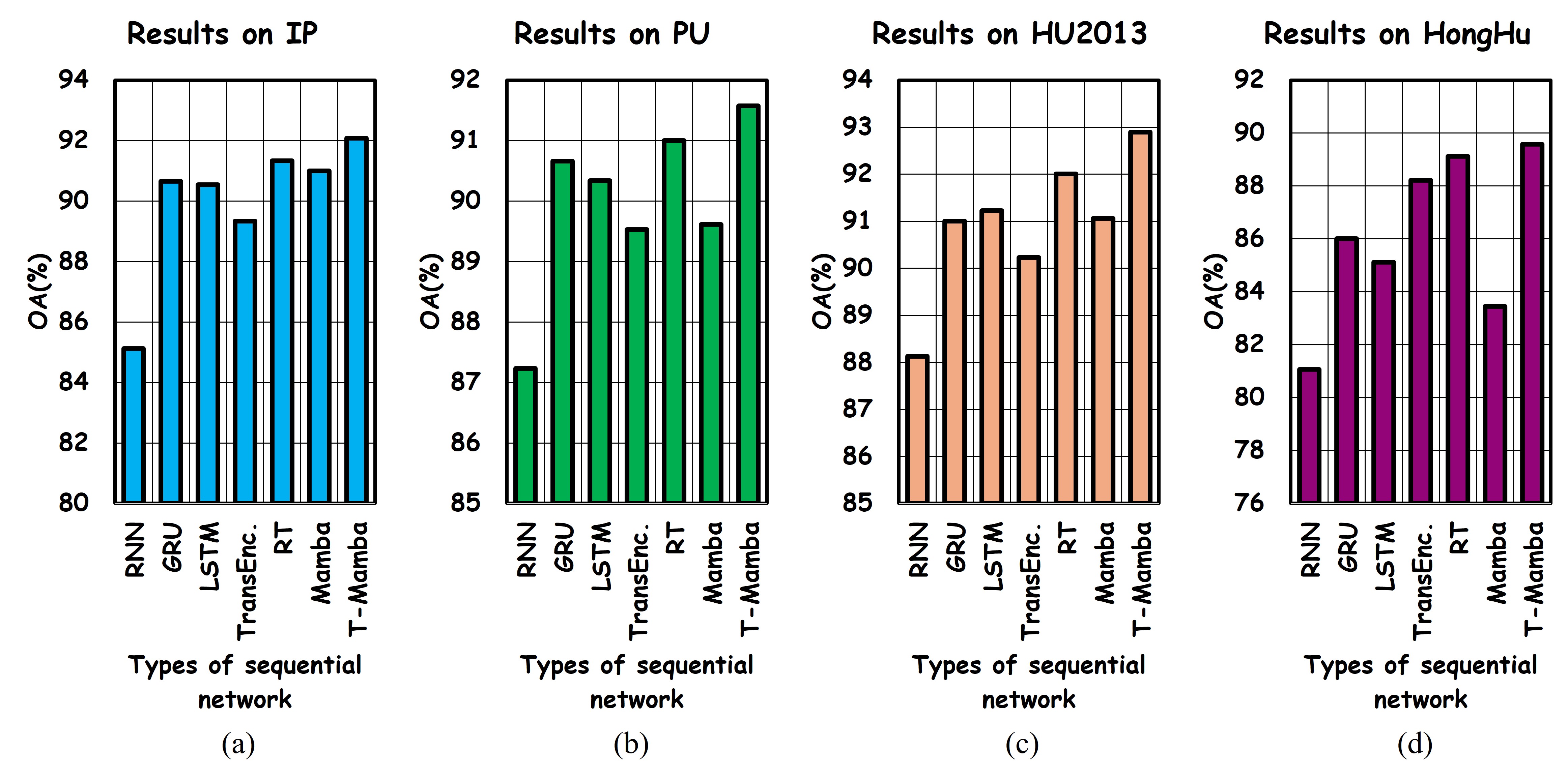}
    \caption{The comparison between using different sequential networks (i.e., RNN, GRU, LSTM, Transformer encoder, RT, Mamba, and T-Mamba). We have to note that all networks are not sharing parameters for four types of MCS in MiM model. Experimental results are listed in (a) IP dataset, (b) PU dataset, (c) HU2013 dataset, and (d) HongHu dataset. }
    \label{fig:comparison of sequential models_1}
\end{figure}
\begin{figure}[t]
    \centering
    \includegraphics[scale=0.5]{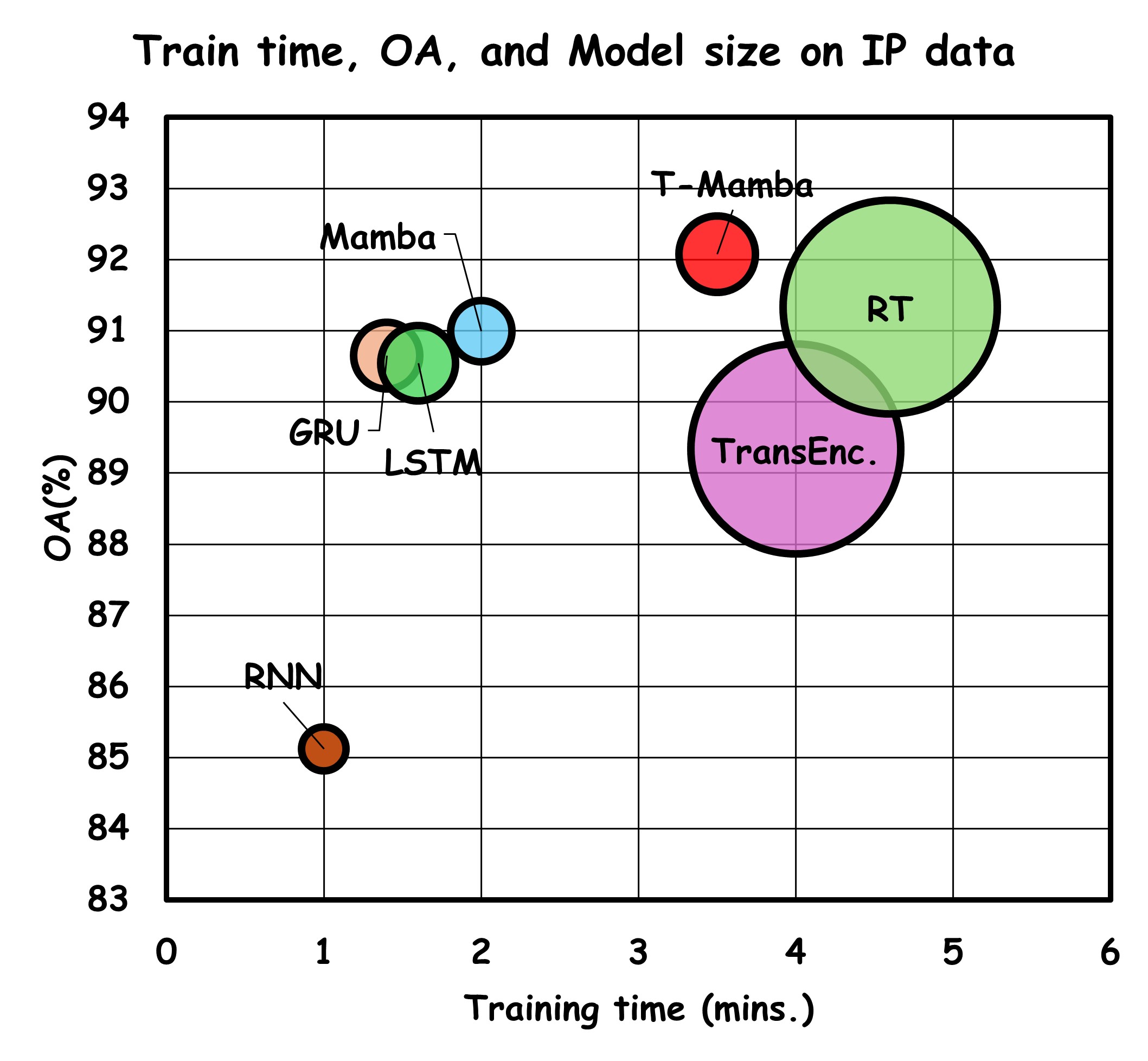}
    \caption{The comparison on the training time, OA(\%) and model size by using different sequential networks, aligning with Fig. \ref{fig:comparison of sequential models_1}.}
    \label{fig:comparison of sequential models_2}
\end{figure}

\subsubsection{\textbf{Comparison with Other Sequential Networks}}
Following the development of the T-Mamba, we conducted comparative experiments against other established sequential networks: RNN, GRU, LSTM, general Transformer encoder, RT (a hybrid of LSTM and Transformer) \cite{multi-rt}, and a general Mamba model \cite{mamba}. These models all process vector inputs at each timestep, allowing us to integrate them directly into our system for a straightforward comparison. We have to note that all networks are not sharing parameters for four types of MCS in MiM model. All tests were performed under identical conditions, with the results shown in Figs. \ref{fig:comparison of sequential models_1} and \ref{fig:comparison of sequential models_2}.  

The vanilla RNN displayed the lowest performance, constrained by its basic approach to processing sequential data. Both GRU and LSTM showed similar results, but the GRU was more efficient due to its simpler structure with one fewer gate than the LSTM. The RT model, which merges LSTM's sequential processing with the Transformer's self-attention, exhibited enhanced performance, underscoring the benefits of combining these two approaches. Although RT outperformed the standalone LSTM and GRU, it requires a significantly larger parameter set.

Our newly designed T-Mamba, which enhances the basic Mamba encoder with tokenization capabilities, does add some parameters but achieves considerable improvements in classification performance. In comparison to the other models, T-Mamba strikes an optimal balance between model size and efficacy. It surpasses the general Transformer encoder in efficiency, offering a more lightweight and robust solution. While RT provides competitive performance, its larger model size and higher resource consumption are less than ideal, making T-Mamba the more advantageous choice in terms of efficiency and effectiveness.

\begin{sidewaystable}[t]
\centering
\caption{Different classification methods in terms of OA on the \underline{\textbf{Indian Pines}} dataset with \underline{\textbf{data augmentation}}. The \textcolor{red}{\textbf{best one}} is highlighted as red. The \textcolor{blue}{\textbf{second best}} is highlighted as blue. \label{tab:data augment}}

\begin{tabular}{|c||c||c|c|c|c|c|c|c|c|c|}
\hline
                 & Num. Train & vanilla ViT & SpeFormer & MAEST & HiT & SSFTT & 3DViT & HUSST & ViM & MiM\\ \hline
                
Scena. 1 & 1024 & 85.163 & 94.123 & 97.323 & 97.211 & 97.002 & \textcolor{blue}{\textbf{97.328}} & 96.892 & 94.317 & \textcolor{red}{\textbf{97.332}} \\ \hline
Scena. 2 & 4170 & 90.669 & 96.128 & \textcolor{red}{\textbf{99.002}} & 98.386 & \textcolor{blue}{\textbf{98.991}} & 98.001 & 97.669 & 95.659 & 98.239 \\ \hline
\end{tabular}

\end{sidewaystable}

\subsubsection{\textbf{Further Discussion on Mamba and Transformer under Data Augmentation}}
Mamba was introduced as an alternative to the Transformer for efficient and lightweight sequential feature encoding. However, unlike the Transformer, which employs global attention, Mamba processes sequences in a linear-like fashion. The Transformer's self-attention mechanism has gained widespread adoption across various applications due to its ability to generalize well with sufficient data, making extensive training datasets a necessity for optimal performance. Against this backdrop, we conducted further experiments to evaluate Mamba-based methods against Transformer-based approaches.

These experiments centered on the augmentation of training samples. We established two scenarios: 1) using a general sample of 10\% randomly selected from the ground truth, a common practice in HSI classification, and 2) augmenting the data through geometric transformations—45°, 90°, and 135° rotations, and vertical and horizontal flips—as outlined in \cite{hsimamba}.

Taking the IP dataset as an example, where labeled pixels across land-cover categories are typically imbalanced and scarce, we initially selected 1024 training samples in the scenario-1, significantly more than in previous experiments with disjoint and fixed training sets. For the augmented scenario-2, we applied geometric transformations to disjoint samples, yielding a substantially increased pool of 4170 training samples (695 multiplied by 6).

The results, detailed in Table \ref{tab:data augment}, indicate that increasing the training sample size to nearly double improves the performance of all models, including our Mamba-based model, although the accuracy gap between methods narrows considerably. This is further evidenced by the shrinking difference in performance between SpeFormer and ViM, two purely model-driven approaches.

However, when the number of training samples are expanded six-fold to 4170, our MiM model was no longer the top performer. Instead, MAEST, which integrates the MAE with prior masks, achieved the best results. Transformer-based models generally performed well. This improvement is likely due to the Transformer's mechanism, which effectively utilizes ample training samples to enhance its recognition and generalization capabilities, as supported by literature suggesting that Transformers can outperform CNNs when sufficient data is available.

Thus, while our MiM model excels in scenarios with limited, disjoint, and fixed samples—conditions that mimic real-world HSI applications. The Transformer-based methods tend to perform better under conditions of abundant training data.

\section{Conclusion\label{sec:conlusion}}
In this study, we introduced a novel and pioneering approach to HSI classification utilizing the newly developed Mamba architecture. Specifically, we implemented a centralized MCS for multi-directional feature aggregation, creating a lighter and more streamlined framework for visual cognition alongside the Mamba sequential model. To further enhance this architecture, we designed a T-Mamba encoder, incorporating bi-directional GDM and a STL to refine feature quality and support downsampling-driven semantic token representation. Meanwhile, porposed STF is effectively fusing learned abstract representations with original sequence features. Additionally, we integrated a WMF module and a multi-scale loss function to optimize decoding efficiency. As the pioneering application of a vision Mamba-based model for HSI classification, our results have significantly outperformed other baselines and state-of-the-art methods with limited and disjoint data samples. However, with augmented training samples, Transformer-based models tend to be better. Moving forward, we plan to extend our idea to include a 3D version of the Mamba model where the scanning consistency on spectral information may work.


\begin{thebibliography}{00}
\bibitem{c1}
M. Ahmad, \textit{et al.}, |Hyperspectral Image Classification—Traditional to Deep Models: A Survey for Future Prospects,| in \textit{IEEE J. Sel. Topics Appl. Earth Observ. Remote Sens.}, vol. 15, pp. 968-999, 2022, doi: 10.1109/JSTARS.2021.3133021.

\bibitem{hyperspectral imaging system}
Y. Tang, \textit{et al.},  |Active and Low-Cost Hyperspectral Imaging for the Spectral Analysis of a Low-Light Environment,| \textit{Sensors}, vol. 2023, no. 3, pp. 1437, 2023, doi: 10.3390/s23031437.

\bibitem{c2}
S. Li, \textit{et al.}, |Deep Learning for Hyperspectral Image Classification: An Overview,| \textit{IEEE Trans. Geosci. Remote Sens.}, vol. 57, no. 9, pp. 6690-6709, 2019, doi: 10.1109/TGRS.2019.2907932.

\bibitem{agriculture}
A. Fong, \textit{et al.}, |Farm to Table: Applications for New Hyperspectral Imaging Technologies in Precision Agriculture, Food Quality and Safety,| \textit{2020 Conference on Lasers and Electro-Optics (CLEO)}, San Jose, CA, USA, 2020, pp. 1-2.

\bibitem{environment}
M. Y. Teng, \textit{et al.}, |Investigation of epifauna coverage on seagrass blades using spatial and spectral analysis of hyperspectral images,| in \textit{2013 5th Workshop on Hyperspectral Image and Signal Processing: Evolution in Remote Sensing (WHISPERS 2013)}, Québec, CA, pp. 1-4, 2013, doi: 10.1109/WHISPERS.2013.8080658.

\bibitem{geology}
S. Qian, \textit{et al.}, |Overview of Hyperspectral Imaging Remote Sensing from Satellites,| \textit{Advances in Hyperspectral Image Processing Techniques}, IEEE, pp.41-66, 2023, doi: 10.1002/9781119687788.ch2.

\bibitem{urban_planning}
N. A, \textit{et al.}, |Current Advances in Hyperspectral Remote Sensing in Urban Planning,| \textit{2022 Third International Conference on Intelligent Computing Instrumentation and Control Technologies (ICICICT)}, Kannur, India, pp. 94-98, 2022, doi: 10.1109/ICICICT54557.2022.9917771.

\bibitem{defense}
M. Shimoni, \textit{et al.}, |Hypersectral Imaging for Military and Security Applications: Combining Myriad Processing and Sensing Techniques,| \textit{IEEE Geosci. Remote Sens. Mag.}, vol. 7, no. 2, pp. 101-117, 2019, doi: 10.1109/MGRS.2019.2902525. 

\bibitem{c7}
P. Ghamisi, \textit{et al.}, |Advances in Hyperspectral Image and Signal Processing: A Comprehensive Overview of the State of the Art,| \textit{IEEE Geosci. Remote Sens. Mag.}, vol. 5, no. 4, pp. 37-78, 2017, doi: 10.1109/MGRS.2017.2762087.

\bibitem{c8}
L. He,  \textit{et al.}, |Recent Advances on Spectral–Spatial Hyperspectral Image Classification: An Overview and New Guidelines,| \textit{IEEE Trans. Geosci. Remote Sens.}, vol. 56, no. 3, pp. 1579-1597, 2018,
doi: 10.1109/TGRS.2017.2765364.

\bibitem{c9}
Y. Chen, \textit{et al.}, |Deep Learning-Based Classification of Hyperspectral Data,| \textit{IEEE J. Sel. Topics Appl. Earth Observ. Remote Sens.}, vol. 7, no. 6, pp. 2094-2107, 2014, doi: 10.1109/JSTARS.2014.2329330.

\bibitem{c10}
Z. Zheng, \textit{et al.}, |FPGA: Fast Patch-Free Global Learning Framework for Fully End-to-End Hyperspectral Image Classification,| \textit{IEEE Trans. Geosci. Remote Sens.}, vol. 58, no. 8, pp. 5612-5626, 2020, doi: 10.1109/TGRS.2020.2967821.

\bibitem{rnn3}
F. Zhou, \textit{et al.}, |Hyperspectral image classification using spectral-spatial LSTMs,| \textit{Neurocomputing}, vol. 328, pp. 39-47, 2019, doi: 10.1016/j.neucom.2018.02.105.

\bibitem{rnn4}
R. Hang, \textit{et al.}, |Cascaded Recurrent Neural Networks for Hyperspectral Image Classification,| \textit{IEEE Trans. Geosci. Remote Sens.}, vol. 57, no. 8, pp. 5384-5394, 2019, doi: 10.1109/TGRS.2019.2899129. 

\bibitem{rnn5}
X. Zhang, \textit{et al.}, |Spatial Sequential Recurrent Neural Network for Hyperspectral Image Classification,| \textit{IEEE J. Sel. Topics Appl. Earth Observ. Remote Sens.}, vol. 11, no. 11, pp. 4141-4155, Nov. 2018, doi: 10.1109/JSTARS.2018.2844873.

\bibitem{rnn6}
S. Cheng, \textit{et al.}, |Multi-scale hierarchical recurrent neural networks for hyperspectral image classification,| \textit{Neurocomputing}, vol. 294, pp. 82-93, 2018, doi: 10.1016/j.neucom.2018.03.012.

\bibitem{song multidirectional rnn}
T. Song, \textit{et al.}, |MSLAN: A Two-Branch Multidirectional Spectral–Spatial LSTM Attention Network for Hyperspectral Image Classification,| \textit{IEEE Trans. Geosci. Remote Sens.}, vol. 60, pp. 1-14, 2022, doi: 10.1109/TGRS.2022.3176216.

\bibitem{vit}
A. Dosovitskiy, \textit{et al.}, “An image is worth 16x16 words: Transformers for image recognition at scale,” \textit{arXiv preprint arXiv:2010.11929}, 2020, doi:  10.48550/arXiv.2010.11929.

\bibitem{vit1}
D. Hong, \textit{et al.}, |SpectralFormer: Rethinking Hyperspectral Image Classification With Transformers,| \textit{IEEE Trans. Geosci. Remote Sens.}, vol. 60, pp. 1-15, 2022, doi: 10.1109/TGRS.2021.3130716.

\bibitem{vit2}
X. Yang, \textit{et al.}, |Hyperspectral Image Transformer Classification Networks,| \textit{IEEE Trans. Geosci. Remote Sens.}, vol. 60, pp. 1-15, 2022, doi: 10.1109/TGRS.2022.3171551. 

\bibitem{vit3}
D. Ibañez, \textit{et al.}, |Masked Auto-Encoding Spectral–Spatial Transformer for Hyperspectral Image Classification,| \textit{IEEE Trans. Geosci. Remote Sens.}, vol. 60, pp. 1-14, 2022, doi: 10.1109/TGRS.2022.3217892.

\bibitem{vit4}
X. Chen, \textit{et al.}, |Hyperspectral Image Classification Based on Multi-stage Vision Transformer with Stacked Samples,| \textit{2021 IEEE Region 10 Conference (TENCON)}, Auckland, New Zealand, 2021, pp. 441-446, doi: 10.1109/TENCON54134.2021.9707289.

\bibitem{vit5}
Y. Qing, \textit{et al.}, “Improved Transformer Net for Hyperspectral Image Classification,” \textit{Remote Sens.}, vol. 13, no. 11, pp. 2216, 2021, doi: 10.3390/rs13112216.

\bibitem{vit6}
X. Qiao, \textit{et al.}, |Multiscale Neighborhood Attention Transformer With Optimized Spatial Pattern for Hyperspectral Image Classification,| \textit{IEEE Trans. Geosci. Remote Sens.}, vol. 61, pp. 1-15, 2023, doi: 10.1109/TGRS.2023.3314550.

\bibitem{vit7}
J. Li, \textit{et al.}, |SCFormer: Spectral Coordinate Transformer for Cross-Domain Few-Shot Hyperspectral Image Classification,| \textit{IEEE Trans. Image Process}, vol. 33, pp. 840-855, 2024, doi: 10.1109/TIP.2024.3351443. 

\bibitem{vit8}
L. Sun, \textit{et al.}, |Spectral–Spatial Feature Tokenization Transformer for Hyperspectral Image Classification,| \textit{IEEE Trans. Geosci. Remote Sens.}, vol. 60, pp. 1-14, 2022, doi: 10.1109/TGRS.2022.3144158. 

\bibitem{vit9}
W. Zhou, \textit{et al.}, |Rethinking Unified Spectral-Spatial-Based Hyperspectral Image Classification Under 3D Configuration of Vision Transformer,| \textit{2022 IEEE International Conference on Image Processing (ICIP)}, Bordeaux, France, 2022, pp. 711-715, doi: 10.1109/ICIP46576.2022.9897603.

\bibitem{vit10}
W. Zhou, \textit{et al.}, |Hierarchical Unified Spectral-Spatial Aggregated Transformer for Hyperspectral Image Classification,| \textit{2022 26th International Conference on Pattern Recognition (ICPR)}, Montreal, QC, Canada, 2022, pp. 3041-3047, doi: 10.1109/ICIP46576.2022.9897603.

\bibitem{vit11}
M. Jiang, \textit{et al.}, |GraphGST: Graph Generative Structure-Aware Transformer for Hyperspectral Image Classification,| \textit{IEEE Trans. Geosci. Remote Sens.}, vol. 62, pp. 1-16, 2024, doi: 10.1109/TGRS.2023.3349076. 

\bibitem{vit12}
B. Tu, \textit{et al.}, |Local Semantic Feature Aggregation-Based Transformer for Hyperspectral Image Classification,| \textit{IEEE Trans. Geosci. Remote Sens.}, vol. 60, pp. 1-15, 2022, doi: 10.1109/TGRS.2022.3201145.

\bibitem{vit13}
S. Mei, \textit{et al.}, |Hyperspectral Image Classification Using Group-Aware Hierarchical Transformer,| \textit{IEEE Trans. Geosci. Remote Sens.}, vol. 60, pp. 1-14, 2022, doi: 10.1109/TGRS.2022.3207933. 

\bibitem{vit14}
S. K. Roy, \textit{et al.}, |Spectral–Spatial Morphological Attention Transformer for Hyperspectral Image Classification,| \textit{IEEE Trans. Geosci. Remote Sens.}, vol. 61, pp. 1-15, 2023, doi: 10.1109/TGRS.2023.3242346.

\bibitem{vit15}
X. Zhang, \textit{et al.}, |A Lightweight Transformer Network for Hyperspectral Image Classification,|  \textit{IEEE Trans. Geosci. Remote Sens.}, vol. 61, pp. 1-17, 2023, doi: 10.1109/TGRS.2023.3297858. 

\bibitem{self-attention}
Z. Niu, \textit{et al.}, |A review on the attention mechanism of deep learning.| \textit{Neurocomputing}, 452, pp. 48-62, 2021, doi: 10.1016/j.neucom.2021.03.091.

\bibitem{cnn1}
A. Ben Hamida, \textit{et al.}, |3-D Deep Learning Approach for Remote Sensing Image Classification,| \textit{IEEE Trans. Geosci. Remote Sens.}, vol. 56, no. 8, pp. 4420-4434, 2018, doi: 10.1109/TGRS.2018.2818945.

\bibitem{cnn2}
Y. Li, \textit{et al.}, “Spectral–Spatial Classification of Hyperspectral Imagery with 3D Convolutional Neural Network,” \textit{Remote Sensing}, vol. 9, no. 1, pp. 67, 2017, doi: 10.3390/rs9010067

\bibitem{cnn3}
M. He, \textit{et al.}, |Multi-scale 3D deep convolutional neural network for hyperspectral image classification,| \textit{2017 IEEE International Conference on Image Processing (ICIP)}, Beijing, China, 2017, pp. 3904-3908, doi: 10.1109/ICIP.2017.8297014.

\bibitem{rnn trans 1}
J. Wensel, \textit{et al.}, |ViT-ReT: Vision and Recurrent Transformer Neural Networks for Human Activity Recognition in Videos,| \textit{IEEE Access}, vol. 11, pp. 72227-72249, 2023, doi: 10.1109/ACCESS.2023.3293813.

\bibitem{rnn trans 2}
M. Gehrig, \textit{et al.}, |Recurrent Vision Transformers for Object Detection with Event Cameras,| \textit{2023 IEEE/CVF Conference on Computer Vision and Pattern Recognition (CVPR)}, Vancouver, BC, Canada, 2023, pp. 13884-13893, doi: 10.1109/CVPR52729.2023.01334.

\bibitem{rnn trans 3}
X. Xue, \textit{et al.}, “Convolutional Recurrent Neural Networks with a Self-Attention Mechanism for Personnel Performance Prediction,” \textit{Entropy}, vol. 21, no. 12, pp. 1227, 2019, doi: 10.3390/e21121227.

\bibitem{rnn trans 4}
Y. Tatsunami, \textit{et al.}, |Sequencer: Deep lstm for image classification.| \textit{Advances in Neural Information Processing Systems 35}, pp: 38204-38217, 2022, doi: 10.48550/arXiv.2205.01972.

\bibitem{multiscanicpr}
W. Zhou, \textit{et al.}, |Multi-Scanning Based Recurrent Neural Network for Hyperspectral Image Classification,| \textit{2020 25th International Conference on Pattern Recognition (ICPR)}, Milan, Italy, 2021, pp. 4743-4750, doi: 10.1109/ICPR48806.2021.9413071.

\bibitem{multi-lstm}
W. Zhou, \textit{et al.}, |Multiscanning Strategy-Based Recurrent Neural Network for Hyperspectral Image Classification,| \textit{IEEE Trans. Geosci. Remote Sens.}, vol. 60, pp. 1-18, 2022, doi: 10.1109/TGRS.2021.3138742.

\bibitem{multi-rt}
W. Zhou, \textit{et al.}, |Multiscanning-Based RNN–Transformer for Hyperspectral Image Classification,|  \textit{IEEE Trans. Geosci. Remote Sens.}, vol. 61, pp. 1-19, 2023, doi: 10.1109/TGRS.2023.3277014.

\bibitem{s4}
A. Gu, \textit{et al.}, “Efficiently modeling long sequences with structured state spaces,” \textit{arXiv preprint arXiv:2111.00396}, 2021, doi: 10.48550/arXiv.2111.00396.

\bibitem{mamba}
A. Gu, \textit{et al.}, “Mamba: Linear-time sequence modeling with selective state spaces,” \textit{arXiv preprint arXiv:2312.00752}, 2023, doi: 10.48550/arXiv.2312.00752.

\bibitem{survey1}
H. Zhang, \textit{et al.}, |A Survey on Visual Mamba.| \textit{Applied Sciences}, vol. 14, no. 13, p.5683, 2024, doi: 10.3390/app14135683.

\bibitem{survey2}
R. Xu, \textit{et al.}, |A Survey on Vision Mamba: Models, Applications and Challenges.| \textit{arXiv preprint arXiv:2404.18861}, 2024, doi: 10.48550/arXiv.2404.18861.

\bibitem{survey3}
X. Liu, \textit{et al.}, |Vision Mamba: A Comprehensive Survey and Taxonomy.| \textit{arXiv preprint arXiv:2405.04404}, 2024, doi: 10.48550/arXiv.2405.04404.

\bibitem{ssm360}
X. Wang, \textit{et al.}, |State Space Model for New-Generation Network Alternative to Transformers: A Survey,| \textit{arXiv preprint arXiv:2404.09516}, 2024, doi: 10.48550/arXiv.2404.09516.

\bibitem{vim}
L. Zhu, \textit{et al.}, |Vision mamba: Efficient visual representation learning with bidirectional state space model,| \textit{arXiv preprint arXiv:2401.09417}, 2024, doi: 10.48550/arXiv.2401.09417.

\bibitem{rsmamba}
K. Chen, \textit{et al.}, |Rsmamba: Remote sensing image classification with state space model,| \textit{IEEE Geosci. Remote Sens. Lett.}, 2024, vol. 21, pp. 1-5, 2024, doi: 10.1109/LGRS.2024.3407111.

\bibitem{vmamba}
Y. Liu, \textit{et al.}, |Vmamba: Visual state space model,| \textit{arXiv preprint arXiv:2401.10166}, 2024, doi: 10.48550/arXiv.2401.10166.

\bibitem{changemamba}
H. Chen, \textit{et al.}, |Changemamba: Remote sensing change detection with spatio-temporal state space model,| \textit{IEEE Trans. Geosci. Remote Sens.}, vol. 62, pp. 1-20, 2024, doi: 10.1109/TGRS.2024.3417253. 

\bibitem{rs3mamba}
X. Ma, \textit{et al.}, |Rs3mamba: Visual state space model for remote sensing images semantic segmentation.| \textit{IEEE Geosci. Remote Sens. Lett.}, vol. 21, pp. 1-5, 2024, doi: 10.1109/LGRS.2024.3414293.

\bibitem{continuous system theory}
E. F. Cellier, \textit{et al.},  |Continuous system modeling,| \textit{Springer Science \& Business Media}, 2013.

\bibitem{ODE}
R. Chen, \textit{et al.}, |Neural ordinary differential equations,| \textit{Advances in neural information processing systems 31}, 2018.

\bibitem{LSSL}
G. Albert, \textit{et al.}, |Combining recurrent, convolutional, and continuous-time models with linear state space layers,| \textit{Advances in neural information processing systems 34}: 572-585, 2021.

\bibitem{softplus}
H. Zheng, \textit{et al.}, |Improving deep neural networks using softplus units,| \textit{2015 International Joint Conference on Neural Networks (IJCNN)}, Killarney, pp. 1-4, 2015, doi: 10.1109/IJCNN.2015.7280459.

\bibitem{rsmamba2}
S. Zhao, \textit{et al.}, |RS-Mamba for Large Remote Sensing Image Dense Prediction,| \textit{arXiv preprint arXiv:2404.02668}, 2024, doi: 10.48550/arXiv.2404.02668.

\bibitem{spectralmamba}
Y. Jing, \textit{et al.}, |Spectralmamba: Efficient mamba for hyperspectral image classification.| \textit{arXiv preprint arXiv:2404.08489}, 2024, doi: 10.48550/arXiv.2404.08489.

\bibitem{hsimamba}
J. Yang, \textit{et al.}, |Hsimamba: Hyperpsectral imaging efficient feature learning with bidirectional state space for classification.| \textit{arXiv preprint arXiv:2404.00272}, 2024, doi: 10.48550/arXiv.2404.00272.

\bibitem{dualmamba}
J. Sheng, \textit{et al.}, |DualMamba: A Lightweight Spectral-Spatial Mamba-Convolution Network for Hyperspectral Image Classification.| \textit{arXiv preprint arXiv:2406.07050}, 2024, doi: 10.48550/arXiv.2406.07050.

\bibitem{tokenlearner}
R. Michael, \textit{et al.}, |Tokenlearner: What can 8 learned tokens do for images and videos?.| \textit{arXiv preprint arXiv:2106.11297}, 2021, doi: 10.48550/arXiv.2106.11297.


\end{thebibliography}
\end{document}